\let\oldvec\vec% Store \vec in \oldvec
\documentclass[natbib,twocolumn]{svjour3} 
\let\vec\oldvec% Restore \vec from \oldvec
\usepackage[colorlinks = true,
            linkcolor = blue,
            urlcolor  = blue,
            citecolor = blue,
            anchorcolor = blue]{hyperref}
            
\usepackage[pdftex]{graphicx}
\usepackage{amsmath,amssymb,subfig}
\usepackage{multirow,bm}
\usepackage[title]{appendix}

\usepackage{scalerel}
\usepackage{tikz}
\usetikzlibrary{svg.path}
\definecolor{orcidlogocol}{HTML}{A6CE39}
\tikzset{
  orcidlogo/.pic={
    \fill[orcidlogocol] svg{M256,128c0,70.7-57.3,128-128,128C57.3,256,0,198.7,0,128C0,57.3,57.3,0,128,0C198.7,0,256,57.3,256,128z};
    \fill[white] svg{M86.3,186.2H70.9V79.1h15.4v48.4V186.2z}
                 svg{M108.9,79.1h41.6c39.6,0,57,28.3,57,53.6c0,27.5-21.5,53.6-56.8,53.6h-41.8V79.1z M124.3,172.4h24.5c34.9,0,42.9-26.5,42.9-39.7c0-21.5-13.7-39.7-43.7-39.7h-23.7V172.4z}
                 svg{M88.7,56.8c0,5.5-4.5,10.1-10.1,10.1c-5.6,0-10.1-4.6-10.1-10.1c0-5.6,4.5-10.1,10.1-10.1C84.2,46.7,88.7,51.3,88.7,56.8z};
  }
}

\newcommand\orcidicon[1]{\href{https://orcid.org/#1}{\mbox{\scalerel*{
\begin{tikzpicture}[yscale=-1,transform shape]
\pic{orcidlogo};
\end{tikzpicture}
}{|}}}}

\DeclareMathOperator*{\argmin}{arg\,min}
\newcommand{\R}{\mathbb{R}}
\newcommand{\barr}{\mathbf{\bar r}}

\journalname{International Journal of Computer Vision}

\begin{document}

\title{What does 2D geometric information really tell us about 3D face shape?}

%\titlerunning{Short form of title}        % if too long for running head

\author{Anil Bas\textsuperscript{1,2}\orcidicon{0000-0002-3833-6023} \and 
        William A. P. Smith\textsuperscript{1}
}

\authorrunning{A. Bas and W. A. P. Smith} % if too long for running head

\institute{Anil Bas \at
              \email{ab1792@york.ac.uk}
            \and
           William A. P. Smith \at
              \email{william.smith@york.ac.uk}
            \and
           \textsuperscript{1} {Department of Computer Science, University of York, York, UK}\\
           \textsuperscript{2} {Department of Computer Engineering, Marmara University, Istanbul, Turkey}
}

\date{Received: 16 April 2018 / Accepted: 6 July 2019}

\maketitle

\begin{abstract}
A face image contains geometric cues in the form of configurational information and contours that can be used to estimate 3D face shape. While it is clear that 3D reconstruction from 2D points is highly ambiguous if no further constraints are enforced, one might expect that the face-space constraint solves this problem. We show that this is not the case and that geometric information is an ambiguous cue. There are two sources for this ambiguity. The first is that, within the space of 3D face shapes, there are flexibility modes that remain when some parts of the face are fixed. The second occurs only under perspective projection and is a result of perspective transformation as camera distance varies. Two different faces, when viewed at different distances, can give rise to the same 2D geometry. To demonstrate these ambiguities, we develop new algorithms for fitting a 3D morphable model to 2D landmarks or contours under either orthographic or perspective projection and show how to compute flexibility modes for both cases. We show that both fitting problems can be posed as a separable nonlinear least squares problem and solved efficiently. We demonstrate both quantitatively and qualitatively that the ambiguity is present in reconstructions from geometric information alone but also in reconstructions from a state-of-the-art CNN-based method.
\keywords{3D morphable face model \and Shape ambiguity \and Perspective projection \and Landmarks}
\end{abstract}

\section{Introduction}
\label{sec:intro}

A 2D image of a face contains various cues that can be exploited to estimate 3D shape. In this paper, we explore to what degree 2D geometric information allows us to estimate 3D face shape. This is sometimes referred to as ``configurational'' information and includes the relative layout of features (usually encapsulated in terms of the position of semantically meaningful landmark points) and contours (caused by occluding boundaries or texture edges). The advantage of using such cues is that they provide direct information about the shape of the face, without having to model the photometric image formation process and to interpret appearance. 

% Not sure we need this paragraph. It's just making it clear that we're not talking about photometric cues or ambiguities.
Although photometric information does provide a cue to the 3D shape of a face \citep{PAMI}, it is a fragile cue because it requires estimates of lighting, camera properties and reflectance properties making it difficult to apply to ``in the wild'' images. Moreover, in some conditions, the shape-from-shading cue may be entirely absent. Perfectly ambient light cancels out all shading other than ambient occlusion which provides only a very weak shape cue \citep{Prados:09}. For this reason, the use of geometric information has proven very popular in 3D face reconstruction \citep{Blanz:04b,OSPAMI,Patel:09,Knothe:06,cao2014displaced,Bas:16}. Landmark detection on highly uncontrolled face images is now a mature research field with benchmarks \citep{sagonas2016300} providing an indication of likely accuracy. Landmarks are often used to initialise or constrain the fitting of 3D morphable models (3DMMs) to images while denser 2D geometric information such as the occluding boundary are used in some of the state-of-the-art methods.

\begin{figure}[!t]
\centering
\begin{tabular}{@{\hspace{0.05cm}}c@{\hspace{0.05cm}}c@{\hspace{0.05cm}}c@{\hspace{0.05cm}}c@{\hspace{0.05cm}}}
 \includegraphics[clip=true,height=2.3cm]{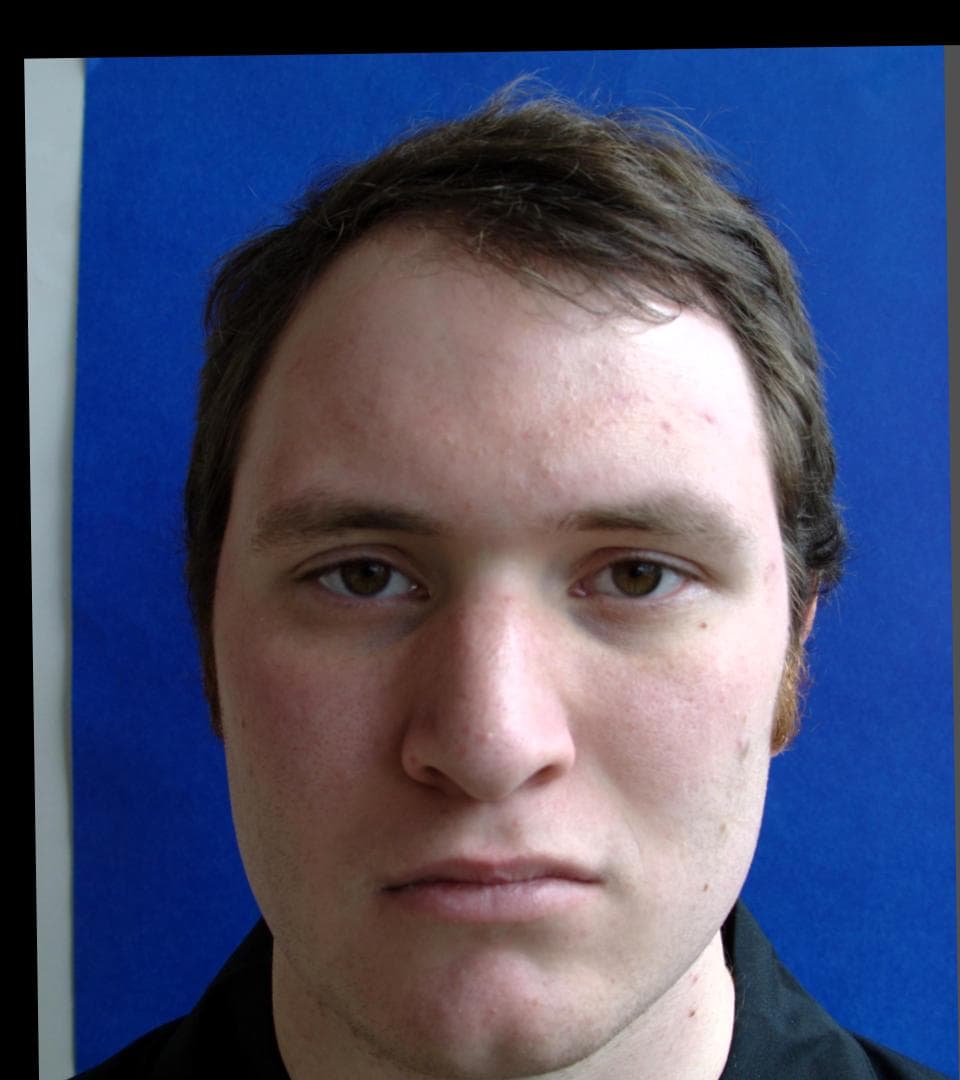} & \includegraphics[clip=true,height=2.3cm]{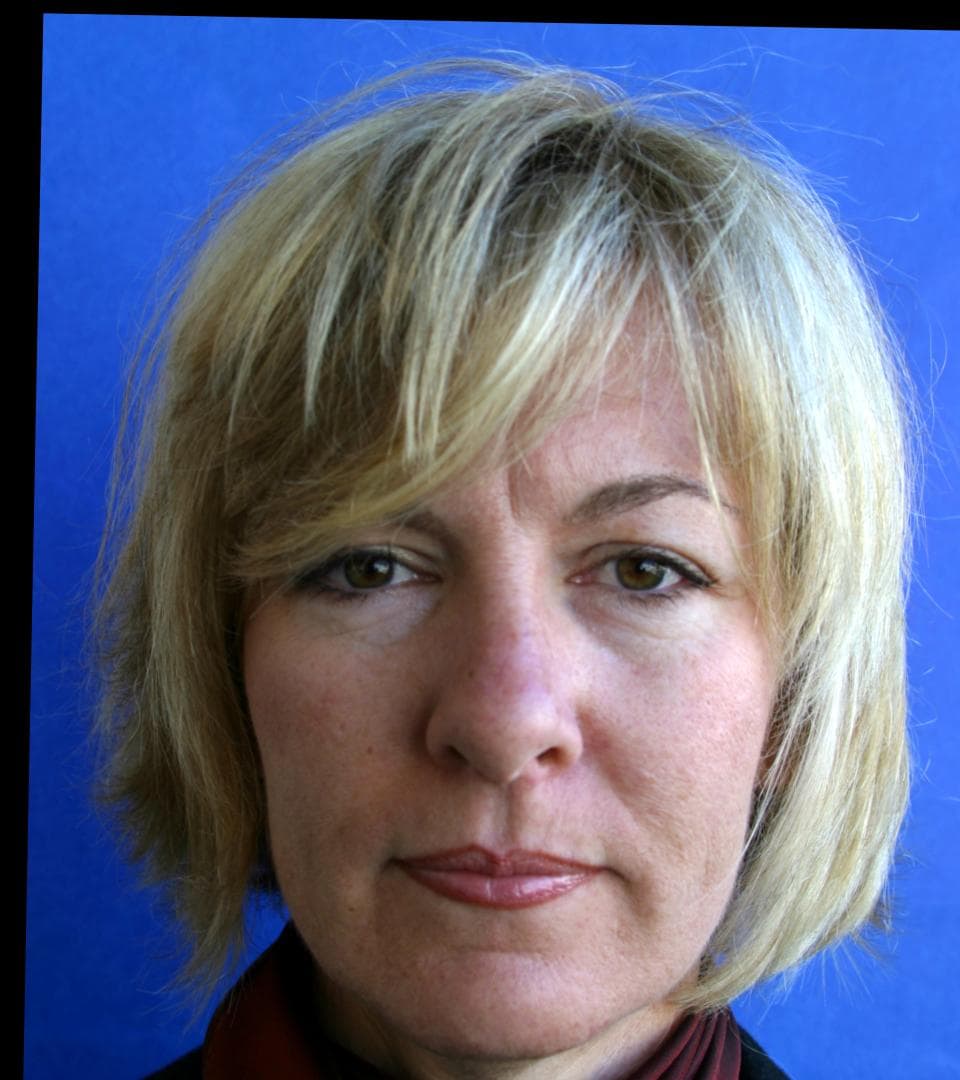} & \includegraphics[clip=true,height=2.3cm]{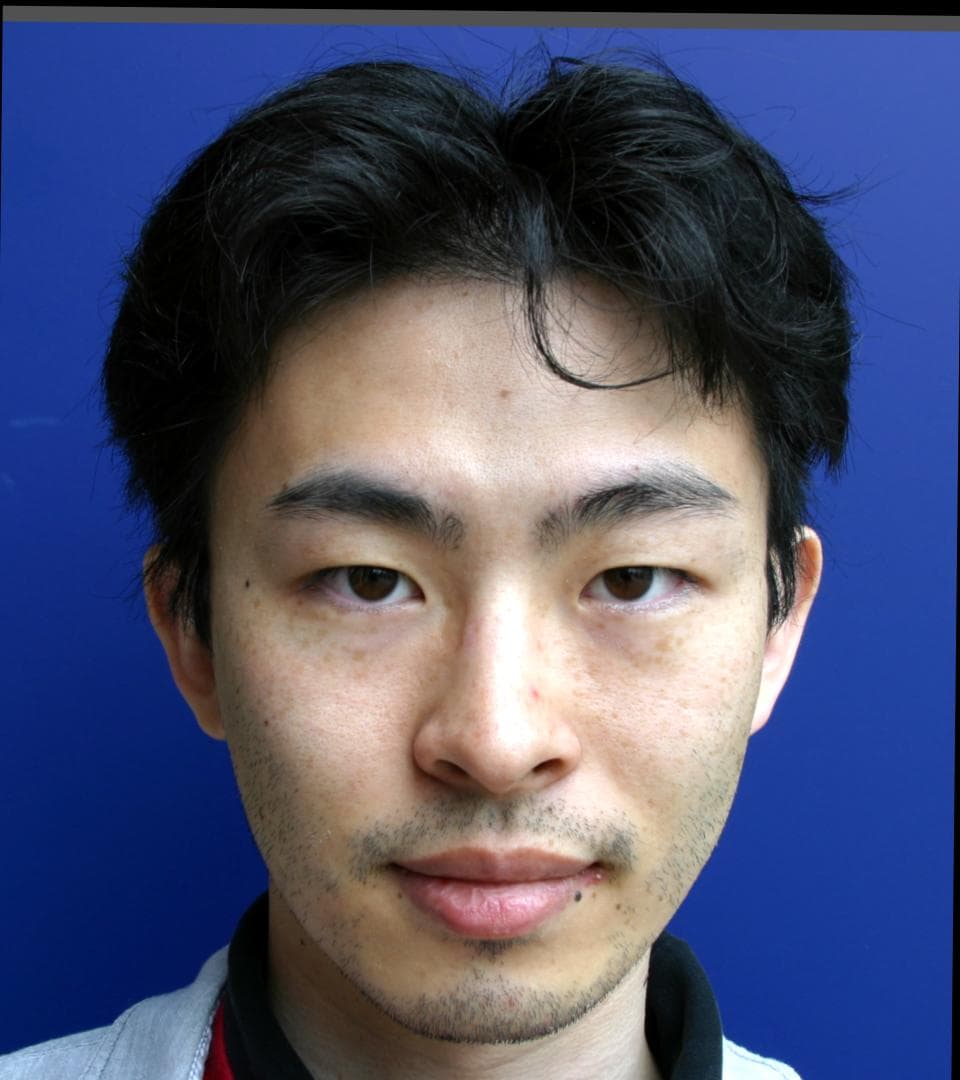} & \includegraphics[clip=true,height=2.3cm]{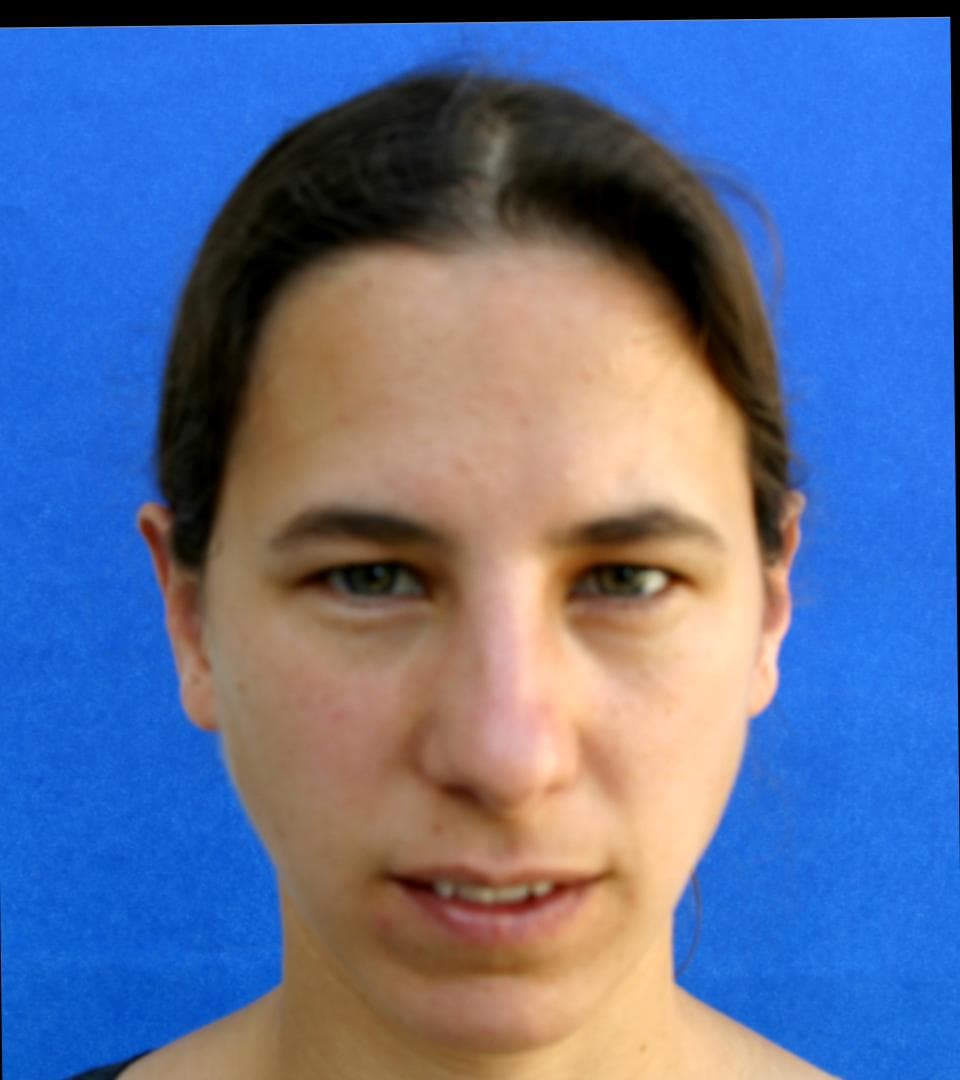} \\
\includegraphics[clip=true,height=2.3cm]{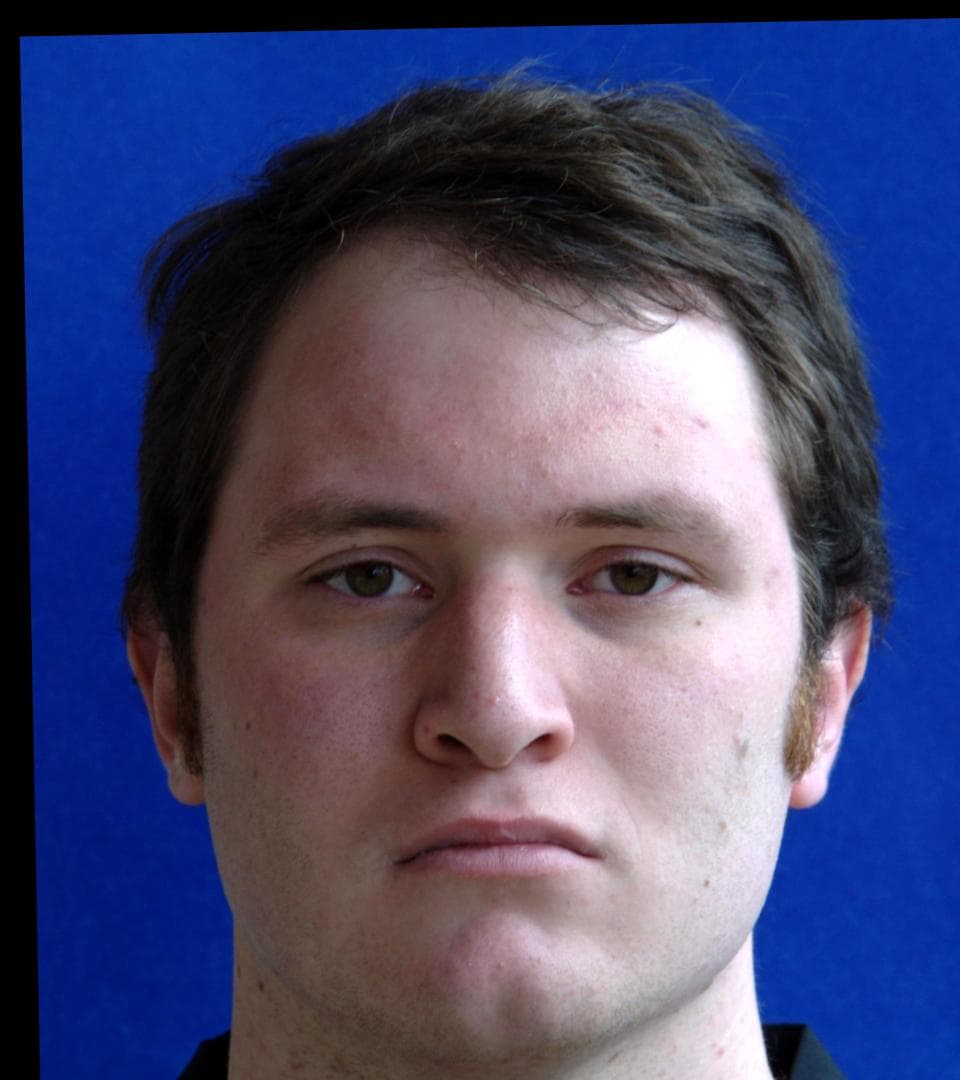} & \includegraphics[clip=true,height=2.3cm]{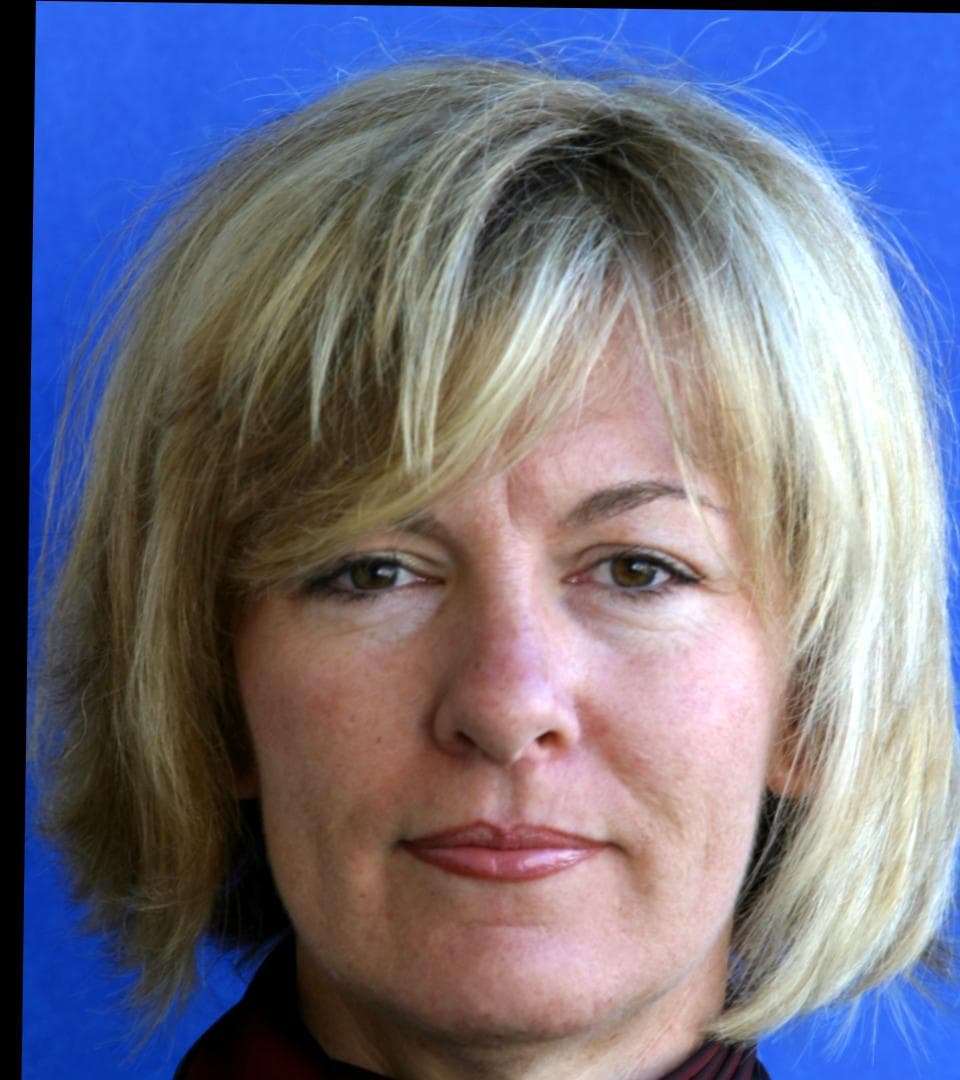} & \includegraphics[clip=true,height=2.3cm]{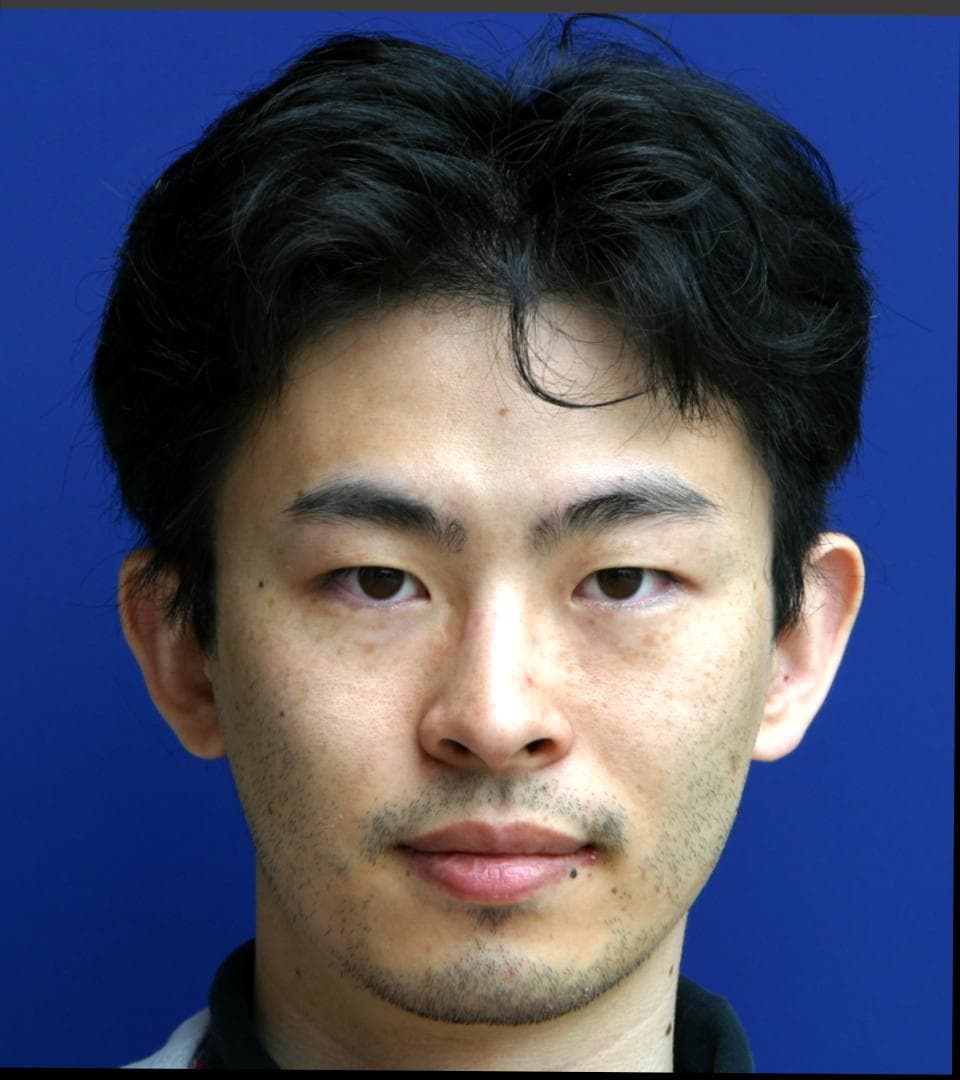} & \includegraphics[clip=true,height=2.3cm]{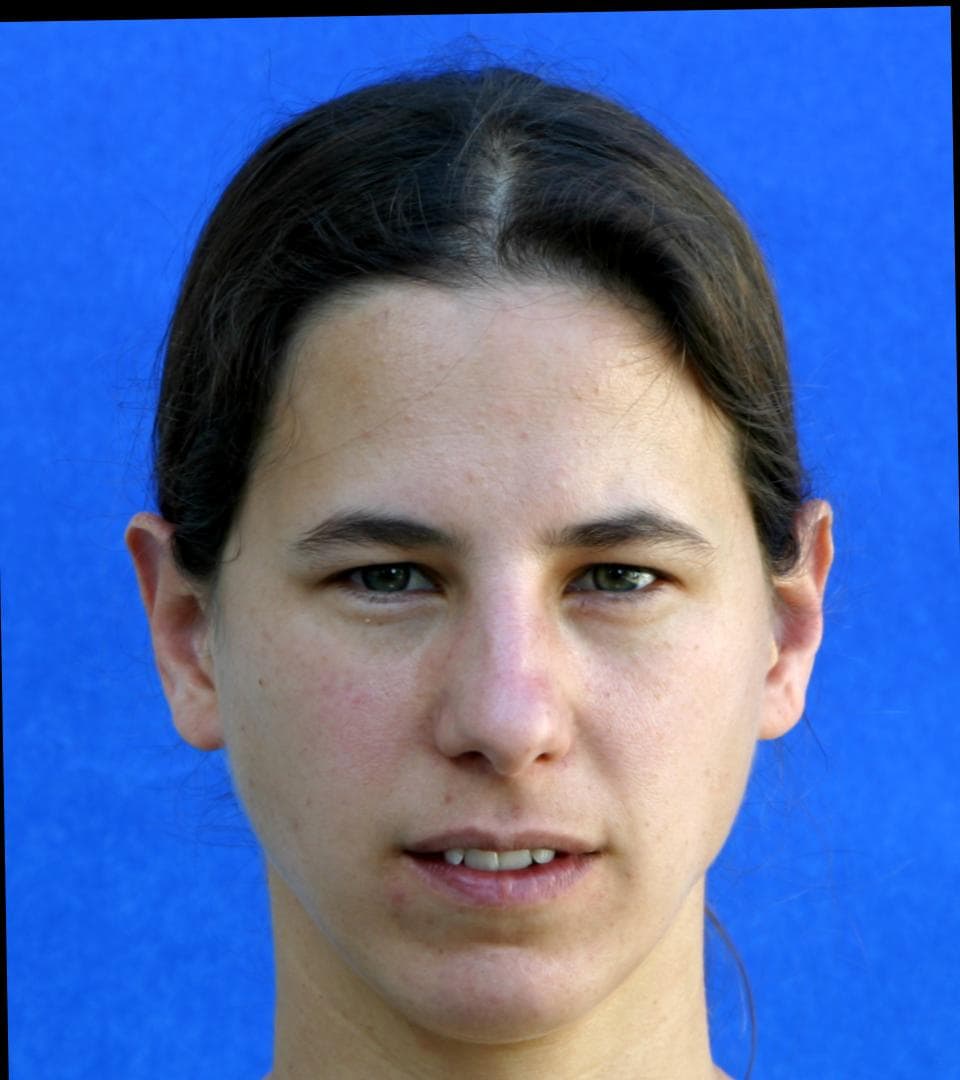} \\
\end{tabular}
\caption{Perspective transformation of real faces from the CMDP dataset \citep{Burgos:14}. The subject is the same in each column and the same camera and lighting is used. The change in viewing distance (60cm top row, 490cm bottom row) induces a significant change in projected shape.}
\label{fig:examples}
\end{figure}

In this paper we show that 2D geometric information only provides a partial constraint on 3D face shape. In other words, face landmarks or occluding contours are an ambiguous shape cue. Rather than try to explain 2D geometric data with a single, best fitting 3D face, we seek to recover a subspace of possible 3D face shapes that are consistent with the 2D data. ``Consistent'' here means that the model explains the data within the tolerance with which we can hope to locate these features within a 2D image. For example, state-of-the-art automatic face landmarking provides a mean landmark error under 4.5\% of interocular distance for only 50\% of images (according to the second conduct of the 300 Faces in the Wild challenge \citep{sagonas2016300}). We show how to compute this subspace and show that it contains very significant shape variation. The ambiguity arises for two reasons. The first is that, within the space of possible faces (as characterised by a 3DMM) there are degrees of flexibility that do not change the 2D geometric information when projection parameters are fixed (this applies to both orthographic and perspective projection). The second is caused by the nonlinear effect of perspective.

When a human face is viewed under perspective projection, its 2D shape varies with the distance between the camera and subject. The effect of perspective transformation is to distort the relative distances between facial features and can be quite dramatic. When a face is close to the camera, it appears taller and slimmer with the features closest to the camera (nose and mouth) appearing relatively larger and the ears appearing smaller and partially occluded. As distance increases and the shape converges towards the orthographic projection, faces appear broader and rounder with ears that protrude further. We show some examples of this effect in Fig.~\ref{fig:examples}. Images are taken at 60cm and 490cm. Each face is cropped and rescaled such that the interocular distance is the same. The distortion caused by perspective transformation is clearly visible. This effect leads to the second ambiguity. Namely that, two different (but natural) 3D face shapes viewed at different distances can give rise to the same 2D geometric features. 

In order to demonstrate both ambiguities, we propose novel algorithms for fitting a 3DMM to 2D geometric information and extracting the subspace of possible 3D shapes. Our contribution is to observe that, under both orthographic and perspective projection, model fitting can be posed as a separable nonlinear least squares optimisation problem that can be solved efficiently without requiring any problem specific optimisation method, initialisation or parameter tuning. % more here
In addition, we use real face images to verify that the ambiguity is present in actual faces. We show that, on average, 2D geometry is more similar between different faces viewed at the same distance than it is between the same face viewed at different distances.
We present quantitative and qualitative results on synthetic 2D geometric data created by projection of real 3D scans. We also present qualitative results on real images from the Caltech Multi-Distance Portraits (CMDP) dataset \citep{Burgos:14}.

\section{Related Work}
\label{sec:related}

{\bf 3D face shape from 2D geometric information}\ \ Facial landmarks, i.e. points with well defined correspondence between identities, are used in a number of ways in face processing. Most commonly, they are used for registration and normalisation, as is done in training an Active Appearance Model \citep{Cootes:98} or in CNN-based face recognition frameworks \citep{Taigman:14}. For this reason, there has been sustained interest in building feature detectors capable of accurately labelling face landmarks in uncontrolled images \citep{sagonas2016300}.

Motivated by the recent improvements in the robustness and efficiency of 2D facial feature detectors, a number of researchers have used the position of facial landmarks in a 2D image as a cue for 3D face shape. In particular, by fitting a 3DMM to these detected landmarks \citep{Blanz:04b,OSPAMI,Patel:09,Knothe:06}. All of these methods assume an affine camera and hence the problem reduces to a multilinear problem in the unknown shape and camera parameters.
The problem of interpreting 3D face shape from 2D landmark positions is related to the problem of non-rigid structure from motion \citep{hartley2008perspective}. However, in that case, the basis set describing the non-rigid deformations is unknown but multiple views of the deforming object are available. In our case, the basis set is known (it is ``face space'' - represented here by a 3DMM) but only a single view of the face is available. 
Some work has considered other 2D shape features besides landmark points. \cite{Keller:07} fit a 3DMM to contours (both silhouettes and inner contours due to texture, shape and shadowing). \cite{Bas:16} adapt the Iterated Closest Point algorithm to fit to edge pixels with an additional landmark term. They use alternating linear least squares followed by a non-convex refinement. Although not applied to faces, \cite{zhou20153d} propose a convex relaxation of the shape-from-landmarks energy. Several recent works \citep{cao20133d,cao2014displaced,saito2016real} use landmark fitting to generate ground truth to train a direct image-to-shape parameters regressor. Again, the landmark fitting optimisation is performed using alternating minimisation, this time under perspective projection with a given focal length. Interestingly, \cite{cao2014displaced} explicitly note that varying the focal length leads to different shapes and use binary search to find the one that gives lowest residual error.

A related problem is to describe the remaining flexibility in a statistical shape model that is partially fixed. If the position of some points, curves or subset of the surface is known, the goal is to characterise the space of shapes that approximately fit these observations. \cite{Albrecht:08} show how to compute the subspace of faces with the same profile. \cite{Luthi:09} extended this approach into a probabilistic setting.
% Fitting to partial data
%A popular approach is to fit a statistical model to 2D landmarks detected using a feature detector. This is largely 
% Contours

The vast majority of 2D face analysis methods that involve estimation of 3D face shape or fitting of a 3D face model assume a linear camera (such as scaled orthographic/weak perspective or affine) \citep{Blanz:04b,OSPAMI,Patel:09,Knothe:06}. Such a camera does not introduce any nonlinear perspective transformation. While this assumption is justified in applications where the subject-camera distance is likely to be large, any situation where a face may be viewed from a small distance must account for the effects of perspective (particularly common due to the popularity of the ``selfie'' format). For this reason, in this paper we consider both orthographic and perspective camera models.

% Landmarks

% 300 faces in the wild, landmark fitting generally

We emphasise that we study the ambiguities only in a monocular setting and, for the perspective case, assuming no geometric calibration. Multiview constraints would reduce or remove the ambiguity. For example, \cite{Amberg:07} describe an algorithm for fitting a 3DMM to stereo face images. In this case, the stereo disparity cue used in their objective function conveys depth information which helps to resolve the ambiguity. However, note that even here, their solution is unstable when camera parameters are unknown. They introduce an additional heuristic constraint on the focal length, namely they restrict it to be between 1 and 5 times the sensor size. 

{\bf Deep model-based face analysis}\ \ 
While the methods above rely on explicit features such as detected landmarks, state of the art methods for 3DMM fitting use deep convolutional neural networks (CNNs) that can learn to exploit any combination of features. Typically, these methods train a CNN to regress 3DMM parameters directly from an input image using a variety of different forms of supervision. \cite{tran2017regressing} perform supervised, discriminative training by first running a multi-image fitting method \citep{Piotraschke2016} on sets of images of the same person and then training the network to predict these parameters from single images. Their multi-image fitting method is based on weighted averaging of single image fits that are themselves initialised by landmark fitting. This initial landmark fit is subject to the ambiguities described in this paper, though the subsequent use of appearance-based losses may not be. However, the latest state-of-the-art in analysis-by-synthesis based fitting suggests that even using dense appearance information the ambiguity may still exist. \cite{schonborn2017markov} use a sampling approach based on Markov Chain Monte Carlo to estimate the full posterior distribution using a hybrid loss including landmarks and appearance error. They note a very high posterior standard deviation in estimated distance from the camera concluding that the ambiguity under perspective cannot be resolved.

The latest state-of-the-art in regression-based fitting \cite{RingNet:CVPR:2019} relies entirely on landmark reprojection error, again subject to the ambiguities we describe. \cite{tewari2017mofa} propose to use a model-based decoder (differentiable renderer) such that the estimated shape, texture, pose and illumination parameters can be rendered back into an image and a self-supervised appearance loss computed. We draw particular attention to the fact that this method incorporates a landmark loss. The appearance loss only provides a useful gradient for training when already close to a good solution, so the landmark loss is essential to coarsely train the network. This loss is subject to exactly the ambiguities we describe in this paper. In addition, during training, the learning rate on the Z translation (i.e.~subject-camera distance) is set three orders of magnitude lower than all other parameters. In other words, the network essentially learns to reconstruct faces assuming a fixed face distance. The idea of self-supervision has been extended in a number of ways. \cite{tran2018nonlinear} make the 3DMM itself learnable. \cite{tewari2018self} learn a corrective space to add details not captured by the model. \cite{Genova_2018_CVPR} learn to regress from face identity parameters to 3DMM parameters such that the rendered face encodes to similar identity parameters to the original image. 

CNNs have also been used to directly estimate \emph{correspondence} between a 3DMM and a 2D face image, without explicitly estimating 3DMM shape parameters or pose. Unlike landmarks, this correspondence is dense, providing a 2D location for every visible vertex. This was first proposed by \cite{guler2017densereg} who use a fully convolutional network and pose the continuous regression task as a coarse to fine classification task. \cite{yu2017learning} take a similar approach but go further by using the correspondences to estimate 3D face shape by fitting a 3DMM. \cite{wu2018convolutional} learn this fitting process as well.
\cite{sela2017unrestricted} take a multitask learning approach by training a CNN to predict both correspondence and facial depth. 
In all cases, this estimated dense correspondence provides an ambiguous shape cue, exactly as we describe in this paper.

{\bf Faces under perspective projection}\ \ The effect of perspective transformation on face appearance has been studied from both a computational and psychological perspective previously. In psychology, \cite{Liu:03,Liu:06} show that human face recognition performance is degraded by perspective transformation. \cite{perona2007new,bryan2012perspective} investigated a different effect, noting that perspective distortion influences social judgements of faces. In art history, \cite{Latto:07} discuss how uncertainty regarding subject-artist distance when viewing a painting results in distorted perception. They show that perceptions of body weight from face images are influenced by subject-camera distance.

There have been two recent attempts to address the problem of estimating subject-camera distance from monocular, perspective views of a face \citep{Flores:13,Burgos:14}. The idea is that the configuration of projected 2D face features conveys something about the degree of perspective transformation. \cite{Flores:13} approach the problem using exemplar 3D face models. They fit the models to 2D landmarks using perspective-n-point \citep{lepetit2009epnp} and use the mean of the estimated distances as the estimated subject-camera distance. \cite{Burgos:14} on the other hand work entirely in 2D. 
They present a fully automated process for estimating 2D landmark positions to which they apply a linear normalisation. 
Their idea is to describe 2D landmarks in terms of their offset from mean positions, with the mean calculated either across views at different distances of the same face, or across multiple identities at the same distance. They can then perform regression to relate offsets to distance. They compare performance to humans and show that they are relatively bad at judging distance given only a single image.

%Despite the simplicity of their method, they are able to accurately sort 
%In the harder task of estimating distance given a single image of an unseen subject, their approach does less well. 

Our results highlight the difficulty that both of these approaches face. Namely that many interpretations of 2D facial landmarks are possible, all with varying subject-camera distance. We approach the problem in a different way by showing how to solve for shape parameters when the subject-camera distance is known. We can then show that multiple explanations are possible. The perspective ambiguity is hinted at in the literature, e.g.~\cite{booth20183d} state ``we found that it is beneficial to keep the focal length constant in most cases, due to its ambiguity with $t_z$'', but never explored in a rigourous manner.

\cite{fried2016perspective} explore the effect of perspective in a synthesis application. They use a 3D head model to compute a 2D warp to simulate the effect of changing the subject-camera distance, allowing them to approximate appearance at any distance given a single image. \cite{valente2015perspective} also proposed a method to warp a 2D image to compensate for perspective. However, their goal was to improve the performance of face recognition systems that they showed are sensitive to such transformations.

\cite{schumacher2012facial} investigate ambiguities from a perceptual point of view.
They explore whether, after seeing a frontal view, participants accept a 3D reconstruction as the correct profile as often as they do for the original profile. 
It shows that human observers consider the reconstructed shape equally plausible as ground truth, even if it differs significantly from ground truth and even if choices include the original profile of the face.

{\bf Other ambiguities}\ \ 
There are other known ambiguities in the monocular estimation of 3D shape.
% Bas relief
The bas relief ambiguity \citep{Belhumeur:99} arises in photometric stereo with unknown light source directions. A continuous class of surfaces (differing by a linear transformation) can produce the same set of images when an appropriate transformation is applied to the illumination and albedo. 
For the particular case of faces, \cite{Georghiades:01} resolve this ambiguity by exploiting the symmetries and similarities in faces. Specifically they assume: bilateral symmetry; that the forehead and chin should be at approximately the same depth; and that the range of facial depths is about twice the distance between the eyes.

In the {\it hollow face illusion} \citep{Hill:94}, shaded images of concave faces are interpreted as convex faces with inverted illumination. The illusion even holds when the hollow face is moving, with rotations being interpreted in reverse. This is a binary version of the bas relief ambiguity occurring when both convex and concave faces are interpreted as convex so as to be consistent with prior knowledge.

More generally, ambiguities in surface reconstruction have been considered in a number of settings. \cite{ecker2008semidefinite} consider the problem of reconstructing a smooth surface from local information that contains a discrete ambiguity. The ambiguities studied here are in the local surface orientation or gradient, a problem that occurs in photometric shape reconstruction. \cite{salzmann2007deformable} study the ambiguities that arise in monocular nonrigid structure from motion under perspective projection. 

Like us, \cite{moreno2013stochastic} also explore ambiguities in shape-from-landmarks in the context of objects represented by a linear basis (in their case, nonrigid deformations of an object rather than the space of faces). However, unlike in this paper, they assume that the intrinsic camera parameters are known. Hence, they do not model the perspective ambiguity that we describe (in which a change in distance is compensated by a change in focal length). Different to our flexibility modes, instead of analytically deriving a subspace, they use stochastic sampling to explore the set of possible solutions. They attempt to select from within this space using additional information provided by motion or shading.

%\edit{Although both \cite{moreno2013stochastic} and our study both explore the space of feasible solutions, it differs considerably from ours. 
%First,
%Second, they do not consider the ambiguity arises from the effect of perspective transformation. 
%Third, the main aim of our study is to compute and verify that these ambiguities exist if no further constraints are enforced while \cite{moreno2013stochastic} grounded on the assumption of ...}

% NRSFM ambiguities to go here?

In an early version of this work \citep{smith2016perspective}, we considered only the effect of perspective and assumed that rotation and translation were fixed. Here we go further by also considering orthographic projection and showing how to compute flexibility modes. Moreover, we show how model fitting can be posed as a separable nonlinear least squares problem, including solving for rotation and translation, and present more comprehensive experimental results. Finally, we consider not only landmarks but also show how to fit to contours where model-image correspondence is not known.

\begin{table}[t]
\centering
\noindent\resizebox{.48\textwidth}{!}{
\begin{tabular}{|c|p{4.5cm}|l|}
\hline
\textbf{Symbol}         & \textbf{Description}              & \textbf{Object type}         \\ \hline \hline
$ N $                   & No.~3D vertices                   & $\in\mathbb{Z}$\\ \hline
$ S $                   & No.~model dimensions              & $\in\mathbb{Z}$    \\ \hline
$ L $                   & No.~2D landmarks                  & $\in\mathbb{Z}$    \\ \hline
$ {\bf Q} $             & Principal components              & $\in\R^{3N\times S}$  \\ \hline
$ {\bm \alpha} $        & Shape parameter vector            & $\in\R^{S}$   \\ \hline
$ \bar{\bm \varsigma}$  & Mean face shape                   & $\in\R^{3N}$    \\ \hline
$ {\bf v}_{i} $         & $i$th 3D point (vertex)           & $\in\R^3$    \\ \hline
$ {\bf x}_{i} $         & $i$th 2D point                    & $\in\R^2$       \\ \hline
$ \mathbf{P} $          & Orthographic projection matrix    & $\in\R^{2\times 3}$     \\ \hline
${\bf R} $              & Rotation matrix                   & $\in SO(3)$    \\ \hline
$ {\bf r} $             & Axis-angle vector                 & $\in\R^3$     \\ \hline
$ {\bf t} $             & Translation vector                & $\in\R^2$ or $\R^3$    \\ \hline
$ s $                   & Scale                             & $\in\R_{>0}$    \\ \hline
$ f $                   & Focal length                      & $\in\R_{>0}$     \\ \hline
$ \mathbf{K} $          & Camera intrinsics                 & $\in\R^{3\times 3}$    \\ \hline
$ \varepsilon $         & Objective function                & $\in\R_{\geq 0}$    \\ \hline
$ \mathbf{d} $          & Vector of residuals               & $\in\R^{2L}$ or $\R^{3L}$ \\ \hline
$ \mathbf{I}_n $        & Identity matrix                   & $\in\{0,1\}^{n\times n}$  \\ \hline
$ \mathbf{1}_n $        & Column vector of ones             & $\in\{1\}^n$   \\ \hline
$ \mathbf{J} $          & Jacobian matrix                   & $\in\R^{2L\times 4}$ or $\R^{3L\times 4}$   \\ \hline
$ t_z $                 & Face-camera distance              & $\in\R_{>0}$     \\ \hline
$ k $                   & Threshold value                   & $\in\R_{>0}$     \\ \hline
$ {\bm \Pi} $           & 2D Projection                     & $\in\R^{2L\times S}$ or $\R^{3L\times S}$\\ \hline
$ {\bf f} $             & Flexibility modes                 & $\in\R^S$ \\ \hline
$ \lambda_i $           & $i$th eigenvalue                  & $\in\R$ \\ \hline
$ {\cal B} $            & Occluding boundary vertices       & $\subset \{1,\dots,N\}$ \\ \hline
$ \otimes $             & Kronecker product                 & Operator    \\ \hline
\end{tabular}}
\caption{Definition of symbols.}
\label{tab:def}
\end{table}

\section{Preliminaries}

Our approach is based on fitting a 3DMM to 2D landmark observations under either orthographic or perspective projection. Hence, we begin by describing the 3DMM and the scaled orthographic and pinhole projection model. We provide the definition of symbols in Table~\ref{tab:def}.

\subsection{3D Morphable Model}

A 3DMM is a deformable mesh whose vertex positions, ${\bm \varsigma}({\bm \alpha})$, are determined by the shape parameters ${\bm \alpha}\in \R^{S}$. Shape is described by a linear subspace model learnt from data using principal component analysis (PCA) \citep{Blanz:03b}. So, the shape of any object from the same class as the training data can be approximated as:
\begin{equation}
{\bm \varsigma}({\bm \alpha})= {\bf Q}{\bm \alpha}+\bar{\bm \varsigma}, \label{eqn:shapemodel}
\end{equation}
where the vector ${\bm \varsigma}({\bm \alpha})\in \R^{3N}$ contains the coordinates of the $N$ vertices, stacked to form a long vector: ${\bm \varsigma}=[u_{1}, v_{1}, w_{1}, \dots, u_{N}, v_{N}, w_{N}]^{\textrm{T}}$, ${\bf Q}\in\R^{3N\times S}$ contains the $S$ retained principal components and $\bar{\bm \varsigma}\in\R^{3N}$ is the mean shape. Hence, the $i$th vertex is given by: ${\bf v}_{i}=[\varsigma_{3i-2}, \varsigma_{3i-1}, \varsigma_{3i}]^{\textrm{T}}$.

For convenience, we denote the sub-matrix corresponding to the $i$th vertex as ${\bf Q}_i\in\R^{3\times S}$ and the corresponding vertex in the mean face shape as $\bar{\bm \varsigma}_i\in\R^3$, such that the $i$th vertex is given by:
$
{\bf v}_i = {\bf Q}_i{\bm \alpha}+\bar{\bm \varsigma}_i.
$
%Similarly, we define the row corresponding to the $u$ component of the $i$th vertex as ${\bf P}_{iu}$ (similarly for $v$ and $w$) and define the $u$ component of the $i$th mean shape vertex as $\bar{s}_{iu}$ (similarly for $v$ and $w$).

Since the morphable model that we use has meaningful units (i.e. it was constructed from scans where vertex positions were recorded in metres) we do not need a scale parameter to transform from model to world coordinates.

\subsection{Scaled Orthographic Projection}

The scaled orthographic, or weak perspective, projection model assumes that variation in depth over the object is small relative to the mean distance from camera to object. Under this assumption, the projection of a 3D point ${\bf v}=[u, v, w]^{\textrm{T}}$ onto the 2D point ${\bf x}=[x, y]^{\textrm{T}}$ is given by ${\bf x}=\textrm{\bf SOP}[{\bf v},{\bf R},{\bf t}_{\textrm{2d}},s] \in \R^2$ which does not depend on the distance of the point from the camera, but only on a uniform scale $s$ given by the ratio of the focal length of the camera and the mean distance from camera to object:
\begin{equation}
 \textrm{\bf SOP}[{\bf v},{\bf R},{\bf t}_{\textrm{2d}},s] = s\mathbf{P}{\bf Rv}+s{\bf t}_{\textrm{2d}}
\end{equation}
where 
\begin{equation*}
\mathbf{P}=
\begin{bmatrix}
 1 & 0 & 0 \\
 0 & 1 & 0
\end{bmatrix}
\end{equation*}
is a projection matrix and the pose parameters ${\bf R}\in SO(3)$, ${\bf t}_{\textrm{2d}}\in\R^2$ and $s\in\R^+$ are a rotation matrix, 2D translation and scale respectively. In order to constrain optimisation to valid rotation matrices, we parameterise the rotation matrix by an axis-angle vector ${\bf R}({\bf r})$ with ${\bf r}\in\R^3$. The conversion from an axis-angle representation to a rotation matrix is given by:
\begin{equation}
    \mathbf{R}(\mathbf{r}) = \cos\theta\mathbf{I} + \sin\theta \begin{bmatrix} \barr \end{bmatrix}_{\times} + (1-\cos\theta)\barr\barr^{\textrm{T}},\label{eqn:rtoR}
\end{equation}
where $\theta=\|\mathbf{r}\|$ and $\barr=\mathbf{r} / \| \mathbf{r} \|$ and
\begin{equation*}
\begin{bmatrix} \mathbf{a} \end{bmatrix}_{\times} = \begin{bmatrix} 0 & -a_3 & a_2 \\ a_3 & 0 & -a_1 \\ -a_2 & a_1 & 0 \end{bmatrix}
\end{equation*}
is the cross product matrix.

\subsection{Perspective camera model}

The nonlinear perspective projection of the 3D point ${\bf v}=[u, v, w]^{\textrm{T}}$ onto the 2D point ${\bf x}=[x, y]^{\textrm{T}}$ is given by the pinhole camera model ${\bf x}=\textrm{\bf pinhole}[{\bf v},\mathbf{K},\mathbf{R},\mathbf{t}_{\textrm{3d}}] \in \R^2$ where $\mathbf{R}\in SO(3)$ is a rotation matrix and $\mathbf{t}_{\textrm{3d}}=[t_x, t_y, t_z]^{\textrm{T}}$ is a 3D translation vector which relate model and camera coordinates (the extrinsic parameters). The matrix:
\begin{equation*}
\mathbf{K}=\begin{bmatrix}
f & 0 & c_x  \\
0 & f & c_y  \\
0 & 0 & 1 
\end{bmatrix}
\end{equation*}
contains the intrinsic parameters of the camera, namely the focal length $f$ and the principal point $(c_x,c_y)$. We assume that the principal point is known (often the centre of the image is an adequate estimate) and parameterise the intrinsic matrix by its only unknown ${\bf K}(f)$. Note that varying the focal length amounts only to a uniform scaling of the projected points in 2D. This corresponds exactly to the scenario in Fig.~\ref{fig:examples}. There, subject-camera distance was varied before rescaling each image such that the interocular distance was constant, effectively simulating a lack of calibration information.
This nonlinear projection can be written in linear terms by using homogeneous representations $\tilde{\bf v}=[u, v, w, 1]^{\textrm{T}}$ and $\tilde{\bf x}=[x, y, 1]^{\textrm{T}}$:
\begin{equation}
 \gamma\tilde{\bf x}=\mathbf{K}
\begin{bmatrix}
\mathbf{R} & \mathbf{t}_{\textrm{3d}}
\end{bmatrix}
\tilde{\bf v},\label{eqn:linearcammodel}
\end{equation}
where $\gamma$ is an arbitrary scaling factor.

\section{Shape-from-landmarks}\label{sec:fitting}

In this section, we describe a novel method for fitting a 3DMM to a set of 2D landmarks. Here, ``landmarks'' can be interpreted quite broadly. It simply means a point for which both the 2D position and the corresponding vertex in the morphable model are known. Later, we will relax this requirement by showing how to establish these correspondences for points on the occluding boundary that do not have clear semantic meaning in the way that a typical landmark does.

We assume that $L$ 2D landmark positions ${\bf x}_i=\left[x_i, y_i\right]^{\textrm{T}}$ ($i=1\dots L$) have been observed. Without loss of generality, we assume that the $i$th landmark corresponds to the $i$th vertex in the morphable model. 

\begin{figure*}[!t]
\noindent\resizebox{\textwidth}{!}{
\centering
\begin{tabular}{ccccc}
\includegraphics[height=5cm, clip=true]{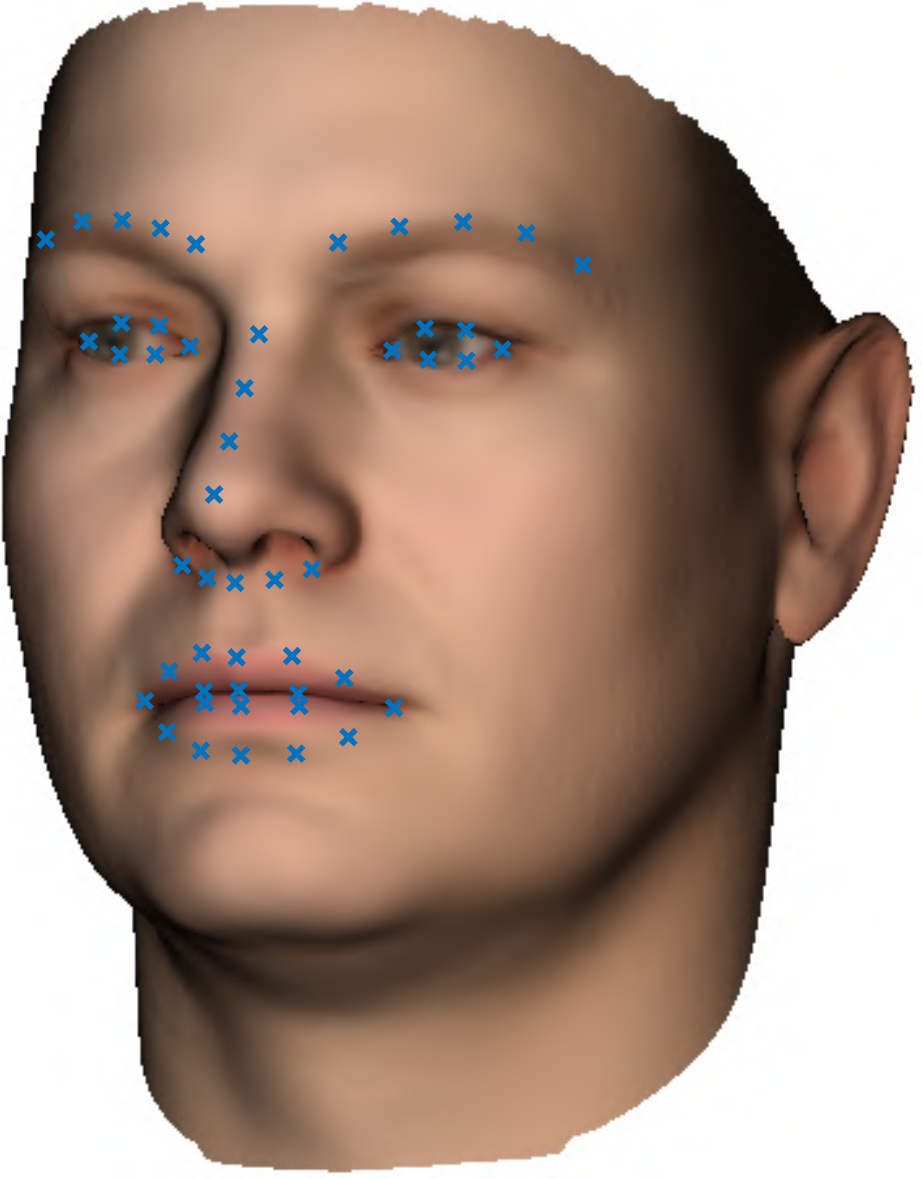}&
\includegraphics[height=5cm, clip=true]{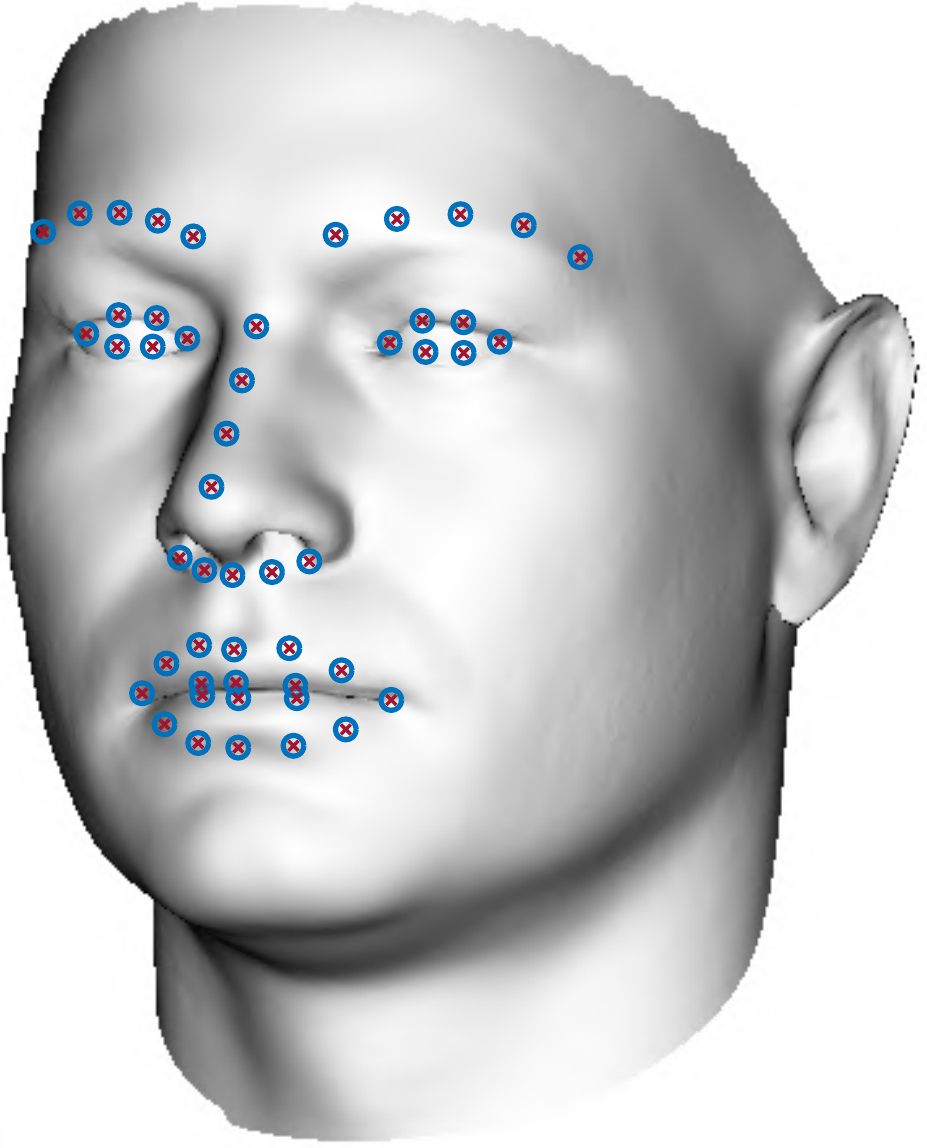}&
\includegraphics[height=5cm, clip=true]{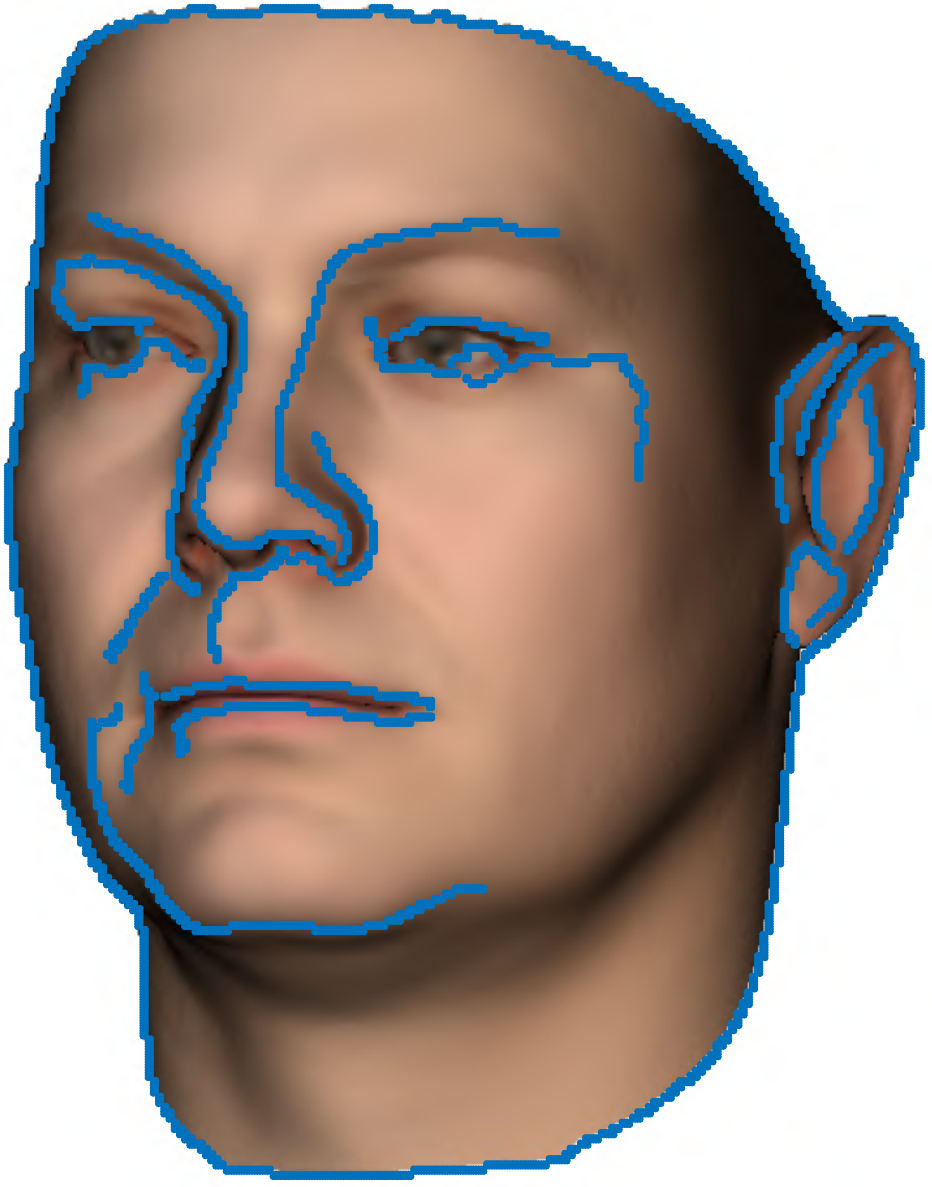}&
\includegraphics[height=5cm, clip=true]{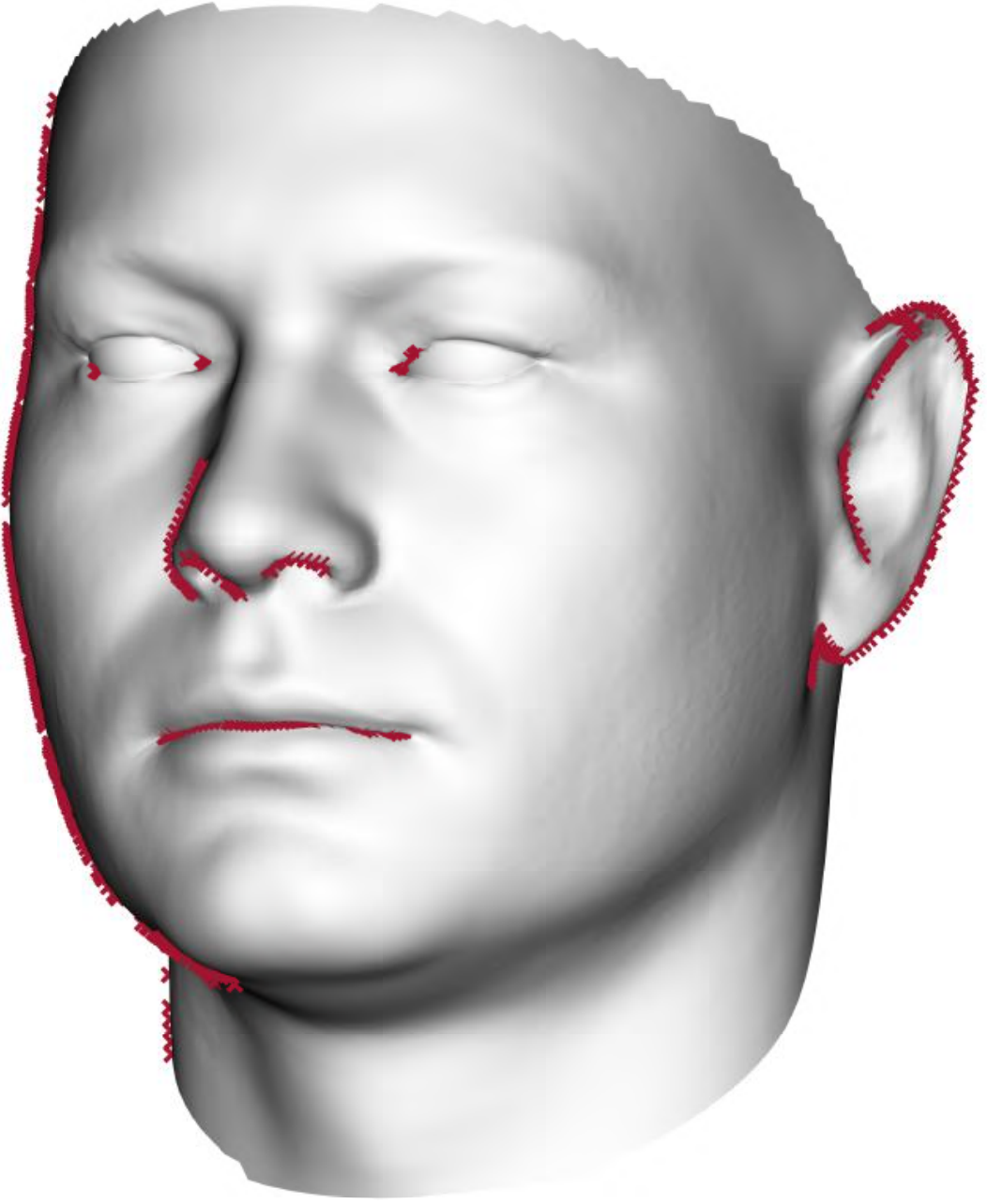}&
\includegraphics[height=5cm, clip=true]{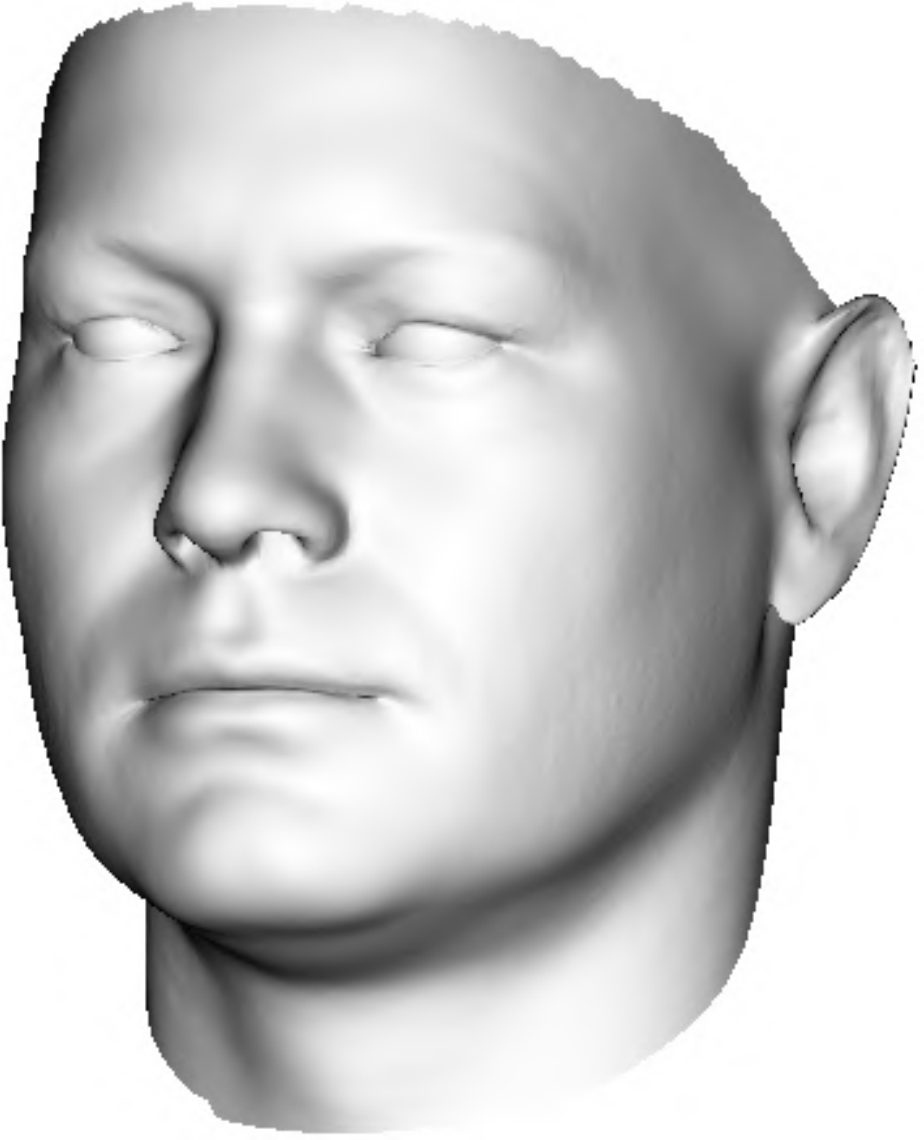}
\end{tabular}}
\caption{Overview of estimating shape from geometric information. From left to right: Input image with landmarks; shape-from-landmarks (Section \ref{sec:fitting}) with image landmarks shown as red crosses and projected model landmarks shown as blue circles;
input image with edge pixels shown in blue;
shape-from-contours (Section \ref{sec:contours}) with occluding boundary vertices labelled with red crosses; 
final reconstruction.}
\label{fig:banner}
\end{figure*}

The objective is to find the shape, pose and camera parameters that, when projected to 2D, minimise the sum of squared distances over all landmarks. We introduce objective functions for the orthographic and perspective cases and then show how they can be expressed as separable nonlinear least squares problems. Fig.~\ref{fig:banner} provides an overview of estimating shape from geometric information.

\subsection{Orthographic objective function}

In the orthographic case, we seek to minimise the following objective function:
\begin{multline}
\varepsilon_{\textrm{ortho}}({\bf r},{\bf t}_{\textrm{2d}},s,{\bm \alpha})= \\
\mathbf{d}_{\textrm{ortho}}({\bf r},{\bf t}_{\textrm{2d}},s,{\bm \alpha})^{\textrm{T}}\mathbf{d}_{\textrm{ortho}}({\bf r},{\bf t}_{\textrm{2d}},s,{\bm \alpha}),\label{eqn:orthoobj}
\end{multline}
where the vector of residuals $\mathbf{d}_{\textrm{ortho}}({\bf r},{\bf t}_{\textrm{2d}},s,{\bm \alpha})\in\R^{2L}$ is given by:
\begin{multline}
\mathbf{d}_{\textrm{ortho}}({\bf r},{\bf t}_{\textrm{2d}},s,{\bm \alpha}) = \\
\begin{bmatrix}
\mathbf{x}_1-\textrm{\bf SOP}\left[{\bf Q}_1{\bm \alpha}+\bar{\bm \varsigma}_1,\mathbf{R}(\mathbf{r}),\mathbf{t}_{\textrm{2d}},s\right] \\
\vdots \\
\mathbf{x}_L-\textrm{\bf SOP}\left[{\bf Q}_L{\bm \alpha}+\bar{\bm \varsigma}_L,\mathbf{R}(\mathbf{r}),\mathbf{t}_{\textrm{2d}},s\right]
\end{bmatrix}. \label{eqn:orthores}
\end{multline}
These residuals are linear in the shape parameters, translation vector and scale but nonlinear in the rotation vector. Previous work has treated this as a multilinear optimisation problem and used alternating coordinate descent. Instead, we observe that the problem can be treated as linear in the shape and translation parameters simultaneously and nonlinear in scale and rotation.

\subsection{Perspective objective function}

In the perspective case, we seek to minimise the following objective function:
\begin{multline}
\varepsilon_{\textrm{persp}}({\bf r},{\bf t}_{\textrm{3d}},f,{\bm \alpha}) = \\
\mathbf{d}_{\textrm{persp}}({\bf r},{\bf t}_{\textrm{3d}},f,{\bm \alpha})^{\textrm{T}}\mathbf{d}_{\textrm{persp}}({\bf r},{\bf t}_{\textrm{3d}},f,{\bm \alpha}), \label{eqn:perspobj}
\end{multline}
where the vector of residuals $\mathbf{d}_{\textrm{persp}}({\bf r},{\bf t}_{\textrm{3d}},f,{\bm \alpha})\in\R^{2L}$ is given by:
\begin{multline}
\mathbf{d}_{\textrm{persp}}({\bf r},{\bf t}_{\textrm{3d}},f,{\bm \alpha}) = \\
\begin{bmatrix}
\mathbf{x}_1-\textrm{\bf pinhole}\left[{\bf Q}_1{\bm \alpha}+\bar{\bm \varsigma}_1,{\bf K}(f),\mathbf{R}(\mathbf{r}),\mathbf{t}_{\textrm{3d}} \right] \\
\vdots \\
\mathbf{x}_L-\textrm{\bf pinhole}\left[{\bf Q}_L{\bm \alpha}+\bar{\bm \varsigma}_L,{\bf K}(f),\mathbf{R}(\mathbf{r}),\mathbf{t}_{\textrm{3d}} \right] 
\end{bmatrix}.\label{eqn:perspres}
\end{multline}
These residuals are nonlinear in all parameters and non-convex due to the perspective projection. However, we can use the direct linear transformation (DLT) \citep{Hartley:03} to transform the problem to a linear one. The solution of this easier problem provides a good initialisation for nonlinear optimisation of the true objective.

From (\ref{eqn:shapemodel}) and (\ref{eqn:linearcammodel}) we have a linear similarity relation for each landmark point:
\begin{equation}
\left[ \begin{array}{c}
{\bf x}_i \\
1
\end{array}
\right]\sim{\bf K}
\left[
\begin{array}{cc}
{\bf R} & {\bf t}
\end{array}
\right]\left[ \begin{array}{c}
{\bf Q}_i{\bm \alpha}+\bar{\bm \varsigma}_i \\
1
\end{array}
\right],
\end{equation}
where $\sim$ denotes equality up to a non-zero scalar multiplication. We rewrite as a collinearity condition:
\begin{equation}
\left[ \begin{array}{c}
{\bf x}_i \\
1
\end{array}
\right]_{\times} {\bf K}
\left[
\begin{array}{cc}
{\bf R} & {\bf t}
\end{array}
\right] \left[ \begin{array}{c}
{\bf Q}_i{\bm \alpha}+\bar{\bm \varsigma}_i \\
1
\end{array}
\right] = {\bf 0} \label{eqn:linpersp}
\end{equation}
where ${\bf 0}=[0\ 0\ 0]^{\textrm{T}}$.
%and $\left[ . \right]_{\times}$ is the cross product matrix:
%\begin{equation}
%\left[ {\bf x} \right]_{\times} = \left[ \begin{array}{ccc}
%0 & -x_3 & x_2 \\
%x_3 & 0 & -x_1 \\
%-x_2 & x_1 & 0 \\
%\end{array}
%\right].
%\end{equation}
This means that each landmark yields three equations that are linear in the unknown shape parameters ${\bm \alpha}$ and the translation vector ${\bf t}_{\textrm{3d}}$.

\subsection{Separable nonlinear least squares}
\label{section:snls}

We now show that both objective functions can be written in a separable nonlinear least squares (SNLS) form, i.e. a form that is linear in some of the parameters (including shape) and nonlinear in the remainder. This special form of least squares problem can be solved more efficiently than general least squares problems and may converge when the original problem would diverge \citep{golub2003separable}. SNLS problems are solved by optimising a nonlinear least squares problem only in the nonlinear parameters, hence the problem dimensionality is reduced and the number of parameters that require initial guesses reduced.
For convenience, henceforth we denote by ${\bf Q}_L\in\R^{3L\times S}$ the submatrix of ${\bf Q}$ containing the rows corresponding to the $L$ landmarks (i.e. the first $3L$ rows of ${\bf Q}$).

\subsubsection{Orthographic}

The vector of residuals \eqref{eqn:orthores} in the orthographic objective function \eqref{eqn:orthoobj} can be written in SNLS form as
\begin{equation}
\mathbf{d}_{\textrm{ortho}}({\bf r},{\bf t}_{\textrm{2d}},s,{\bm \alpha}) = \mathbf{A}(\mathbf{r},s)\begin{bmatrix} {\bm \alpha} \\ {\bf t}_{\textrm{2d}} \end{bmatrix} - \mathbf{y}(\mathbf{r},s) \label{eqn:orthoresSLS}
\end{equation}
where $\mathbf{A}(\mathbf{r},s) \in \R^{2L\times S+2}$ is given by
\begin{equation}
\mathbf{A}(\mathbf{r},s) = s\begin{bmatrix} \left( \mathbf{I}_L \otimes \left[\mathbf{P}\mathbf{R}(\mathbf{r})\right] \right)\mathbf{Q}_L & \mathbf{1}_L \otimes \mathbf{I}_2 \end{bmatrix},
\end{equation}
and $\mathbf{y}(\mathbf{r},s) \in \R^{2L}$ is given by
\begin{equation}
\mathbf{y}(\mathbf{r},s) = s \left( \mathbf{I}_L \otimes \left[\mathbf{P}\mathbf{R}(\mathbf{r})\right] \right) \overline{\mathbf{s}}  - [ x_1, y_1, \dots, y_L ]^{\textrm{T}},
\end{equation}
where $\mathbf{I}_L$ is the $L\times L$ identity matrix and $\mathbf{1}_L$ is the length $L$ vector of ones.

Note that the vector of residuals in \eqref{eqn:orthoresSLS} is exactly equivalent to the original one in \eqref{eqn:orthores}.
The optimal solution to the original objective function \eqref{eqn:orthoobj} in terms of the linear parameters is given by:
\begin{equation}
\begin{bmatrix} {\bm \alpha}^* \\ {\bf t}_{\textrm{2d}}^* \end{bmatrix} = \mathbf{A}^+(\mathbf{r},s) \mathbf{y}(\mathbf{r},s)\label{eqn:linoptortho}
\end{equation}
where $\mathbf{A}^+(\mathbf{r},s)$ is the pseudoinverse. Substituting \eqref{eqn:linoptortho} into \eqref{eqn:orthoresSLS} we get a vector of residuals that is exactly equivalent to \eqref{eqn:orthores} but which depends only on the nonlinear parameters:
\begin{equation}
%\min_{\mathbf{r},s} 
\mathbf{d}_{\textrm{ortho}}({\bf r},s) = \mathbf{A}(\mathbf{r},s)\mathbf{A}^+(\mathbf{r},s) \mathbf{y}(\mathbf{r},s) - \mathbf{y}(\mathbf{r},s).%\label{eqn:orthoSLS}
\end{equation}
Substituting this into \eqref{eqn:orthoobj}, we get an equivalent objective function, $\varepsilon_{\textrm{ortho}}({\bf r},s)$, again depending only on the nonlinear parameters. This is a nonlinear least squares problem of very low dimensionality ($[\mathbf{r}\ s]$ is only 4D). We solve this using the trust-region-reflective algorithm for which we require $\mathbf{J}_{\mathbf{d}_{\textrm{ortho}}}({\bf r},s)\in\R^{2L\times 4}$, the Jacobian of the residual function. In Appendix A, we analytically derive $\mathbf{J}_{\mathbf{d}_{\textrm{ortho}}}$. Although computing these derivatives is quite involved, in practice it is still faster than using finite difference approximations.
Once optimal parameters have been obtained by minimising $\varepsilon_{\textrm{ortho}}({\bf r},s)$ then the parameters ${\bm \alpha}^*$ and ${\bf t}^*$ are obtained by \eqref{eqn:linoptortho}. 

If we wish to impose a statistical prior on the shape parameters we can use Tikhonov regularisation, as in \citep{Blanz:04b}, during the solution of \eqref{eqn:linoptortho}.

\subsubsection{Perspective}

The perspective residual function \eqref{eqn:perspres}, linearised via \eqref{eqn:linpersp}, can be written in SNLS form as
\begin{equation}
\mathbf{d}_{\textrm{persp}}^{\textrm{DLT}}({\bf r},{\bf t}_{\textrm{3d}},f,{\bm \alpha}) = 
%\min_{\mathbf{r},\mathbf{t},f,{\bm \alpha}} 
\mathbf{B}(\mathbf{r},f)\begin{bmatrix} {\bm \alpha} \\ {\bf t}_{\textrm{3d}} \end{bmatrix} - \mathbf{z}(\mathbf{r},f)\label{eqn:perspresDLT}
\end{equation}
where $\mathbf{B}(\mathbf{r},f)\in\R^{3L\times S+3}$ is given by:
\begin{equation}
 \mathbf{B}(\mathbf{r},f) = {\bf DE}(f){\bf F}({\bf r}),
\end{equation}
with
\begin{equation*}
 {\bf D} = \textrm{diag}\left( \begin{bmatrix}{\bf x}_1 \\ 1\end{bmatrix}_{\times}, \dots, \begin{bmatrix}{\bf x}_L \\ 1\end{bmatrix}_{\times}\right),\ \ 
 {\bf E}(f) = {\bf I}_L \otimes {\bf K}(f) 
\end{equation*}
and
\begin{equation*}
 {\bf F}({\bf r}) = \begin{bmatrix}
 \left( {\bf I}_L \otimes {\bf R}({\bf r}) \right) {\bf Q}_L & {\bf 1}_L \otimes {\bf I}_3
\end{bmatrix}.
\end{equation*}
The vector $\mathbf{z}(\mathbf{r},f) \in \R^{3L}$ is given by:
\begin{equation*}
\mathbf{z}(\mathbf{r},f) =
 -{\bf D}\left({\bf I}_L \otimes \left[{\bf K}(f)\mathbf{R}(\mathbf{r})\right]\right)\overline{\bf s}
\end{equation*}

Exactly as in the orthographic case, we can write optimal solutions for the linear parameters in terms of the nonlinear parameters and solve a 4D nonlinear minimisation problem in $({\bf r},f)$. In contrast to the orthographic case, this objective is not equivalent to minimisation of the original objective, i.e. the sum of squared perspective reprojection distances in \eqref{eqn:perspobj}. So, we use the SNLS solution to initialise a nonlinear least squares optimisation of the original objective over all parameters, again using trust-region-reflective. In practice, we find that the SNLS solution is already very close to the optimum and that the subsequent nonlinear least squares optimisation usually converges in 2-5 iterations, shown in Fig.~\ref{fig:3b}.

\subsection{Perspective Ambiguities}

Solving the optimisation problems above yields a least squares estimate of the pose and shape of a face, given 2D landmark positions. In Section \ref{sec:flexibility}, we show that for both orthographic and perspective cases, with pose fixed there remain degrees of flexibility that allow the 3D shape to vary without significantly increasing the objective value. However, for the perspective case there is an additional degree of freedom related to the subject-camera distance, i.e. $t_z$. If, instead of allowing $t_z$ to be optimised along with other parameters, we fix it to some chosen value $k$, then we can obtain different shape and pose parameters:
\begin{equation*}
{\bm \alpha}^*(k) = \arg_{{\bm \alpha}}\min_{{\bf r},{\bf t}_{\textrm{3d}},f,{\bm \alpha}} \varepsilon_{\textrm{persp}}({\bf r},{\bf t}_{\textrm{3d}},f,{\bm \alpha}), \ \ \textrm{s.t. }\ t_z=k.
\end{equation*}
Given 2D landmark observations, we therefore have a continuous (nonlinear) space of solutions ${\bm \alpha}^*(k)$ as a function of subject-camera distance. This is the perspective face shape ambiguity. If the mean reprojection error with a value of $k$ other than the optimal one is still smaller than the tolerance of our landmark detector, then shape recovery is ambiguous.

\section{Shape-from-contours}\label{sec:contours}

In order to extend the method in the previous section to also exploit contour information, we follow \cite{Bas:16} and use an iterated closest edge fitting strategy. We assume that manually provided or automatically detected landmarks are available and we initialise by fitting to these using the method in the previous section. Next, we alternate between establishing correspondences and refitting as follows:
\begin{enumerate}
\item Compute occluding boundary vertices for current shape and pose estimate and project to 2D.
\item Correspondence is found between edges detected in the image and the projection of model vertices that lie on the occluding boundary. This is done in a nearest neighbour fashion with some filtering for robustness.
\item With the correspondences to hand, edge vertices can be treated like landmarks with known correspondence and the method from the previous section applied to refit the model (initialising with the nonlinear parameters obtained in the previous iteration and retaining the original landmarks).
\end{enumerate}
These three steps are iterated to convergence.

In detail, we begin by labelling a subset of pixels as edges, stored in the set ${\cal E}=\{(x,y)|(x,y) \textrm{ is an edge}\}$. In practice, we compute edges by applying the Canny edge detector with a fixed threshold to the input image. More robust performance would be obtained by using a problem-specific edge detector such as boosted edge learning. This was recently done for fitting a morphable tooth model to contours in uncontrolled images \citep{wu2016model}. 

Model contours are computed based on the pose and shape parameters as the occluding boundary of the 3D face. The set of occluding boundary vertices, ${\cal B}({\bm \alpha},{\bf r},{\bf t},s)$ (for the orthographic case), are defined as those lying on a mesh edge whose adjacent faces have a change of visibility. This definition encompasses both outer (silhouette) and inner (self-occluding) contours. In addition, we check that potential edge vertices are not occluded by another part of the mesh (using z-buffering) and we ignore edges that lie on a mesh boundary since they introduce artificial edges. In this paper, we deal only with occluding contours (both inner and outer). If texture contours were defined on the surface of the morphable model, it would be straightforward to include these in our approach. 

We find the set of edge/contour pairs, ${\cal N}$, that are mutual nearest neighbours in a Euclidean distance sense in 2D, i.e. $(i^*,(x^*,y^*))\in{\cal N}$ if:
\begin{multline*}
(x^*,y^*)=\\
\argmin_{(x,y)\in{\cal E}} \|[x\ y]^{\textrm{T}} - \textrm{\bf SOP}\left[{\bf Q}_{i^*}{\bm \alpha}+\bar{\bm \varsigma}_{i^*},\mathbf{R}(\mathbf{r}),\mathbf{t}_{\textrm{2d}},s\right] \|^2
\end{multline*}
and
\begin{multline*}
i^*=\\
\argmin_{i\in{\cal B}({\bm \alpha},{\bf r},{\bf t},s)} \|[x^*\ y^*]^{\textrm{T}} - \textrm{\bf SOP}\left[{\bf Q}_i{\bm \alpha}+\bar{\bm \varsigma}_i,\mathbf{R}(\mathbf{r}),\mathbf{t}_{\textrm{2d}},s\right] \|^2.
\end{multline*}
Using mutual nearest neighbours makes the method robust to contours that are partially missed by the edge detector. The perspective case is identical except that the pinhole projection model is used. The correspondence set can be further filtered by excluding some proportion of pairs whose distance is largest or pairs whose distance exceeds a threshold.

\section{Flexibility modes}\label{sec:flexibility}

We now assume that a least squares model fit has been obtained using the method in Section \ref{sec:fitting} (and optionally Section \ref{sec:contours}). This amounts to a shape, ${\bf Q}{\bm \alpha}+\bar{\bm \varsigma}$, determined by the estimated shape parameters and a pose $(\mathbf{r},s,{\bf t}_{\textrm{2d}})$ or $(\mathbf{r},f,{\bf t}_{\textrm{3d}})$ for orthographic or perspective respectively. We now show that there are remaining modes of flexibility in the model fit. Keeping pose parameters fixed, we wish to find perturbations to the shape parameters that change the projected 2D geometry as little as possible (i.e. minimising the increase in the reprojection error of landmark vertices) while changing the 3D shape as much as possible.

Our approach to computing these flexibility modes is an extension of the method of \cite{Albrecht:08}. They considered the problem of flexibility only in a 3D setting where the model is partitioned into a disjoint fixed part and a flexible part. We extend this so that the constraint on the fixed part acts in 2D after orthographic or perspective projection while the flexible part is the 3D shape of the whole face.

In the orthographic case, we define the 2D projection of the principal component directions for the $L$ landmark vertices as:
\begin{equation}
{\bm \Pi}_{\textrm{ortho}} = \left( \mathbf{I}_L \otimes \left(\mathbf{P}\mathbf{R}(\mathbf{r}) \right)\right){\bf Q}_L,
\end{equation}
where ${\bf r}$ is the rotation vector that was estimated during fitting. Intuitively, we seek modes that move the landmark vertices primarily along the projection axis, which depends only on the rotation, and therefore do not move their 2D projection much. Hence, the flexibility modes do not depend on the scale or translation of the fit or even the landmark positions. For the perspective case, we again use the DLT linearisation in (\ref{eqn:linpersp}), leading to the following expression:
\begin{equation}
{\bm \Pi}_{\textrm{persp}} = {\bf D} \left({\bf I}_L \otimes \left( {\bf K}(f)\begin{bmatrix} \mathbf{R}(\mathbf{r}) & {\bf t}_{\textrm{3d}} \end{bmatrix}\mathbf{S}\right)\right){\bf Q}_L,
\end{equation}
where
\begin{equation*}
\mathbf{S} = \begin{bmatrix} 1 & 0 & 0 \\ 0 & 1 & 0 \\ 0 & 0 & 1 \\ 0 & 0 & 0 \end{bmatrix}.
\end{equation*}
Again, $\mathbf{r}$, $f$ and ${\bf t}_{\textrm{3d}}$ are the rotation vector, focal length and translation that were estimated during fitting. By using the DLT linearisation, the intuition here is that we want the camera rays to the landmark vertices to remain as parallel as possible with the homogeneous vectors representing the observed landmarks.

Concretely, we seek flexibility modes, ${\bf f} \in \R^S$, such that ${\bf Q}{\bf f}$ changes as much as possible whilst the 2D projection of the landmarks, given by ${\bm \Pi}_{\textrm{ortho}}{\bf f}$ or ${\bm \Pi}_{\textrm{persp}}{\bf f}$, changes as little as possible. This can be formulated as a constrained maximisation problem:
\begin{equation}
\max_{{\bf f} \in \R^S} \|{\bf Q}{\bf f}\|^2\ \ \textrm{subject to } \|{\bm \Pi}{\bf f}\|^2=c,
\end{equation}
where ${\bm \Pi}$ is one of the projection matrices and $c\in\R^+$ controls how much variation in the 2D projection is allowed (this value is arbitrary since it does not appear in the subsequent flexibility mode computation). Introducing a Lagrange multiplier and differentiating with respect to ${\bf f}$ yields:
\begin{equation}
{\bf Q}^{\textrm{T}}{\bf Q}{\bf f} = \lambda {\bm \Pi}^{\textrm{T}}{\bm \Pi}{\bf f}.
\end{equation}
This is a generalised eigenvalue problem whose solution is a set of flexibility modes ${\bf f}_1, \dots, {\bf f}_S$ along with their corresponding generalised eigenvalue $\lambda_1, \dots, \lambda_S$, sorted in descending order. Therefore, ${\bf f}_1$ is the flexibility mode that changes the 3D shape as much as possible while minimising the change to the projected 2D geometry. If a face was fitted with shape parameters ${\bm \alpha}$ then its shape is varied by adjusting the weight $w$ in: ${\bf Q}({\bm \alpha}+w{\bf f})+\bar{\bm \varsigma}$. 

\begin{figure}[!t] 
\centering
\noindent\resizebox{.49\textwidth}{!}{
\includegraphics[clip]{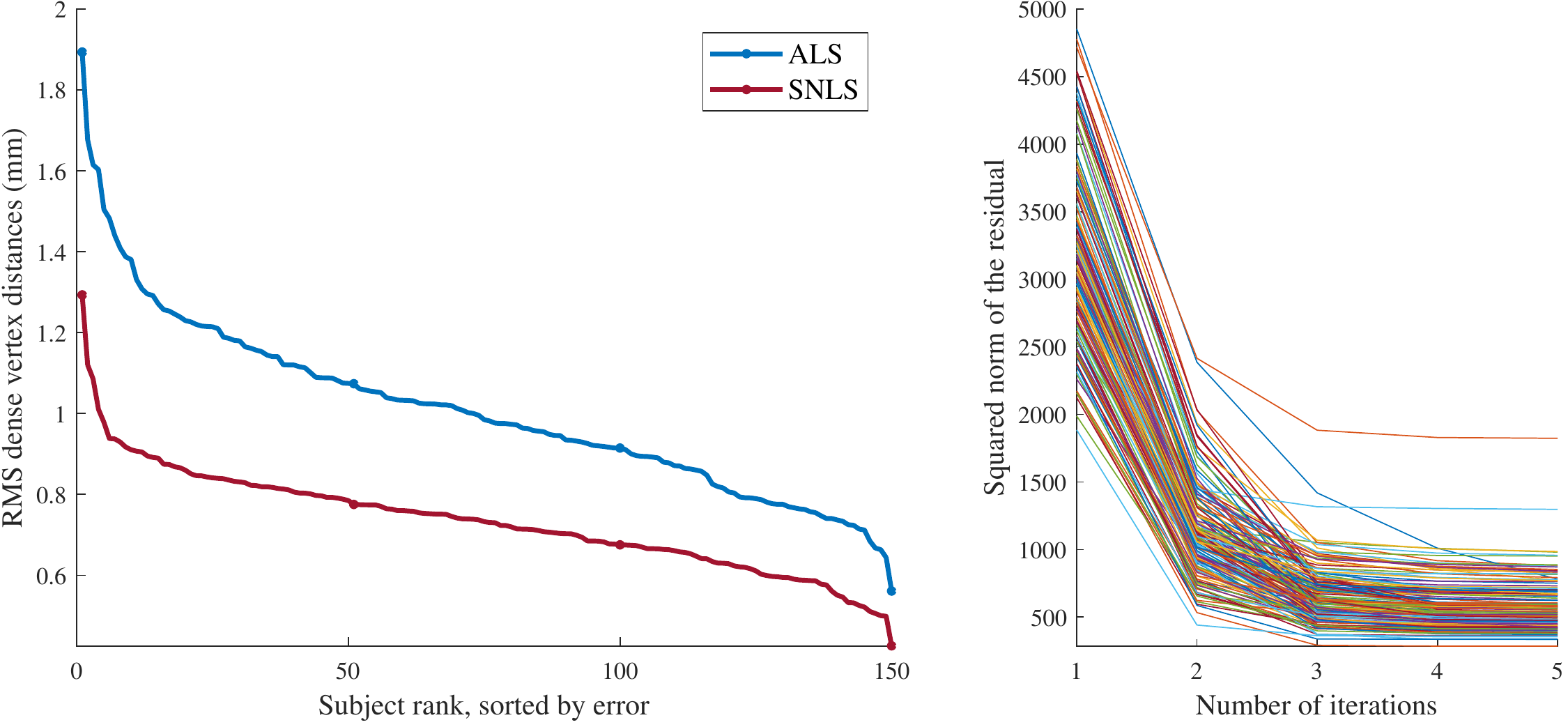}}
%\hspace*{5em}(a)\hspace*{12em}(b
\\[-4mm]
\hspace*{3em}\subfloat[\label{fig:3a}]{}\hspace*{14em}\subfloat[\label{fig:3b}]{}

\caption{(a) Quantitative comparison between alternating linear least squares (ALS) and separable nonlinear least squares (SNLS) on 150 subjects in the Facewarehouse dataset. The average dense surface error is 1.01 mm for ALS and 0.73 mm for SNLS. (b) Convergence rates of nonlinear least squares optimisation.}
\label{fig:alsvssnls}
\end{figure}

We can truncate the number of flexibility modes by setting a threshold $k_1$ on the mean Euclidean distance by which the surface should change and testing whether the corresponding change in mean landmark error is less than a threshold $k_2$. We retain only those flexibility modes where this is the case.

\section{Experimental results}

We now present experimental results to demonstrate the ambiguities that arise in estimating 3D face shape from 2D geometry.
We make use of the Basel Face Model \citep{Paysan:09} (BFM) which is a 3DMM comprising 53,490 vertices and which is trained on 200 faces. We use the shape component of the model only. The model is supplied with 10 out-of-sample faces which are scans of real faces that are in correspondence with the model. We use these for quantitative evaluation on synthetic data. Unusually, the model does not factor out scale, i.e. faces are only aligned via translation and rotation. This means that the vertex positions are in absolute units of distance. This allows us to specify camera-subject distance in physically meaningful units. For all fittings we use Tikhonov regularisation with a low weight.
For sparse (landmark) fitting, where overfitting is more likely, we use $S=70$ dimensions and constrain parameters to be within $k=2$ standard deviations of the mean. For dense fitting, we use all $S=199$ model dimensions and constrain parameters to be $k=3$ standard deviations of the mean.
%and 70 model dimensions for landmark fitting and 199 dimensions for dense fitting.

We make use of two quantitative error measures in our evaluation. For data with ground truth 3D, $d_S$ is the mean Euclidean distance between the ground truth and reconstructed surface after aligning with Procrustes analysis. $d_L$ is the mean distance between observed landmarks and the corresponding projection of the reconstructed landmark vertices, expressed as a percentage of the interocular distance. 
%The interocular distance in these images is 100 pixels so $d_L$ can also be interpreted as the mean landmark error in pixels.

\begin{figure}[!t]
\centering
\subfloat[\label{fig:convergence}]{\includegraphics[height=3.5cm, clip=true,trim=145px 276px 148px 280px]{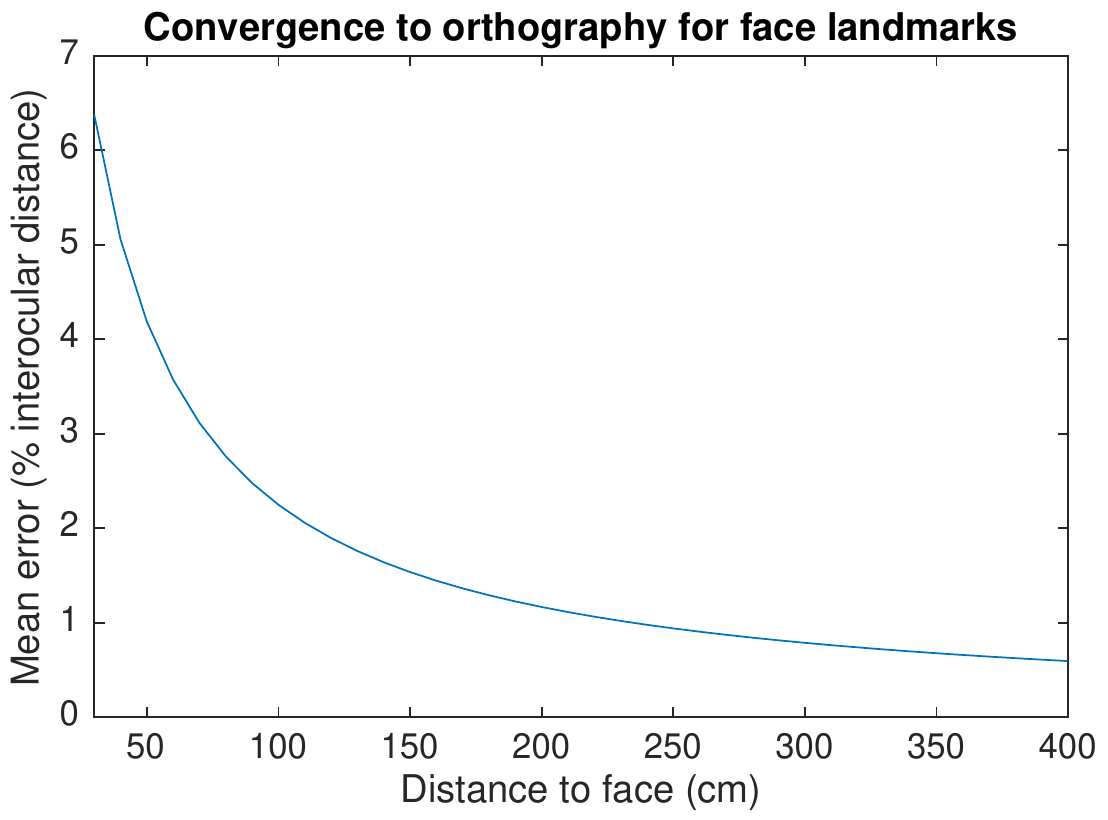}}
\subfloat[\label{fig:distance}]{\includegraphics[height=3.5cm, clip=true,trim=160px 261px 175px 262px]{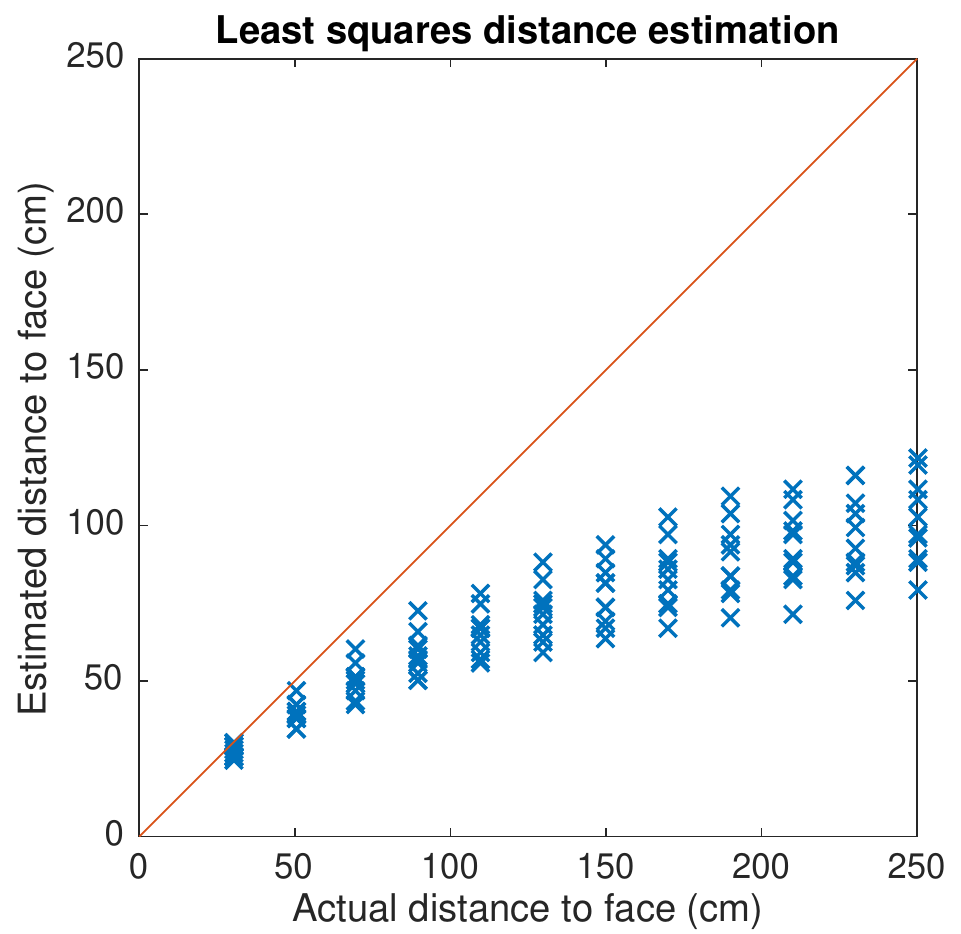}}
\caption{(a) Mean landmark error ($y$ axis) between perspective and orthographic projection, averaged over 10 BFM scans, as subject-camera distance ($x$ axis) is varied. (b) Subject-camera distance estimation by least squares optimisation.}
\vspace{-0.5mm}
\end{figure}

\subsection{SNLS fitting}

In Section \ref{section:snls} we introduced a novel formulation of 3DMM fitting under orthographic and perspective projection using SNLS. Although our goal in this paper is to investigate ambiguities in the 3D interpretation of 2D geometry and not to advance the state of the art in 3DMM fitting, we nevertheless begin by demonstrating that our SNLS formulation is indeed superior to alternating least squares (ALS) as used in previous work \citep{Bas:16,Zhu:15,OSPAMI,cao20133d,cao2014displaced,saito2016real}. In order to evaluate in a realistic setting, we require images with corresponding ground truth 3DMM fits. For this reason, we use the Facewarehouse dataset and model \citep{cao2014facewarehouse}. We use leave-one-out testing, building each model on 149 subjects and testing on the remaining one and use the 74 landmarks provided with the dataset. For this evaluation we test only the orthographic setting. Fig.~\ref{fig:3a} shows the mean Euclidean distance between dense ground truth and estimated face surface in mm after Procrustes alignment. We do not use any regularisation for either algorithm and therefore do not need to choose the weight parameter. For all subjects SNLS achieves a lower error, on average reducing it by about 30\%.

\setlength{\tabcolsep}{2.1pt}
\renewcommand{\arraystretch}{1}
\begin{table}[!t]
\centering
\noindent\resizebox{.48\textwidth}{!}{
\begin{tabular}{|l|c|c|c|c|c|c|}\hline
& \multicolumn{5}{c|}{{\bf Rotation angle}} &  \\ \cline{2-6}
{\bf Method} & $-30^{\circ}$ & $-15^{\circ}$ & $0^{\circ}$ & $15^{\circ}$ & $30^{\circ}$ & {\bf Mean} \\
\hline \hline
\cite{Zhu:15} &4.63&5.09&4.19&5.22&4.92&4.81 \\ \hline
SLNS (Ours) &{\bf 4.53}&{\bf 4.29}&{\bf 4.16}&{\bf 3.99}&{\bf 4.07}&{\bf 4.21} \\ \hline
\end{tabular}}
%}
\caption{Quantitative  comparison  between \cite{Zhu:15}
and SNLS on synthetic data with automatically detected landmarks.
Each cell shows the mean euclidean vertex distance for related pose in mm.
}
\label{tab:hpenvssnls}
\end{table}

As a second experiment, we provide a quantitative fitting comparison on synthetic face images in various poses (rotations of $0^{\circ}$, $\pm 15^{\circ}$ and $\pm 30^{\circ}$ about the vertical axis) which are rendered in orthographic projection from the out-of-sample faces supplied by the BFM.
We use the algorithm of \cite{Zhu:15} with the landmarks detected by the automatic method of \cite{zhu2012face}.
The fitting method and the landmark detector are both publicly available.
Table \ref{tab:hpenvssnls} reports the mean Euclidean distance between ground truth and estimated face surface in mm after Procrustes alignment. 
This shows that our SLNS optimisation provides better overall performance and superior results for all poses.

\subsection{Perspective ambiguity}

We begin by investigating the perspective ambiguity using synthetic data. We use the out-of-sample BFM scans to create input data by choosing pose parameters and projecting the faces to 2D. For sparse landmarks, we use the 70 anthropometric landmarks (due to \citep{Farkas:94}) whose indices in the BFM are known. These landmarks are particularly appropriate as they were chosen so as to best measure the variability in craniofacial shape over a population. In Fig.~\ref{fig:convergence}, we show over what range of distances perspective transformation has a significant effect on 2D face geometry. For each face, we project the 70 landmarks to 2D under perspective projection and measure $d_L$ with respect to the orthographic projection of the landmarks. As $t_z$ increases, the projection converges towards orthography and the error tends to zero. The landmark error falls below 1\% when the distance is around 2.5 metres. Hence, we experiment with distances ranging from selfie distance (30cm) up to this distance.

\setlength{\tabcolsep}{2.1pt}
\renewcommand{\arraystretch}{1}
\begin{table}[!t]
    \centering
    \noindent\resizebox{.4\textwidth}{!}{
    \begin{tabular}{|c||c|c|c|c|c|}%|c|c|c|c|c|}
    \hline
    %Actual & \multicolumn{5}{c||}{{\bf Frontal pose}} \\% & \multicolumn{5}{c|}{{\bf ${\bf 45^{\circ}}$ rotated pose}} \\ \cline{2-11}
    %\cline{2-6}
    Actual  & \multicolumn{5}{c|}{Fitting distance (cm)} \\ % & \multicolumn{5}{c|}{Fitting distance (cm)} \\ \cline{2-11}
        \cline{2-6}
distance (cm) & \textbf{30}    &  \textbf{60}    &  \textbf{120}  & \textbf{240}   & \textbf{Ortho}  \\
    \hline
    \hline
    \multirow{2}{*}{\textbf{30}} &     0.21  &  0.24 &   0.26 &   0.27 &   0.28       \\
    & 7.23  &  9.70  & 13.07 &  14.55 &  14.47   \\
    \hline
    \multirow{2}{*}{\textbf{60}} & 0.30  &  0.26  &  0.27  &  0.27 &   0.28 \\
    & 8.07  &  6.29  &  6.60 &    6.99  &  7.48 \\
    \hline
    \multirow{2}{*}{\textbf{120}} & 0.37  &  0.29 &   0.28  &  0.28  &  0.28 \\
    & 9.52  &  6.17  &  5.38    & 5.39  &  5.62 \\
    \hline
    \multirow{2}{*}{\textbf{240}} & 0.42  &  0.32  &  0.29  &  0.29  &  0.28 \\
    & 10.16  &  6.72 &    5.59 &   5.37 &   5.38 \\
    \hline
    \multirow{2}{*}{\textbf{Ortho}} & 0.47  &  0.35  &  0.31  &  0.30  &  0.29 \\
    & 11.02  &  7.43 &    6.01  &  5.54  &  5.29 \\
    \hline
    \end{tabular}}
    \caption{Quantitative results for the perspective ambiguity on synthetic data. Each cell shows the landmark error, $d_L$ in \%, top and surface error, $d_S$ in mm, bottom.}
    \label{tab:PerspAmbig}
\end{table}

\begin{figure*}[!t]
 \centering
\noindent\resizebox{\textwidth}{!}{
\begin{tabular}{ccccccc}
\hline
\multirow{2}{*}{\bfseries Target} & 
\multicolumn{5}{c}{\bfseries Subject-camera distance (cm)}&
\multirow{2}{*}{\bfseries Error} \\ \cline{2-6}
& $t_z$=30 & $t_z$=60 & $t_z$=90 & $t_z$=120 & $t_z$=240& \\ \hline\\[-0.5em]
%------
\includegraphics[height=4cm, clip=true,trim=20px 5px 25px 35px]{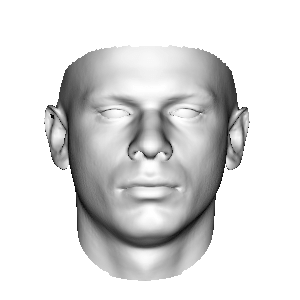}&
\includegraphics[height=4cm, clip=true,trim=20px 5px 25px 35px]{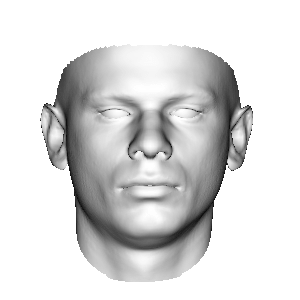}&
\includegraphics[height=4cm, clip=true,trim=20px 5px 25px 35px]{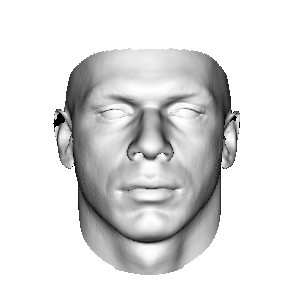}&
\includegraphics[height=4cm, clip=true,trim=20px 5px 25px 35px]{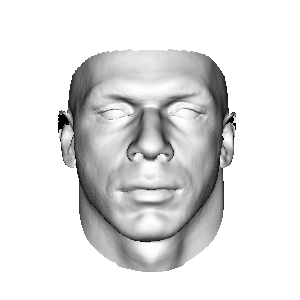}&
\includegraphics[height=4cm, clip=true,trim=20px 5px 25px 35px]{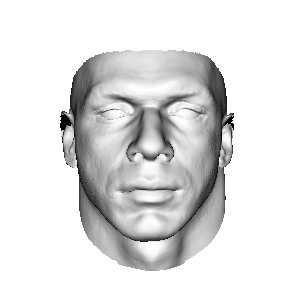}&
\includegraphics[height=4cm, clip=true,trim=20px 5px 25px 35px]{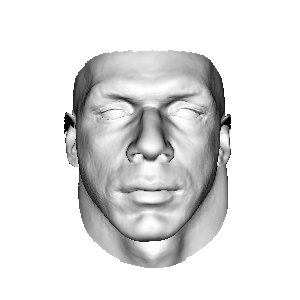}&
\includegraphics[height=4cm, clip=true]{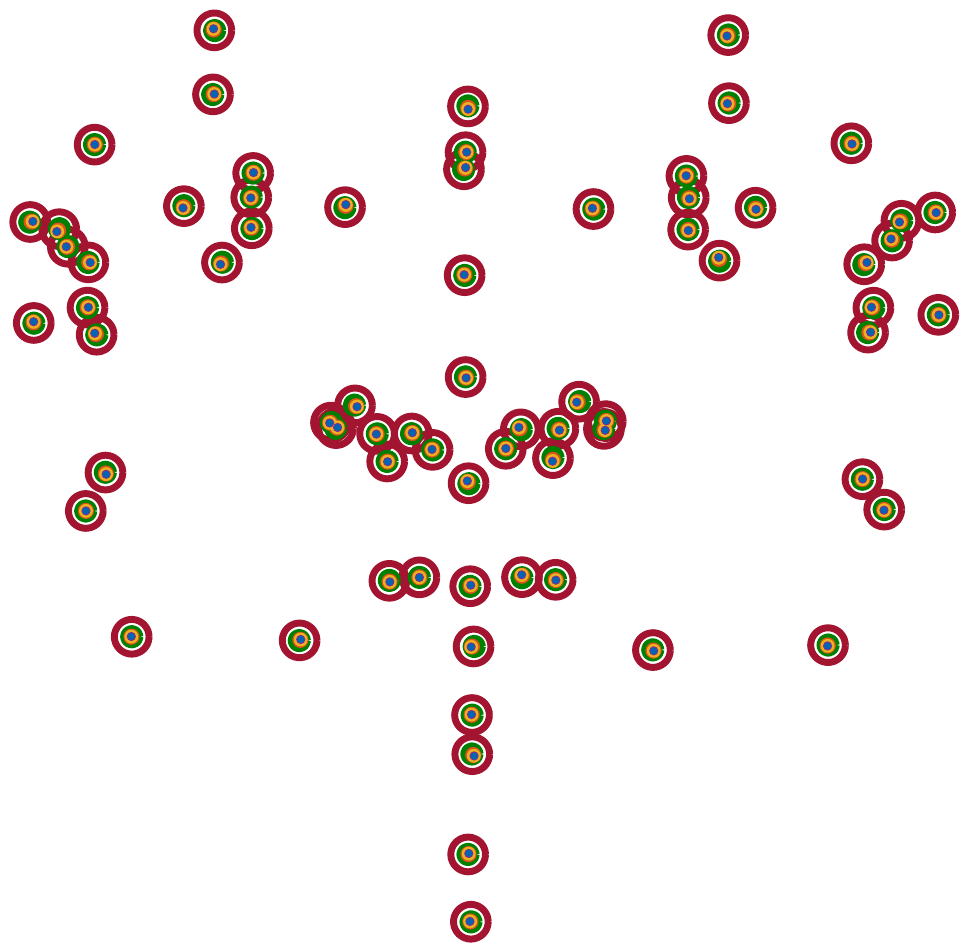}\\
\includegraphics[height=4cm, clip=true,trim=20px 5px 25px 35px]{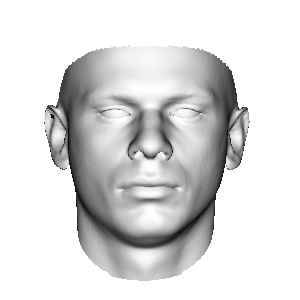}&
\includegraphics[height=4cm, clip=true,trim=20px 5px 25px 35px]{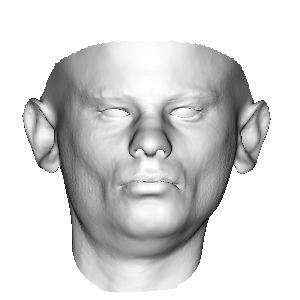}&
\includegraphics[height=4cm, clip=true,trim=20px 5px 25px 35px]{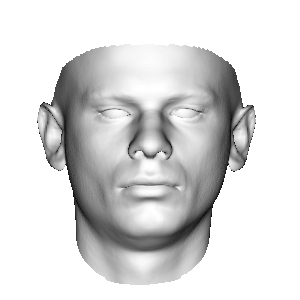}&
\includegraphics[height=4cm, clip=true,trim=20px 5px 25px 35px]{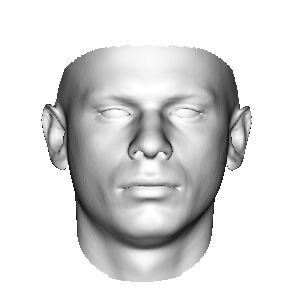}&
\includegraphics[height=4cm, clip=true,trim=20px 5px 25px 35px]{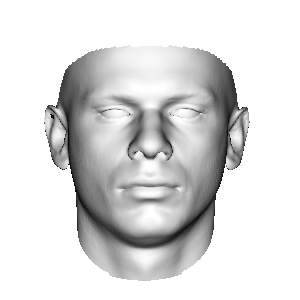}&
\includegraphics[height=4cm, clip=true,trim=20px 5px 25px 35px]{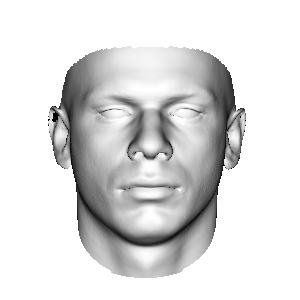}&
\includegraphics[height=4cm, clip=true]{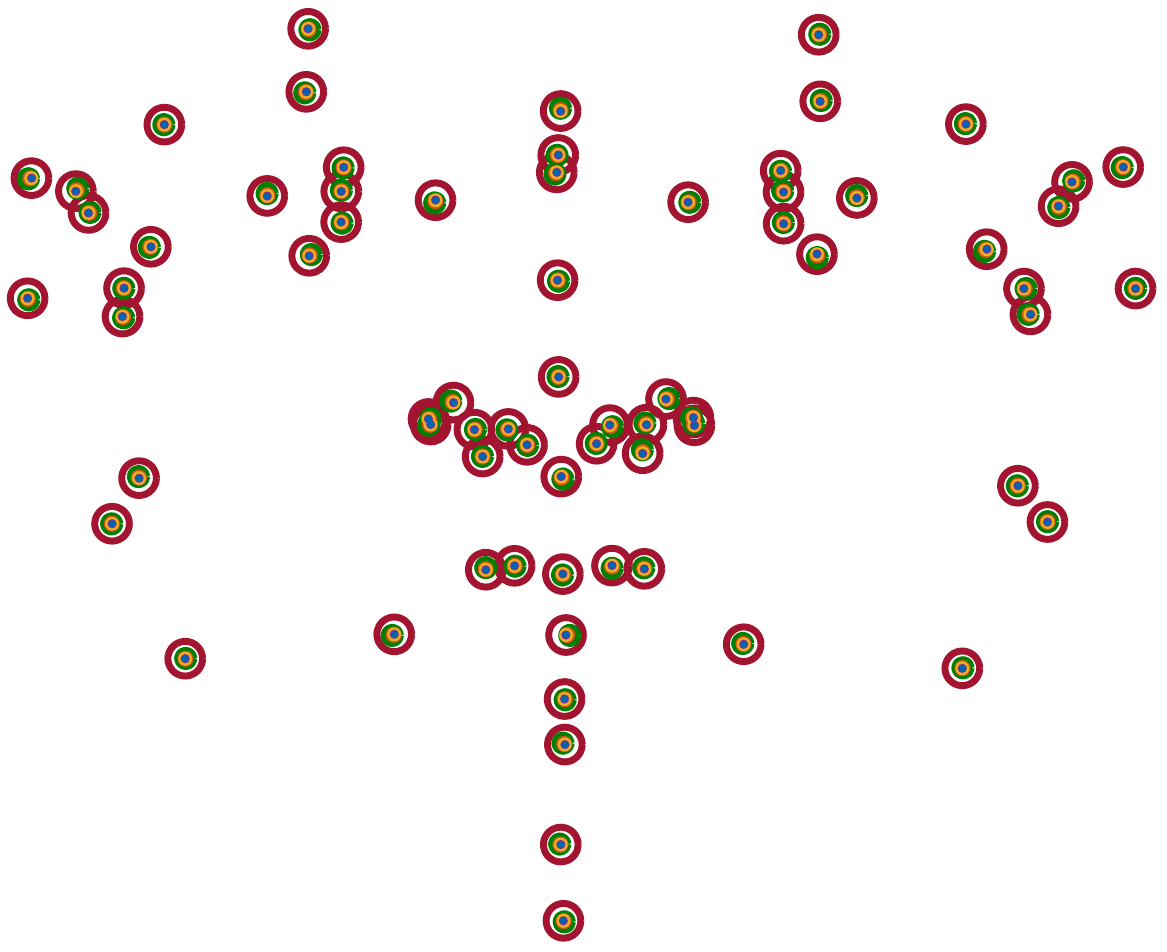}\\
\end{tabular}
}
\caption{Qualitative perspective face shape ambiguity. There is a subspace of possible 3D face shapes with varying subject-camera distance within the landmark tolerance. Target face is at 30cm (first row) and 120cm (second row).}
\label{fig:perspective_fw}
\end{figure*}

%\begin{figure}[!t]
%\centering
%\includegraphics[height=5cm, clip=true,trim=160px 261px 175px 262px]{fig/distance.pdf}
%\caption{Subject-camera distance estimation by least squares optimisation.}
%\label{fig:distance}
%\end{figure}

Our first evaluation of the perspective ambiguity is based on estimating the subject-camera distance as one of the parameters in the least squares fitting process. We use the out-of-sample BFM scans as target faces, vary the subject-camera distance and project the 70 Farkas landmarks to 2D under perspective projection. We use a frontal pose (${\bf r}=[0\ 0\ 0]$) and arbitrarily set the focal length to $f=1$. We initialise the optimisation with the correct focal length and rotation, giving it the best possible chance of estimating the correct distance.  We plot estimated versus ground truth distance in Fig.~\ref{fig:distance}. Optimal performance would see all points falling on the diagonal red line. The distance is consistently under-estimated and the mean percentage error in the estimate is $42\%$. It is clear that the 2D landmarks alone do not contain enough information to accurately estimate subject-camera distance as part of the model fitting process.

We now show that landmarks produced by a real 3D face shape at one distance can be explained by 3D shapes at multiple different distances. In Table \ref{tab:PerspAmbig} we show quantitative results. Each row of the table corresponds to a distance at which we place each of the BFM scans in a frontal pose before projecting to 2D. We then fit to these landmarks with the subject-camera distance assumed to be the value shown in the column. The results show that we are able to explain the data almost as well at the wrong distance as the correct one but the 3D shape is very different, differing by over a 1cm on average. Note that \cite{Burgos:14} found that the difference between landmarks on the same face placed by two different humans was typically 3\% of the interocular distance. Similarly, the 300 faces in the wild challenge \citep{sagonas2016300} found that even the best methods did not obtain better than 5\% accuracy for more than 50\% of the landmarks.
Hence, the difference between target and fitted landmarks is substantially smaller than the accuracy of either human or machine placed landmarks. Importantly, this means that the fitting energy could not be used to resolve the ambiguity. The residual difference between target and fitted landmarks is too small to meaningfully choose between the two solutions.

We now show qualitative examples from the same experiment. 
In Fig.~\ref{fig:perspective_fw} we show orthographic renderings of perspective fits to the face shown in the first column. In the first row, the target landmarks were generated by viewing the face at 30cm, in the second row the face was at 120cm. In each column we show fitting results at different distances. In the final column we show the landmarks of the real face (circles) overlaid with the landmarks from the fitted faces (dots) showing that highly varying 3D faces can produce almost identical 2D landmarks.

%SynthPersComp
\newcommand{\frontimsize}{1.54cm}
\setlength{\tabcolsep}{1pt}
\begin{figure*}[!t]
\centering
\noindent\resizebox{\textwidth}{!}{
\begin{tabular}{cccccccccccc}

& Original & Landmark & Edge & Dense & Original & Landmark & Edge & Dense & Landmark & Edge & Dense \\\cline{2-12}% \hline\hline

\tiny{\(d_S/d_L\)}&&\tiny{5.33mm/0.38\%} &\tiny{5.90mm/0.29\%} &\tiny{6.13mm/0.39\%} &\\
\rotatebox{90}{\scriptsize Perspective}&
\includegraphics[height=\frontimsize, clip=true,trim=75px 10px 80px 35px]{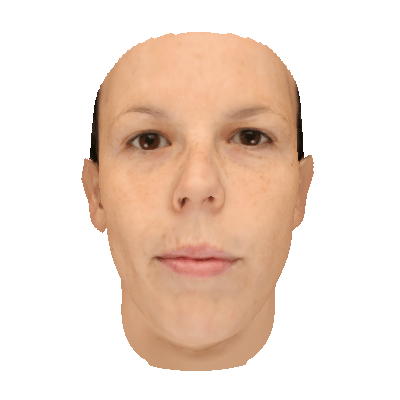}&
\includegraphics[height=\frontimsize, clip=true,trim=75px 10px 80px 35px]{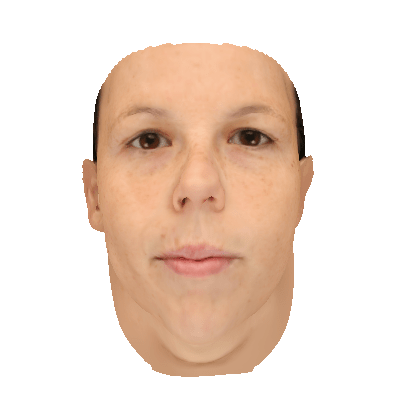}&
\includegraphics[height=\frontimsize, clip=true,trim=75px 10px 80px 35px]{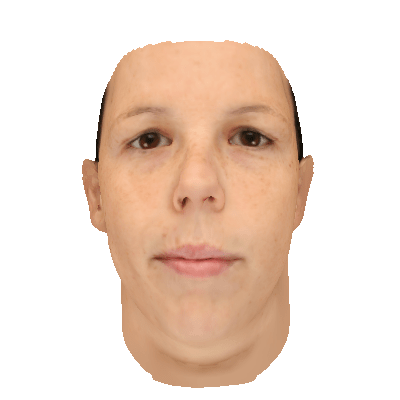}&
\includegraphics[height=\frontimsize, clip=true,trim=75px 10px 80px 35px]{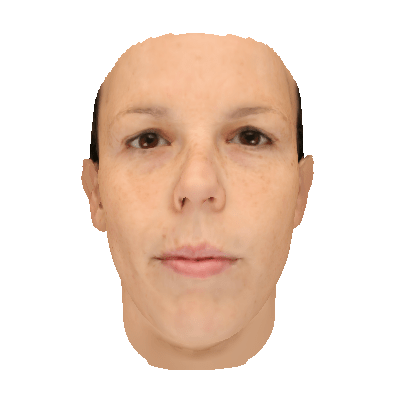}&
\includegraphics[height=\frontimsize, clip=true,trim=75px 10px 80px 35px]{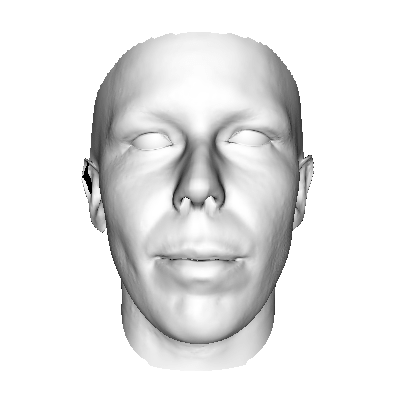}&
\includegraphics[height=\frontimsize, clip=true,trim=75px 10px 80px 35px]{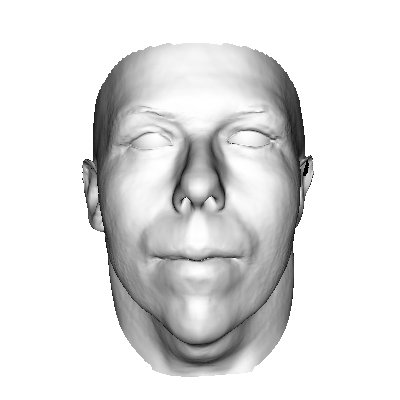}&
\includegraphics[height=\frontimsize, clip=true,trim=75px 10px 80px 35px]{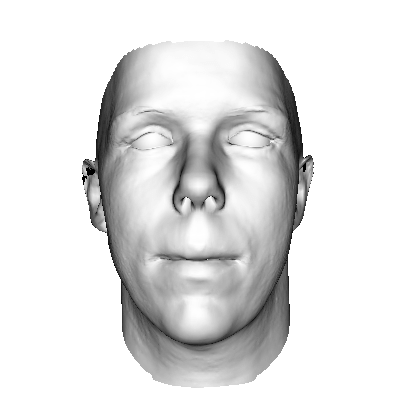}&
\includegraphics[height=\frontimsize, clip=true,trim=75px 10px 80px 35px]{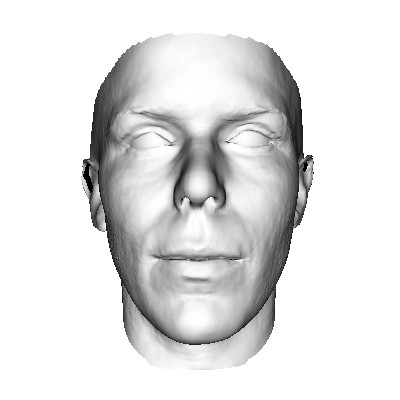}&
\includegraphics[height=\frontimsize, clip=true,trim=75px 10px 80px 35px]{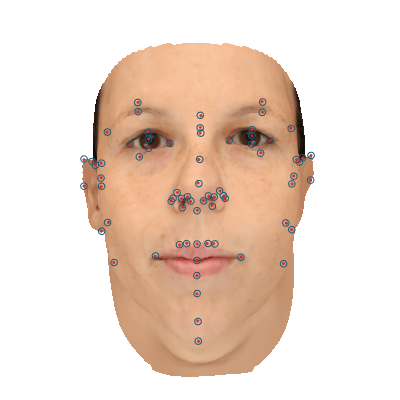}&
\includegraphics[height=\frontimsize, clip=true,trim=75px 10px 80px 35px]{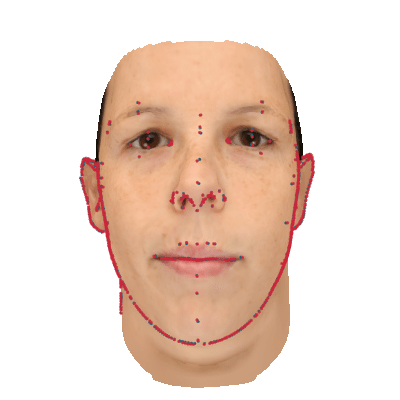}&
\includegraphics[height=\frontimsize, clip=true,trim=75px 10px 80px 35px]{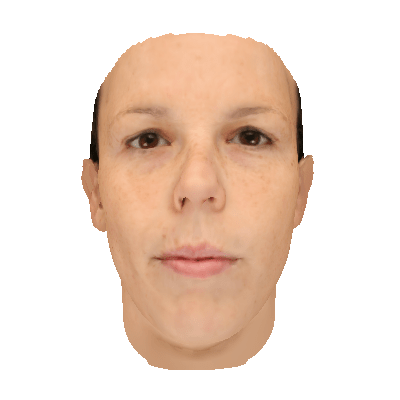}\\
\rotatebox{90}{\scriptsize Orthographic}&
\includegraphics[height=\frontimsize, clip=true,trim=45px 0px 45px 40px]{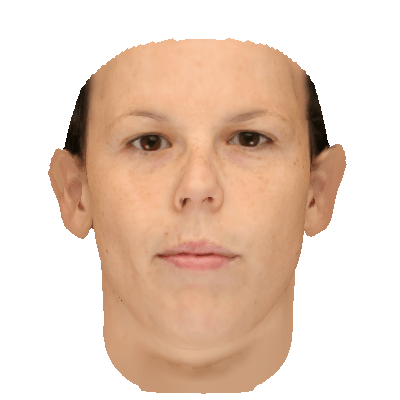}&
\includegraphics[height=\frontimsize, clip=true,trim=45px 0px 45px 40px]{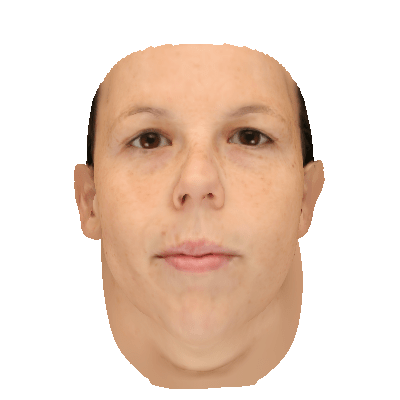}&
\includegraphics[height=\frontimsize, clip=true,trim=45px 0px 45px 40px]{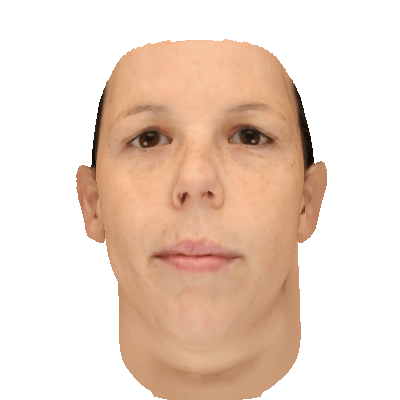}&
\includegraphics[height=\frontimsize, clip=true,trim=45px 0px 45px 40px]{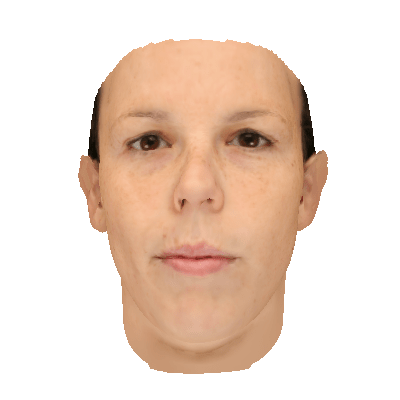}&
\includegraphics[height=\frontimsize, clip=true,trim=45px 0px 45px 40px]{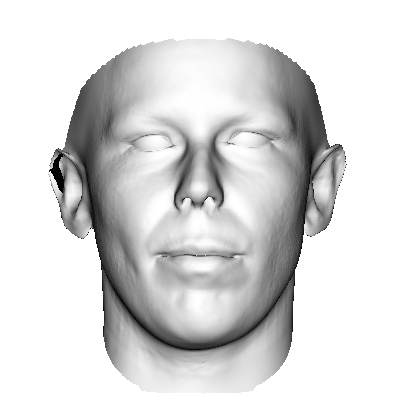}&
\includegraphics[height=\frontimsize, clip=true,trim=45px 0px 45px 40px]{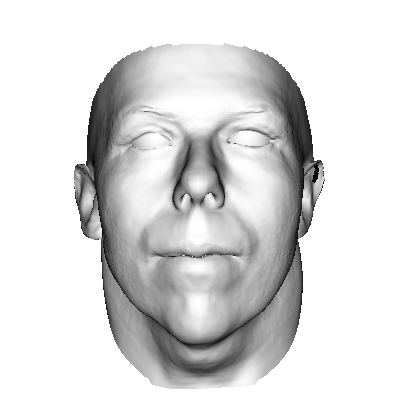}&
\includegraphics[height=\frontimsize, clip=true,trim=45px 0px 45px 40px]{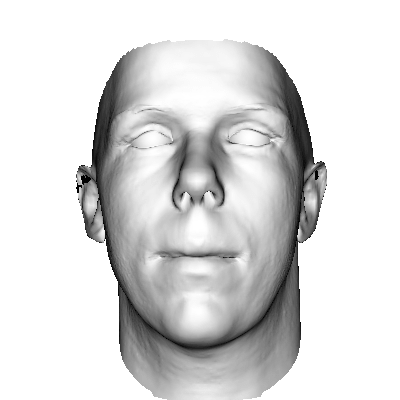}&
\includegraphics[height=\frontimsize, clip=true,trim=45px 0px 45px 40px]{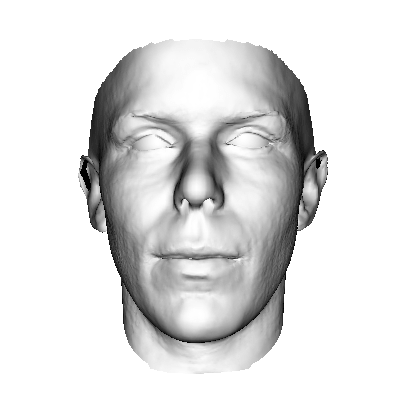}&&&\\
\tiny{\(d_S/d_L\)}&&\tiny{7.12mm/0.82\%} &\tiny{3.88mm/0.63\%} &\tiny{3.51mm/0.68\%} &\\
\rotatebox{90}{\scriptsize Perspective}&
\includegraphics[height=\frontimsize, clip=true,trim=95px 50px 40px 55px]{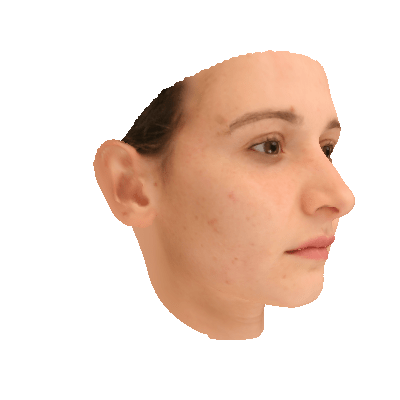}&
\includegraphics[height=\frontimsize, clip=true,trim=95px 50px 40px 55px]{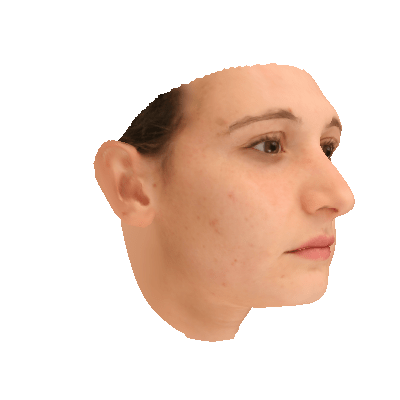}&
\includegraphics[height=\frontimsize, clip=true,trim=95px 50px 40px 55px]{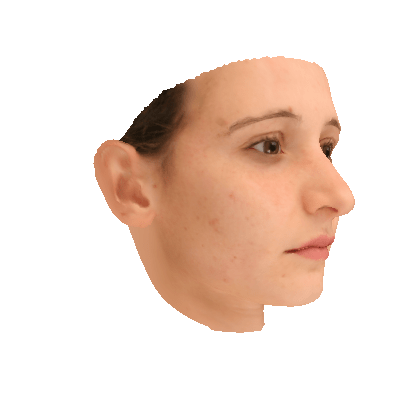}&
\includegraphics[height=\frontimsize, clip=true,trim=95px 50px 40px 55px]{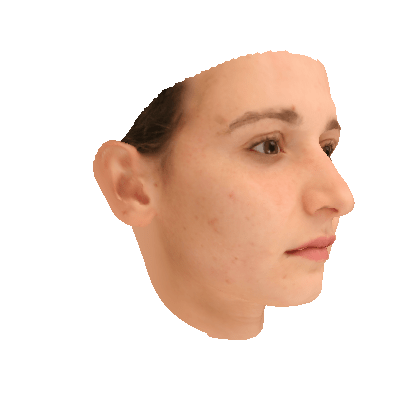}&
\includegraphics[height=\frontimsize, clip=true,trim=95px 50px 40px 55px]{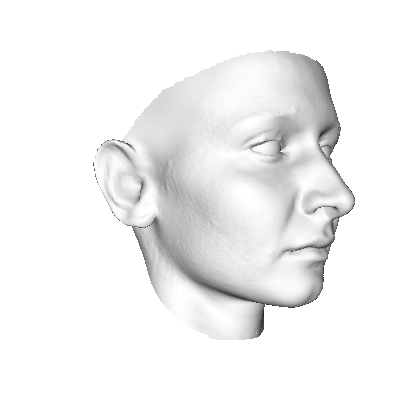}&
\includegraphics[height=\frontimsize, clip=true,trim=95px 50px 40px 55px]{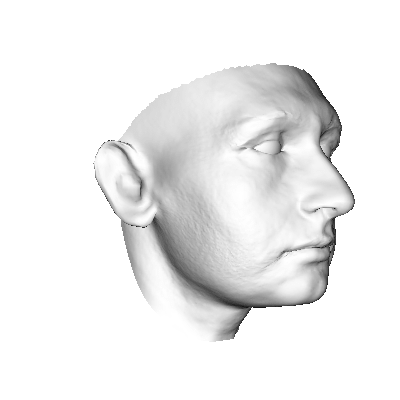}&
\includegraphics[height=\frontimsize, clip=true,trim=95px 50px 40px 55px]{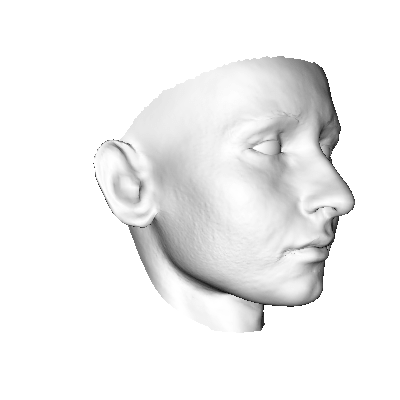}&
\includegraphics[height=\frontimsize, clip=true,trim=95px 50px 40px 55px]{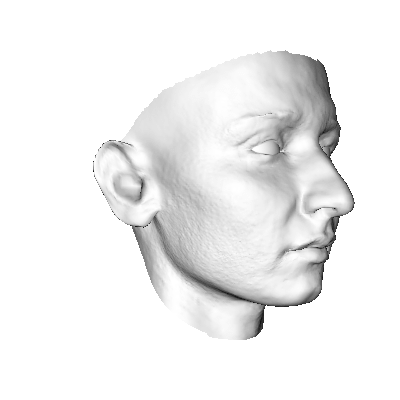}&
\includegraphics[height=\frontimsize, clip=true,trim=95px 50px 40px 55px]{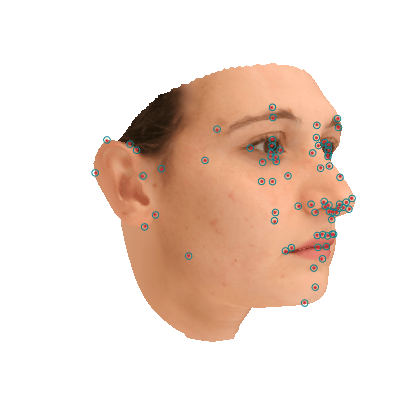}&
\includegraphics[height=\frontimsize, clip=true,trim=95px 50px 40px 55px]{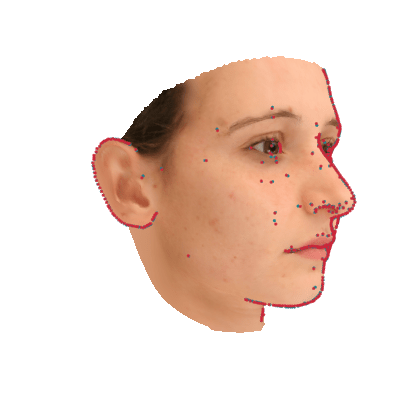}&
\includegraphics[height=\frontimsize, clip=true,trim=95px 50px 40px 55px]{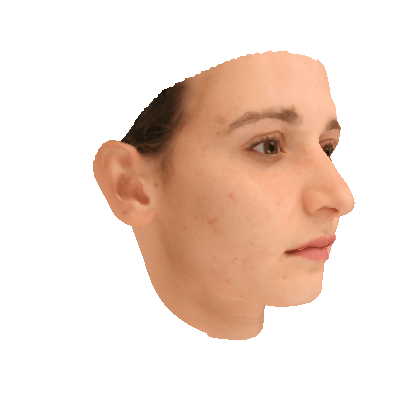}\\
\rotatebox{90}{\scriptsize Orthographic}&
\includegraphics[height=\frontimsize, clip=true,trim=40px 30px 45px 50px]{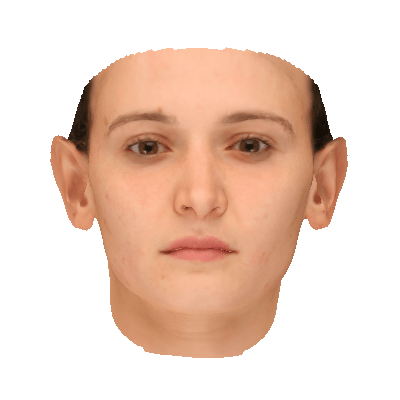}&
\includegraphics[height=\frontimsize, clip=true,trim=40px 30px 45px 50px]{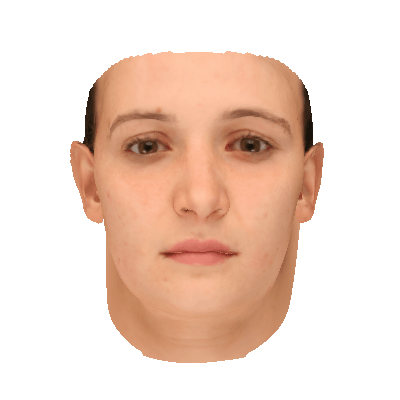}&
\includegraphics[height=\frontimsize, clip=true,trim=40px 30px 45px 50px]{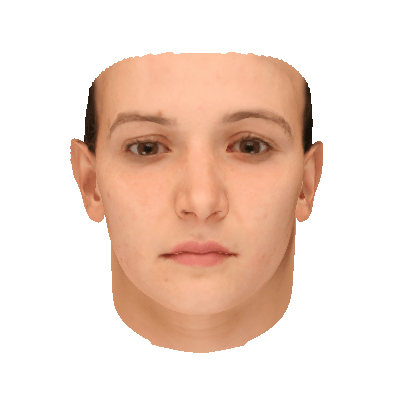}&
\includegraphics[height=\frontimsize, clip=true,trim=40px 30px 45px 50px]{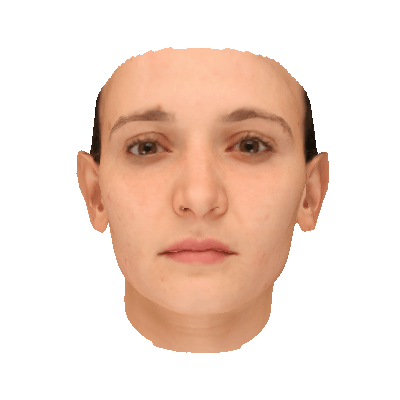}&
\includegraphics[height=\frontimsize, clip=true,trim=40px 30px 45px 50px]{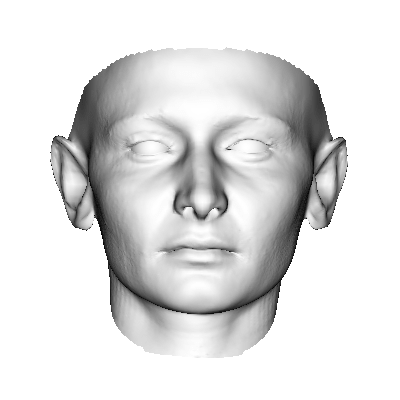}&
\includegraphics[height=\frontimsize, clip=true,trim=40px 30px 45px 50px]{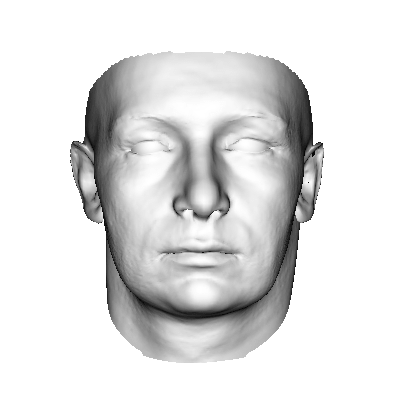}&
\includegraphics[height=\frontimsize, clip=true,trim=40px 30px 45px 50px]{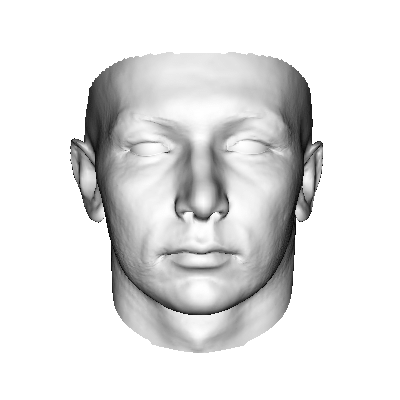}&
\includegraphics[height=\frontimsize, clip=true,trim=40px 30px 45px 50px]{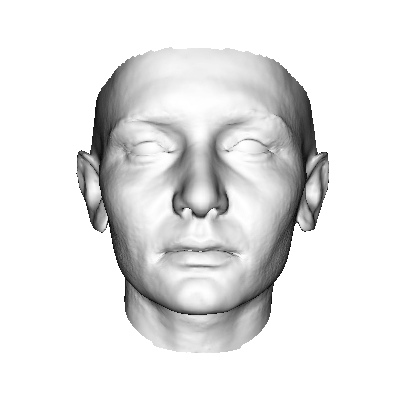}&&&
\end{tabular}
}
\caption{Sparse and dense fitting of the synthetic images. Target at 30cm, fitted results at 120cm.}
\label{fig:SynthPersComp}
\end{figure*}
%Frontal
%Face ID: 1. Level (1:idx 2:edge 3:all):1 | ndims: 70. Landmark distance : 0.378%, surface distance : 5.330 mm.
%Face ID: 1. Level (1:idx 2:edge 3:all):2 | ndims: 70. Landmark distance : 0.294%, surface distance : 5.895 mm.
%Face ID: 1. Level (1:idx 2:edge 3:all):3 | ndims: 199. Landmark distance : 0.389%, surface distance : 6.131 mm.
%Rotated
%Face ID: 8. Level (1:idx 2:edge 3:all):1 | ndims: 70. Landmark distance : 0.823%, surface distance : 7.117 mm.
%Face ID: 8. Level (1:idx 2:edge 3:all):2 | ndims: 70. Landmark distance : 0.627%, surface distance : 3.880 mm.
%Face ID: 8. Level (1:idx 2:edge 3:all):3 | ndims: 199. Landmark distance : 0.684%, surface distance : 3.512 mm.

%SynthPersComp
\setlength{\tabcolsep}{1pt}
\begin{figure*}[!t]
\centering
\noindent\resizebox{\textwidth}{!}{
\begin{tabular}{cccccccccccc}
& Original & Landmark & Edge & Dense & Original & Landmark & Edge & Dense & Landmark & Edge & Dense \\\cline{2-12}%\hline\hline

\tiny{\(d_S/d_L\)}&&\tiny{5.13mm/0.39\%} &\tiny{5.66mm/0.38\%} &\tiny{2.44mm/0.36\%} &\\
\rotatebox{90}{\scriptsize Perspective}&
\includegraphics[height=\frontimsize, clip=true,trim=55px 5px 50px 25px]{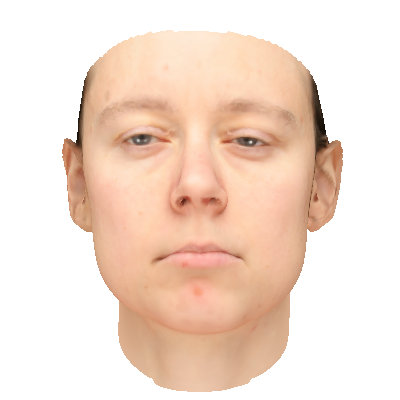}&
\includegraphics[height=\frontimsize, clip=true,trim=55px 5px 50px 25px]{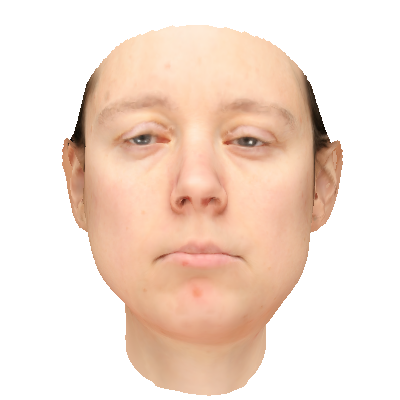}&
\includegraphics[height=\frontimsize, clip=true,trim=55px 5px 50px 25px]{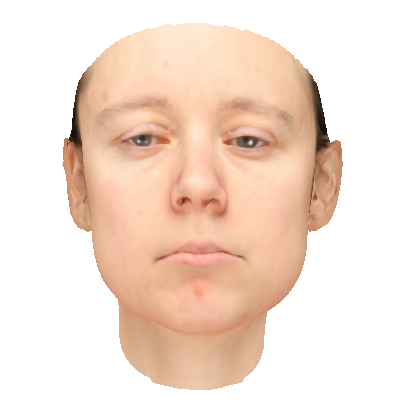}&
\includegraphics[height=\frontimsize, clip=true,trim=55px 5px 50px 25px]{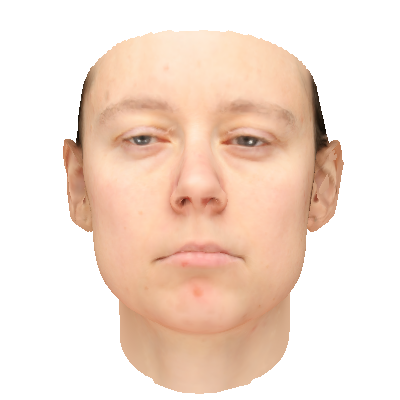}&
\includegraphics[height=\frontimsize, clip=true,trim=55px 5px 50px 25px]{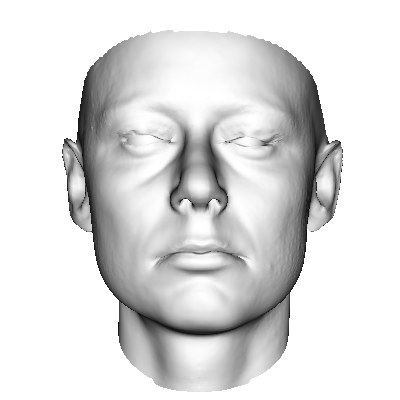}&
\includegraphics[height=\frontimsize, clip=true,trim=55px 5px 50px 25px]{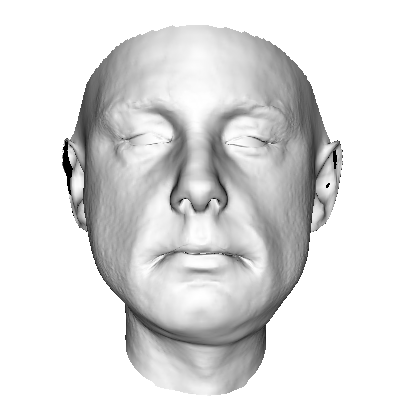}&
\includegraphics[height=\frontimsize, clip=true,trim=55px 5px 50px 25px]{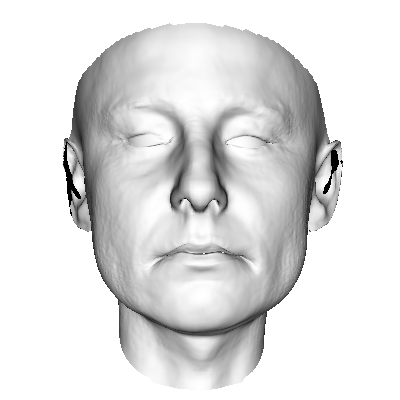}&
\includegraphics[height=\frontimsize, clip=true,trim=55px 5px 50px 25px]{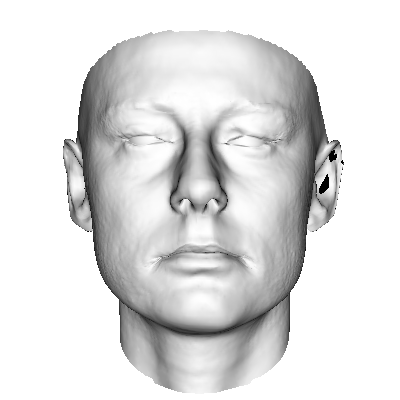}&
\includegraphics[height=\frontimsize, clip=true,trim=55px 5px 50px 25px]{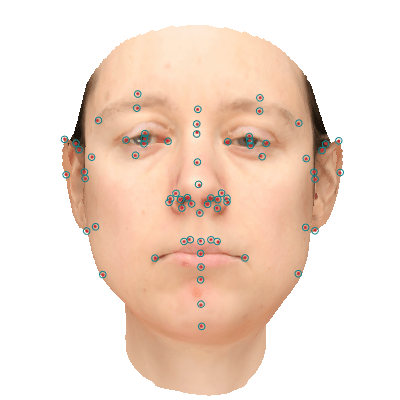}&
\includegraphics[height=\frontimsize, clip=true,trim=55px 5px 50px 25px]{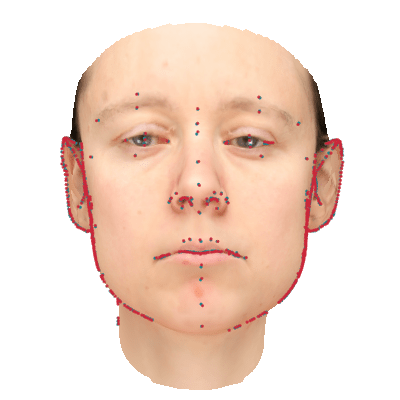}&
\includegraphics[height=\frontimsize, clip=true,trim=55px 5px 50px 25px]{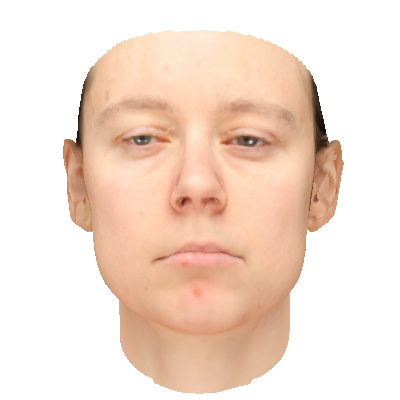}\\
\rotatebox{90}{\scriptsize Orthographic}&
\includegraphics[height=\frontimsize, clip=true,trim=40px 0px 20px 15px]{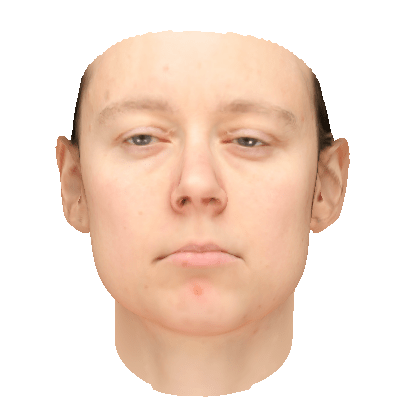}&
\includegraphics[height=\frontimsize, clip=true,trim=40px 0px 20px 15px]{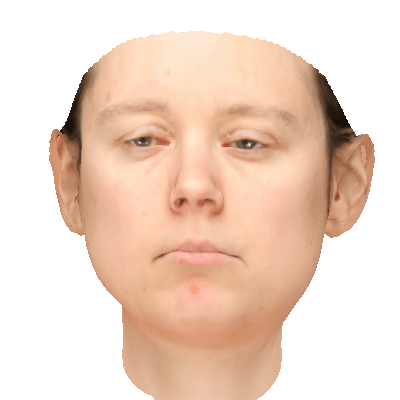}&
\includegraphics[height=\frontimsize, clip=true,trim=40px 0px 20px 15px]{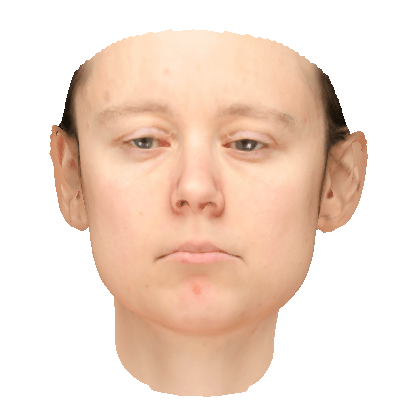}&
\includegraphics[height=\frontimsize, clip=true,trim=40px 0px 20px 15px]{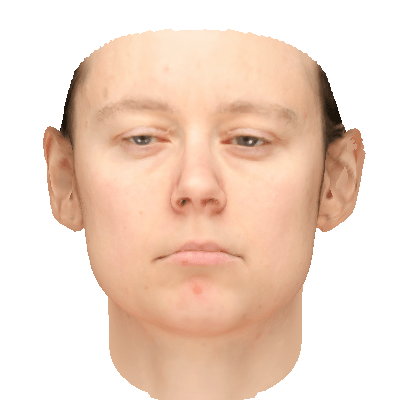}&
\includegraphics[height=\frontimsize, clip=true,trim=40px 0px 20px 15px]{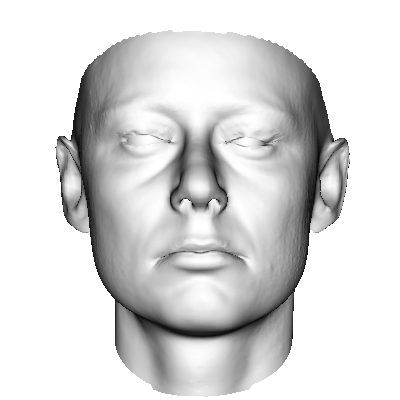}&
\includegraphics[height=\frontimsize, clip=true,trim=40px 0px 20px 15px]{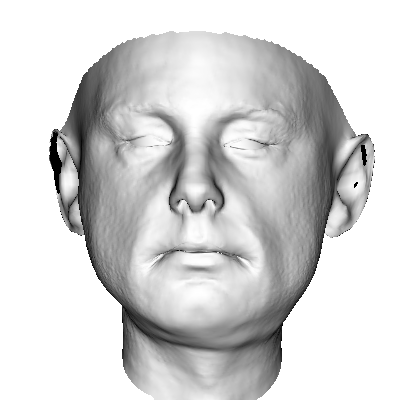}&
\includegraphics[height=\frontimsize, clip=true,trim=40px 0px 20px 15px]{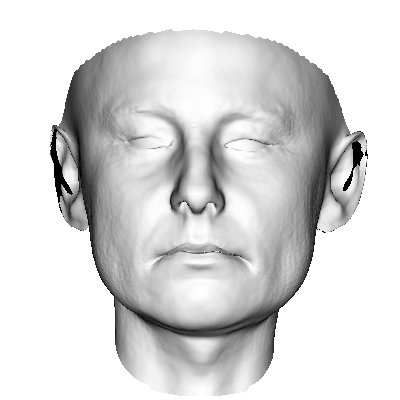}&
\includegraphics[height=\frontimsize, clip=true,trim=40px 0px 20px 15px]{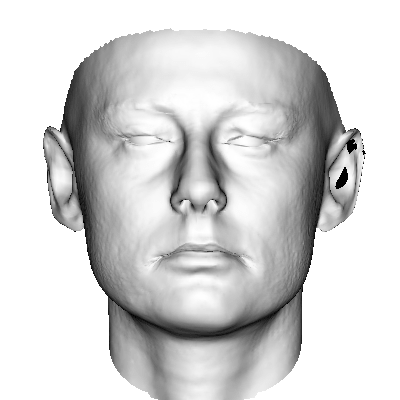}&&&\\
\tiny{\(d_S/d_L\)}&&\tiny{7.15mm/0.51\%} &\tiny{4.70mm/0.52\%} &\tiny{3.63mm/0.49\%} &\\
\rotatebox{90}{\scriptsize Perspective}&
\includegraphics[height=\frontimsize, clip=true,trim=85px 5px 25px 35px]{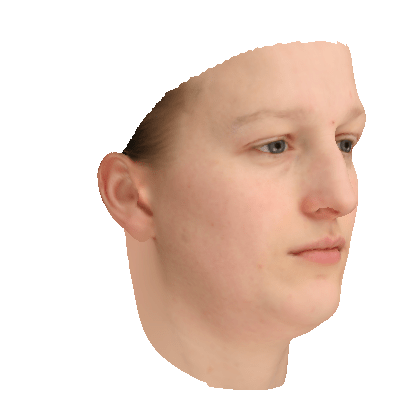}&
\includegraphics[height=\frontimsize, clip=true,trim=85px 5px 25px 35px]{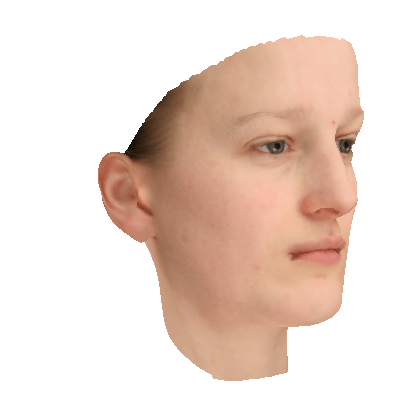}&
\includegraphics[height=\frontimsize, clip=true,trim=85px 5px 25px 35px]{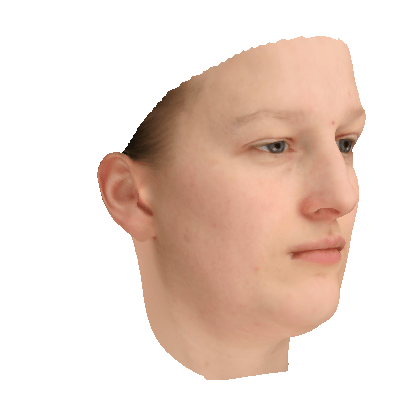}&
\includegraphics[height=\frontimsize, clip=true,trim=85px 5px 25px 35px]{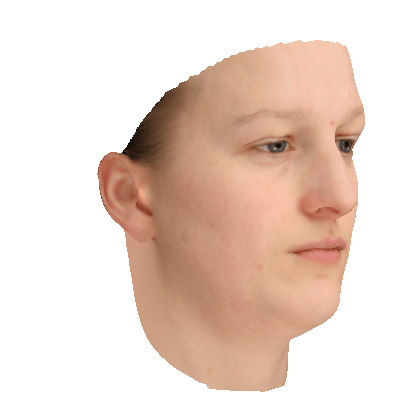}&
\includegraphics[height=\frontimsize, clip=true,trim=85px 5px 25px 35px]{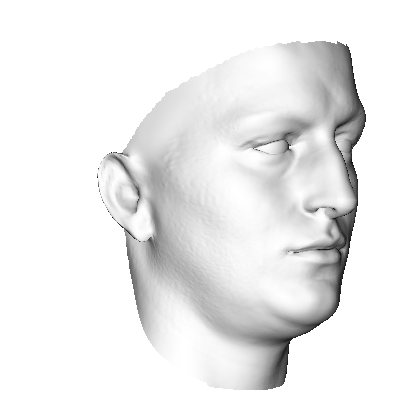}&
\includegraphics[height=\frontimsize, clip=true,trim=85px 5px 25px 35px]{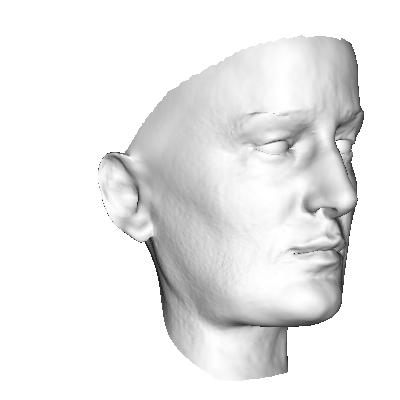}&
\includegraphics[height=\frontimsize, clip=true,trim=85px 5px 25px 35px]{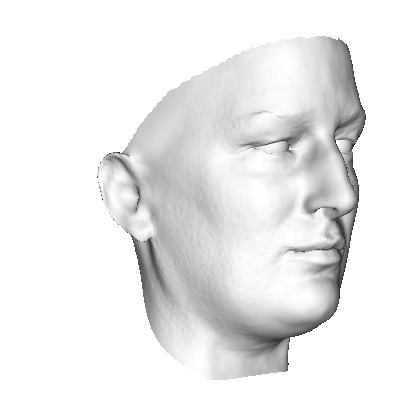}&
\includegraphics[height=\frontimsize, clip=true,trim=85px 5px 25px 35px]{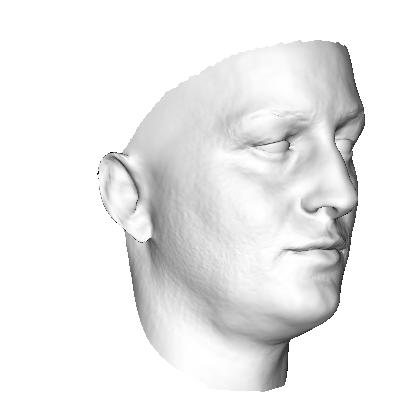}&
\includegraphics[height=\frontimsize, clip=true,trim=85px 5px 25px 35px]{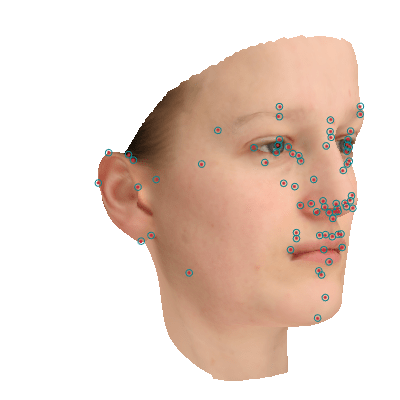}&
\includegraphics[height=\frontimsize, clip=true,trim=85px 5px 25px 35px]{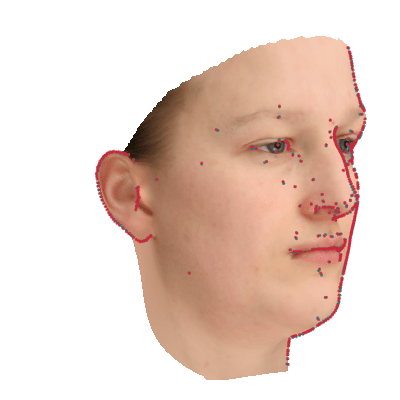}&
\includegraphics[height=\frontimsize, clip=true,trim=85px 5px 25px 35px]{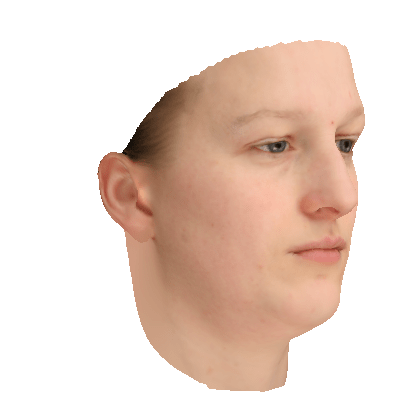}\\
\rotatebox{90}{\scriptsize Orthographic}&
\includegraphics[height=\frontimsize, clip=true, trim=40px 0px 35px 25px]{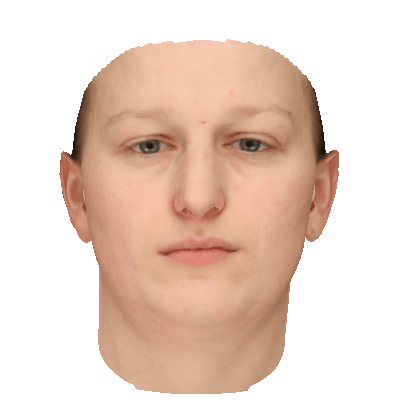}&
\includegraphics[height=\frontimsize, clip=true, trim=40px 0px 35px 25px]{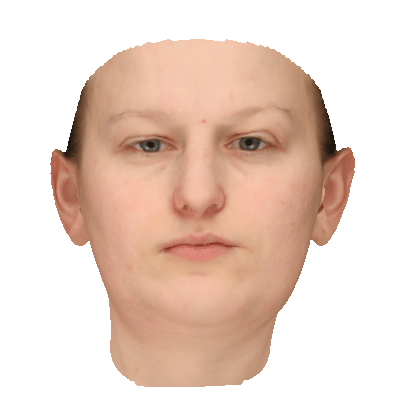}&
\includegraphics[height=\frontimsize, clip=true, trim=40px 0px 35px 25px]{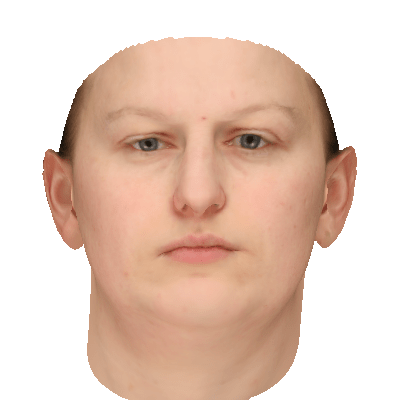}&
\includegraphics[height=\frontimsize, clip=true, trim=40px 0px 35px 25px]{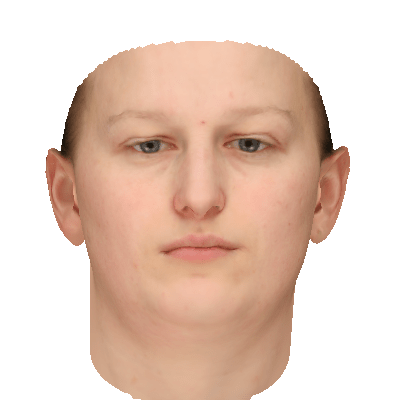}&
\includegraphics[height=\frontimsize, clip=true, trim=40px 0px 35px 25px]{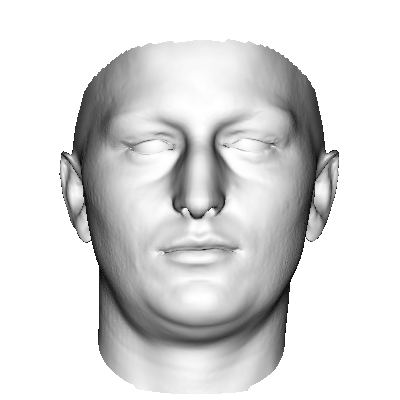}&
\includegraphics[height=\frontimsize, clip=true, trim=40px 0px 35px 25px]{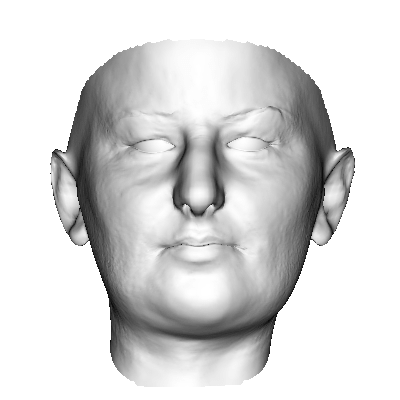}&
\includegraphics[height=\frontimsize, clip=true, trim=40px 0px 35px 25px]{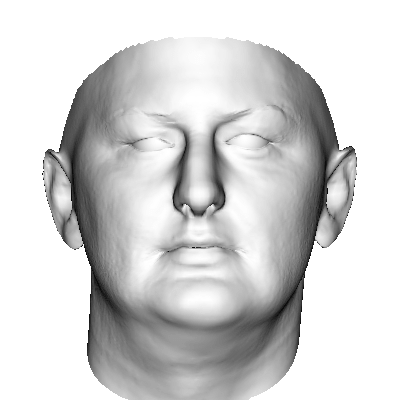}&
\includegraphics[height=\frontimsize, clip=true, trim=40px 0px 35px 25px]{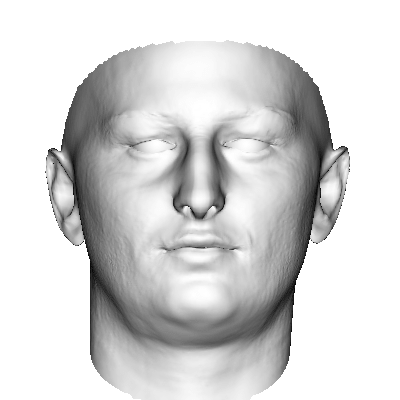}&&&
\end{tabular}
}
\caption{Sparse and dense fitting of the synthetic images. Target at 200cm, fitted results at 60cm.}
\label{fig:SynthPersCompReverse}
%\vspace{-4mm}
\end{figure*}
%Frontal
%Face ID: 5. Level (1:idx 2:edge 3:all):1 | ndims: 70. Landmark distance : 0.391%, surface distance : 5.125 mm.
%Face ID: 5. Level (1:idx 2:edge 3:all):2 | ndims: 70. Landmark distance : 0.376%, surface distance : 5.655 mm.
%Face ID: 5. Level (1:idx 2:edge 3:all):3 | ndims: 199. Landmark distance : 0.361%, surface distance : 2.435 mm.
%Rotated
%Face ID: 10. Level (1:idx 2:edge 3:all):1 | ndims: 70. Landmark distance : 0.507%, surface distance : 7.151 mm.
%Face ID: 10. Level (1:idx 2:edge 3:all):2 | ndims: 70. Landmark distance : 0.516%, surface distance : 4.705 mm.
%Face ID: 10. Level (1:idx 2:edge 3:all):3 | ndims: 199. Landmark distance : 0.494%, surface distance : 3.633 mm.

In Figures \ref{fig:SynthPersComp} and \ref{fig:SynthPersCompReverse} we go further by showing the results of fitting to sparse 2D landmarks (the Farkas feature points), landmarks/edges and all vertices for 4 of the BFM scans (i.e. the targets are real faces). In Fig.~\ref{fig:SynthPersComp}, the target face is close to the camera ($t_z=30$cm) and we fit the model at a far distance ($t_z=120$cm). This configuration is reversed in Fig.~\ref{fig:SynthPersCompReverse} (200cm to 60cm). Since we are only interested in the spatial configuration of features in the image, we show both target and fitted mesh with the texture of the real target face. 
The target perspective projection to which we fit is shown in the first and fifth columns. The fitting result under perspective projection is shown in the second to fourth columns and sixth to eight columns.
To enable comparison between the target and fitted faces, we render them under orthographic projection in rows two and four respectively. The landmarks from the target (plotted as blue circles) and fitted (shown as red dots) face are shown under perspective projection in the column nine. We illustrate edge correspondence (model contours) between faces in the tenth column. In the last column, we average the target and fitted face texture from the dense fitting result, showing that there is no visible difference in the 2D geometry of these two images.
%We now ask whether the perspective ambiguity applies in sparse and dense settings, i.e. whether two different faces can produce the same 2D projections for various vertex density in the mesh. 
%In sparse fitting case, we estimate face shape using only feature points. In the second experiment, we use edge information in addition to landmarks. This gives us better fitting results on both inner (lips, eye corners) and outer (silhouette) model contours. In the dense setting, the 2D shape features are closely reconstructed even when the shaded meshes are viewed. This is despite the target and fitted faces differing by $d_S>1$cm in certain cases. There are clear differences in shading with the dense fittings in Fig.~\ref{fig:SynthPersComp} exhibiting sharper features and hence more dramatic shading and in Fig.~\ref{fig:SynthPersCompReverse}, flatter features and hence flatter shading. 

%These results demonstrate clearly that two faces with significantly different 3D shape can give rise to almost identical 2D landmark positions under perspective projection. This is seen more clearly in rotated views (third rows) of the target and two fitted surfaces. The results in these figures highlight dramatically how faces with very different shapes produce almost identical projections under perspective transformation.
 
%In all cases, $d_S$ is significant, sometimes as low as 0.24cm. On the other hand, $d_L$ is relatively small.

\begin{figure}[!t]
\centering
\begin{tabular}{ccccc}
\bf 488cm & \bf 61cm & \bf Blend & \bf 488cm & \bf 61cm \\
\includegraphics[height=2.3cm, clip=true,trim=900px 10px 850px 235px]{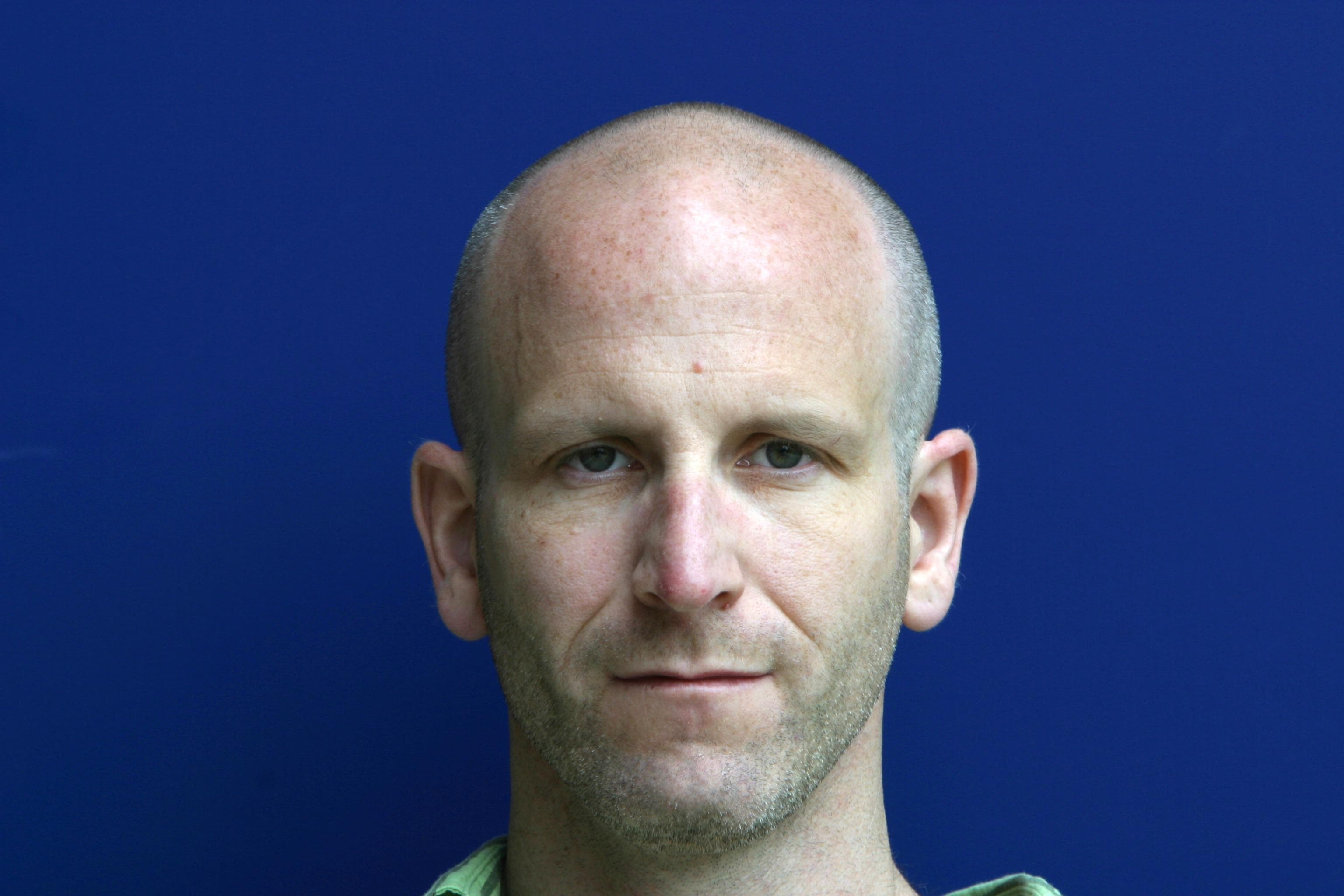}&
\includegraphics[height=2.3cm, clip=true,trim=900px 10px 900px 235px]{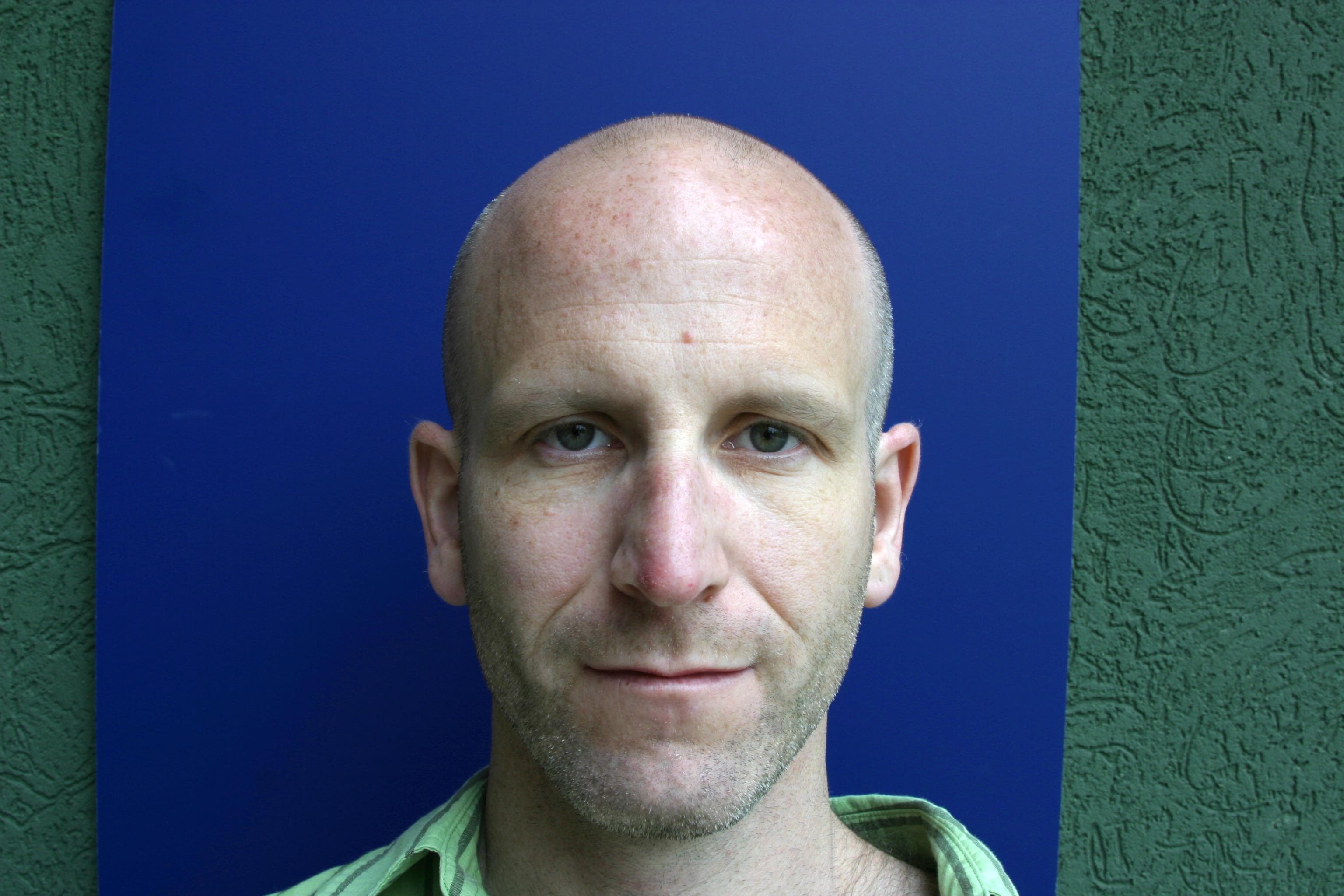}&
\includegraphics[height=2.3cm, clip=true,trim=900px 10px 900px 235px]{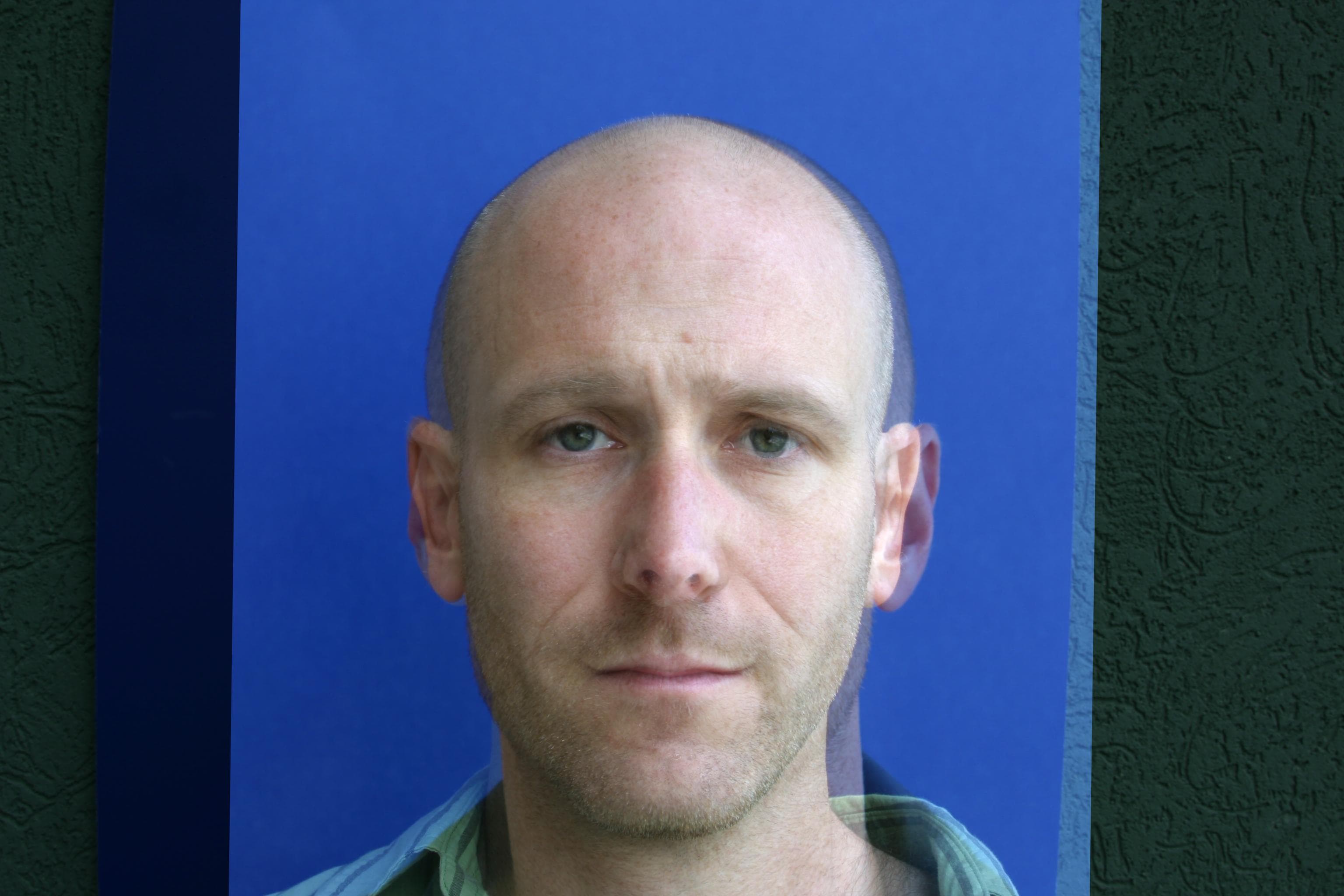}&
\includegraphics[height=2.3cm, clip=true,trim=900px 10px 900px 235px]{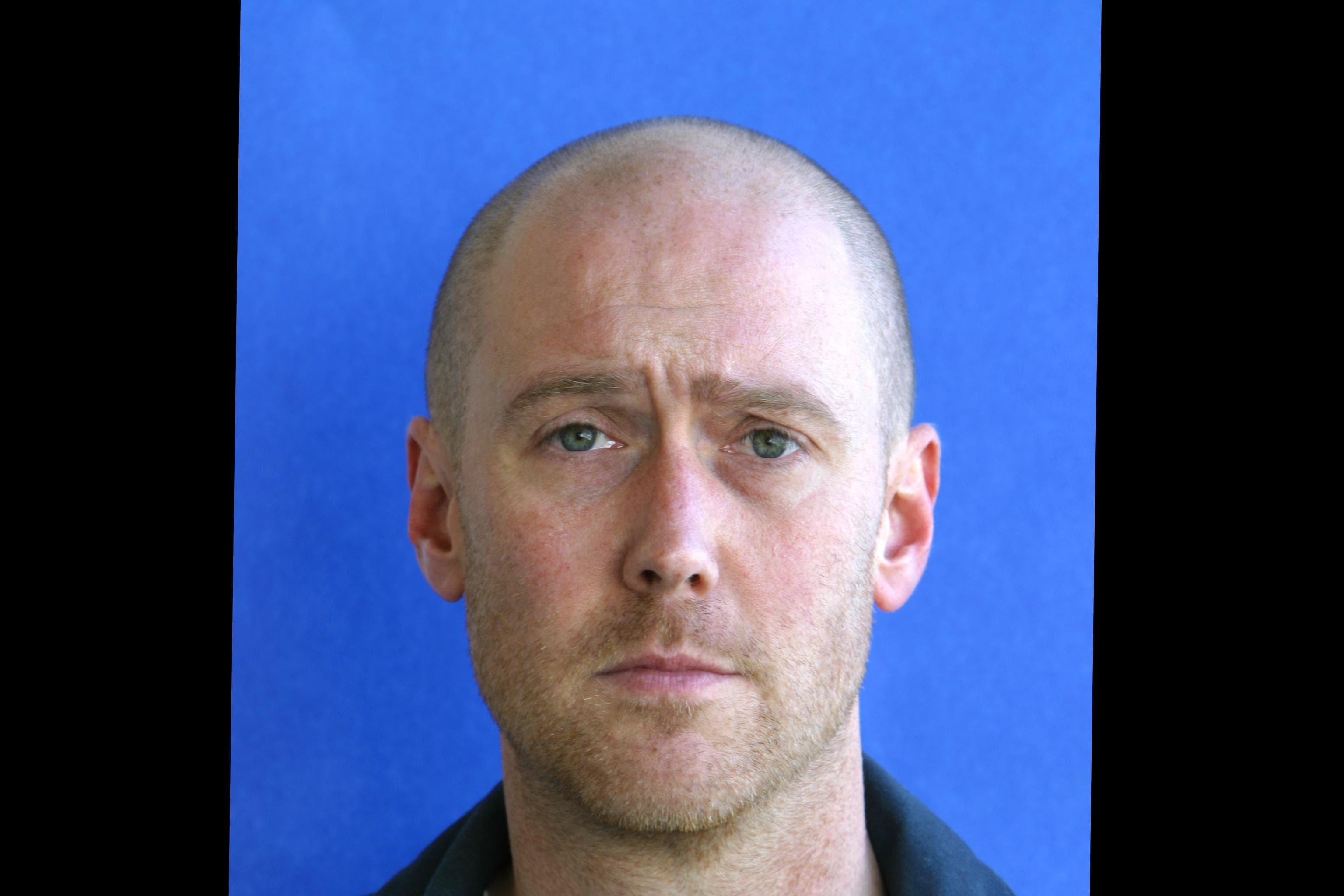}&
\includegraphics[height=2.3cm, clip=true,trim=290px 615px 447px 390px]{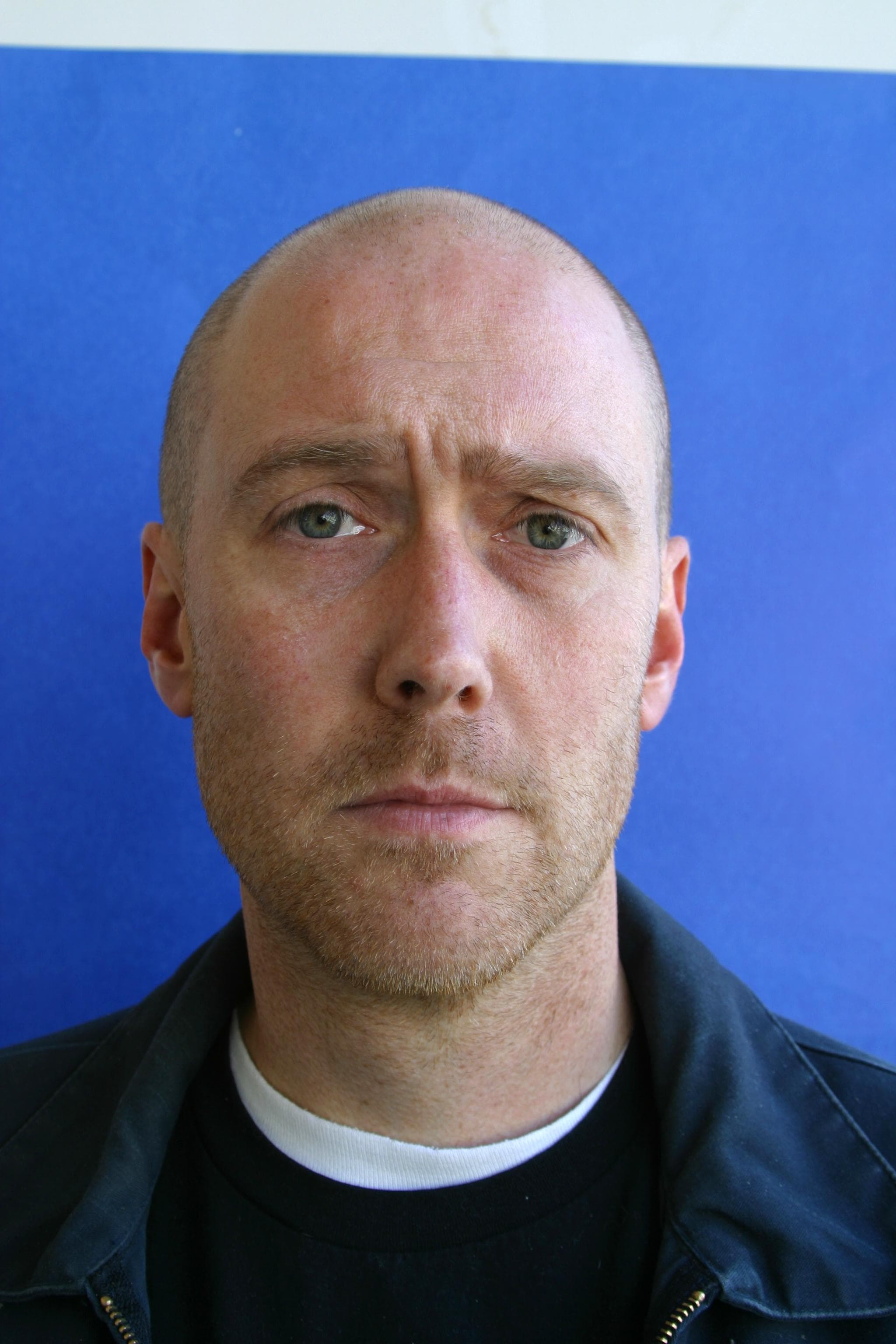}\\
\end{tabular}
\caption{Perspective ambiguity in real faces. Two faces are shown at two different distances. The blend in the middle shows that their 2D geometry is similar when viewed at very different distances.}
\label{fig:RealEg}
\end{figure}

The implication of these results is that, in a sample of real faces, we might expect that two different identities with different face shapes could give rise to approximately the same 2D landmarks when viewed from different distances. We show in Fig.~\ref{fig:RealEg} that this is indeed the case. The Caltech Multi-Distance Portraits dataset \citep{Burgos:14} contains images of 53 subjects viewed at 7 different distances. 55 landmarks are placed manually on each face image. We search for pairs of faces whose landmarks (when viewed at different distances) are close in a Procrustes sense. Despite the small sample size, we find a pair of faces whose mean landmark error is 2.48\% (i.e. they are within the expected accuracy of a landmark detector \citep{sagonas2016300}) when they are viewed at 61cm and 488cm respectively (second and fourth image in the figure). In the third image, we blend these two images to show that their 2D features indeed align well. To highlight that their face shape is in fact quite different, we show their appearance with distances reversed in columns one and five (allowing direct comparison between columns one and four or two and five). E.g. compare column one with column four. The face in column one has larger ears and inner features that are more concentrated towards to the centre of the face compared to the face in column four.

The CMDP data can also be used to demonstrate a surprising conclusion. For all 53 subjects, we compute the mean landmark error between the same identity at 61cm and 488cm which is 3.11\%. Next, for each identity we find the identity at the same distance with the smallest landmark error. Averaged over all identities, this gives a value of 2.86\% for 61cm and 2.83\% for 488cm. We therefore conclude that 2D geometry between different identities at the same distance is more similar than between the same identity at different distances. If the number of identities was increased, the size of this effect would likely increase since the chance of finding closely matching different identity pairs would increase.

\subsection{Beyond geometric cues}

\begin{figure}[!t]
\centering
\noindent\resizebox{.48\textwidth}{!}{
\subfloat[\label{fig:heatmap}]{\includegraphics[height=5cm, clip]{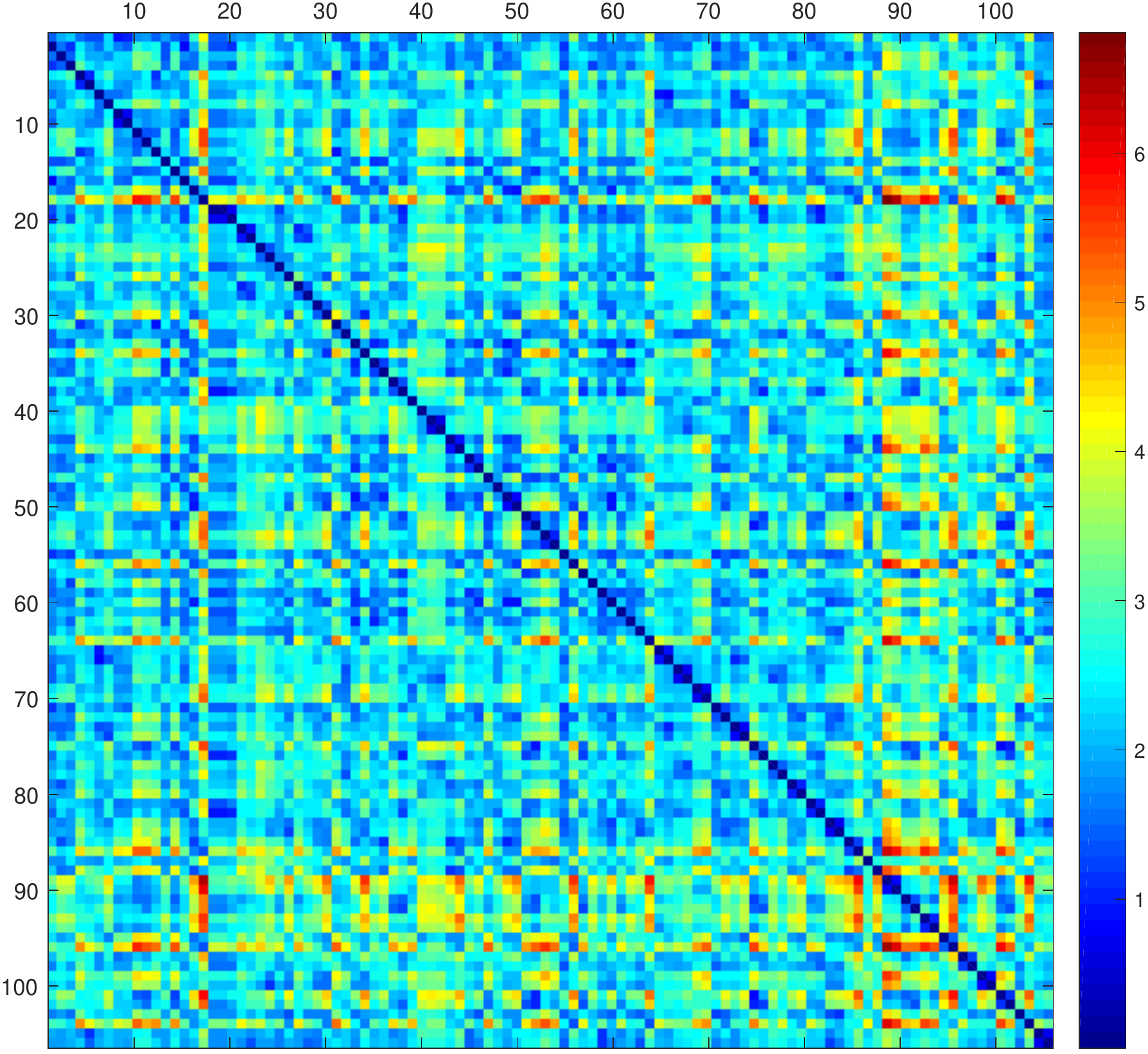}}
\subfloat[\label{fig:binarydistance}]{\includegraphics[height=5cm, clip]{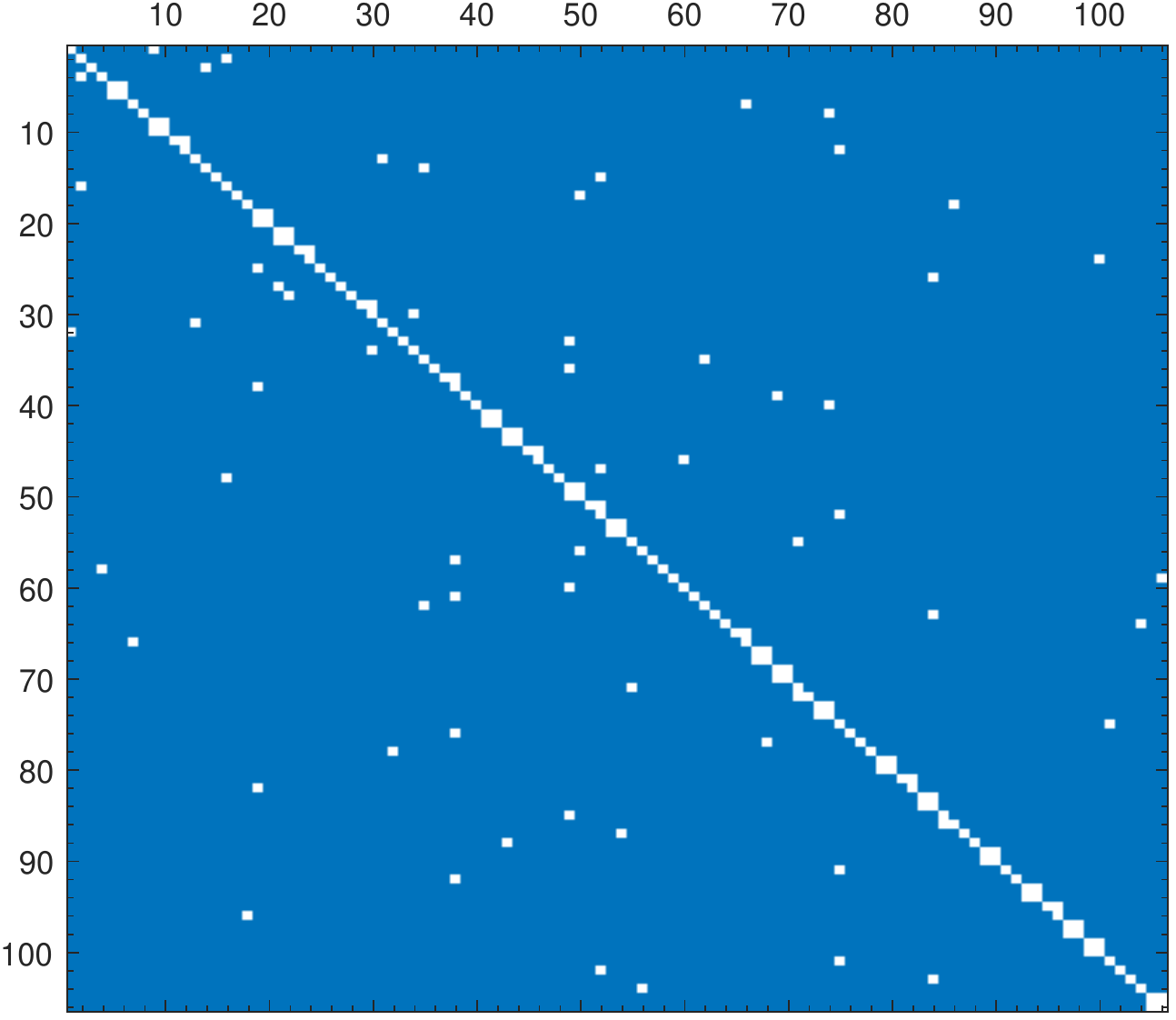}}}
\caption{(a) Heat map and (b) binarised distance matrix visualising similarity between subjects viewed at two different (closest and farthest) distances. We measured the distances between 3D surfaces acquired by running pre-trained \cite{tran2017regressing} on real images from the CMDP dataset. One would expect $2\times 2$ blocks of white on the diagonal if the network is performing perfectly.}
\end{figure}

The fitting methods we propose in this paper use only explicit geometric cues, i.e.~landmarks and contours. State-of-the-art CNN-based methods can exploit any 3D shape cues such as shading, texture, shadows or context from external face features such as hair or clothes or even from background objects. One might suppose that these additional cues resolve the ambiguity we describe. However, we now show that this is not the case. We used the pre-trained network of \cite{tran2017regressing} which is publicly available. This network is trained discriminatively to regress the same 3DMM parameters from different images of the same person. If the training set contained distance variation, then it would be hoped that the network would learn invariance to perspective ambiguities. We ran the network on images of 53 subjects viewed at closest and farthest distances from the CMDP dataset \citep{Burgos:14}. We begin by evaluating the invariance of the shape reconstructions to changes in distance by measuring the mean Euclidean distance after Procrustes alignment between all pairs of 3D reconstructions. This is a standard metric for comparing 3D face reconstructions, e.g.~\cite{RingNet:CVPR:2019,feng2018evaluation}. These comparisons provides a $106\times 106$ distance matrix. One would expect that the shape difference of the same subject viewed at two different distances would be the lowest. However, for the majority of identities, this is not the case. In Fig.~\ref{fig:heatmap} we show the distance matrix (same identity in consecutive positions) and in Fig.~\ref{fig:binarydistance} we binarise this by choosing the best matching shape for each row. Perfect performance would yield $2\times 2$ blocks along the diagonal. We show two examples from this experiment in Fig.~\ref{fig:Tranvis}. These results show that \cite{tran2017regressing} has not learnt invariance to perspective transformation in terms of the metric difference between the shapes themselves.

\begin{figure}[!t]
%\centering
\raggedright
\noindent\resizebox{.45\textwidth}{!}{
\includegraphics[height=5cm, clip]{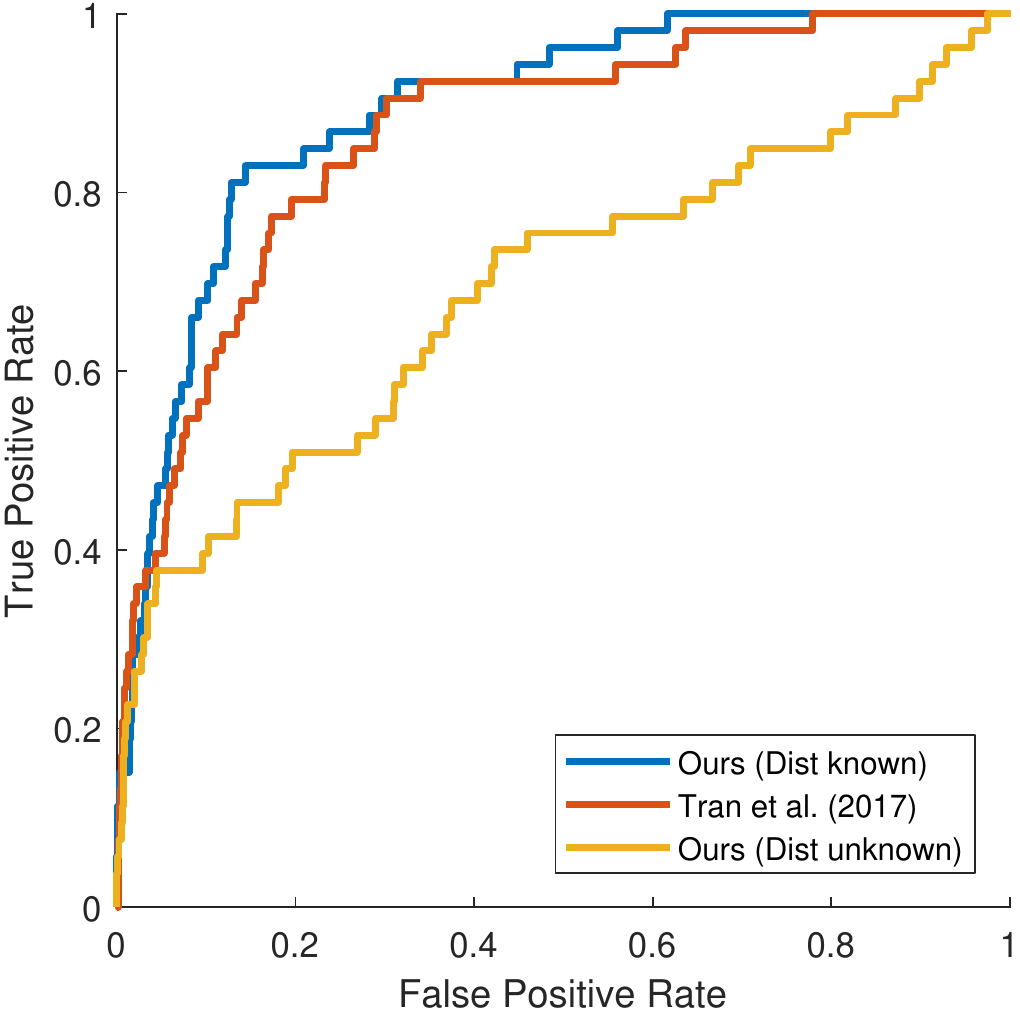}}
\caption{ROC curves of \cite{tran2017regressing} and our method in the distance known and unknown settings on the CMDP dataset.}
%ROC curve. AUC results for Tran: 0.8658 Distknown: 0.8927 Distunknown: 0.69
\label{fig:ROCcurve}
\end{figure}

%%%%%%%%%%%%%%%%%%%%% We can include the recognition experiment here
Another hypothesis is that the shape parameters themselves estimated by \cite{tran2017regressing} may be discriminative across distance for the purposes of recognition. We compute the normalised dot product distance for each shape vector at one distance against all shape vectors at the other distance. This allows us to compare the discriminativeness of the parameters under perspective transformation. We compare against our perspective fitting with either unknown or known subject-camera distance and show ROC curves for all three methods in Fig.~\ref{fig:ROCcurve}. The area under curve (AUC) values for \cite{tran2017regressing} and our method with known distances and unknown distances are 0.866, 0.892 and 0.690, respectively. Using only geometric information and with unknown distance, it is clear that the estimated shape and hence parameters are ambiguous and perform poorly for recognition. \cite{tran2017regressing} has clearly learnt some invariance to distance but performance is still far from perfect on what is a fairly trivial dataset in the context of face recognition. With distance known (and hence the ambiguity avoided), even using only very sparse geometric information we obtain the best performance.
%This provides two 3D reconstructions for each subject. We calculated the mean Euclidean distance after Procrustres alignment between all pairs of reconstructions to form a $106\times 106$ distance matrix. 
%%%%%%%%%%%%%%%%%%%%

%\edit{In the second experiment, we examine the distances between the identity descriptors obtained by pre-trained VGGFace \citep{parkhi2015deep} and VGGFace2\citep{cao2018vggface2} networks.}

% Tran et al results

\subsection{Flexibility modes}

We now explore the flexibility that remains when a model has been fitted to 2D geometric information.
There is a surprising amount of remaining flexibility. Using the 70 Farkas landmark points under orthographic projection in a frontal pose, the BFM has around 50 flexibility modes that change the 3D shape by $k_1=2$mm while inducing a mean change in landmark position of less than $k_2=2$ pixels. Restricting consideration to those flexibility modes where the shape parameter vector remains ``plausible'' (i.e. stays within 3 standard deviations of the expected Mahalanobis length \citep{patel2016manifold}), the number reduces to 7. This still means that knowing the exact 2D location of 70 landmark points only reduces the space of possible 3D face shapes to a 7D subspace of the morphable model.

\newcommand{\frontsizefull}{2.7cm}
\newcommand{\frontsizehalf}{1.6cm}
\begin{figure}[!t]
\centering
\begin{tabular}{ccc}
%\bf 488cm & \bf 61cm & \bf Blend & \bf 488cm & \bf 61cm \\
\includegraphics[width=\frontsizefull, clip=true]{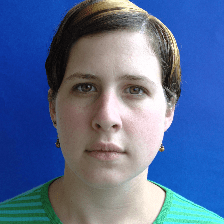}&
\includegraphics[width=\frontsizefull, clip=true]{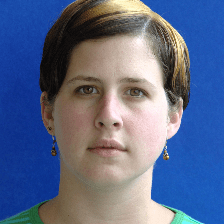}&
\includegraphics[width=\frontsizefull, clip=true]{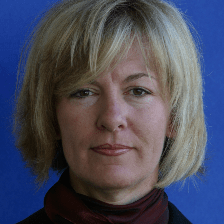}\\
\end{tabular}
\begin{tabular}{ccccc}
\includegraphics[width=\frontsizehalf, clip=true]{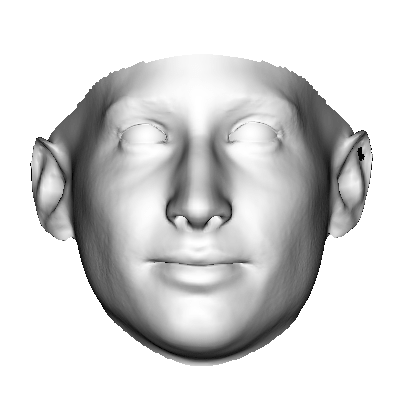}&
\includegraphics[width=\frontsizehalf, clip=true]{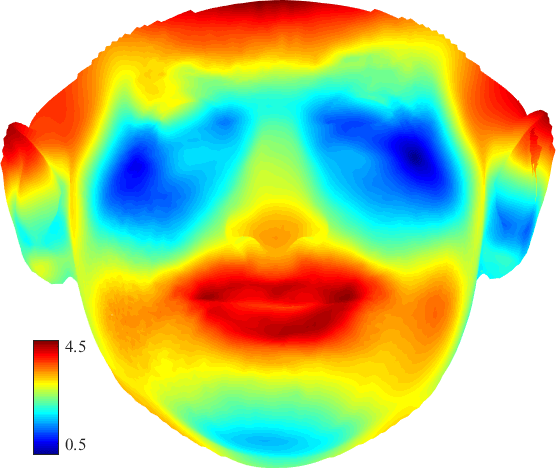}&
\includegraphics[width=\frontsizehalf, clip=true]{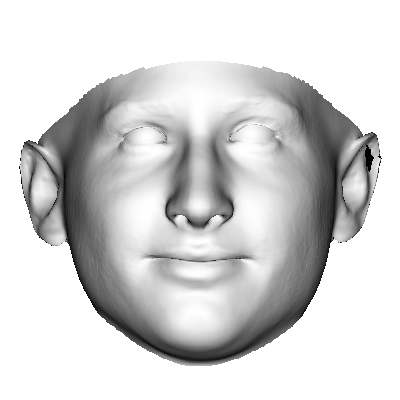}&
\includegraphics[width=\frontsizehalf, clip=true]{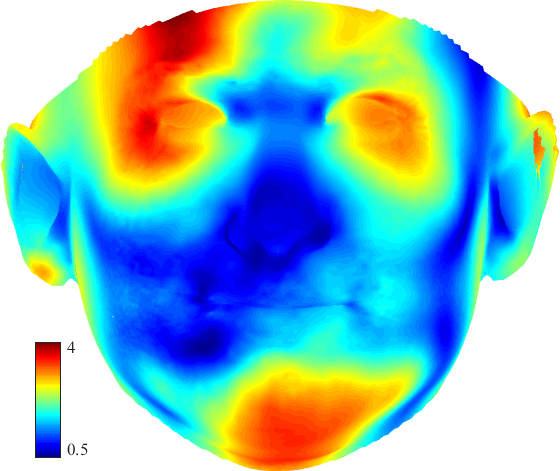}&
\includegraphics[width=\frontsizehalf, clip=true]{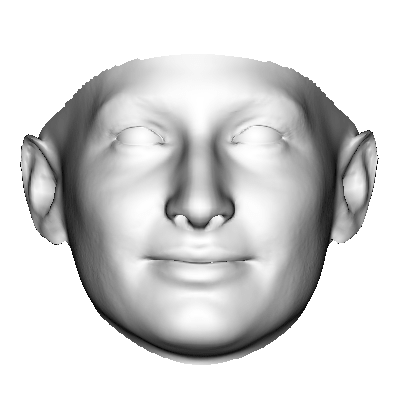}
\end{tabular}
\begin{tabular}{ccc}
%\bf 488cm & \bf 61cm & \bf Blend & \bf 488cm & \bf 61cm \\
\includegraphics[width=\frontsizefull, clip=true]{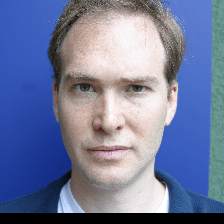}&
\includegraphics[width=\frontsizefull, clip=true]{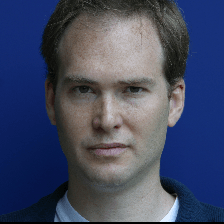}&
\includegraphics[width=\frontsizefull, clip=true]{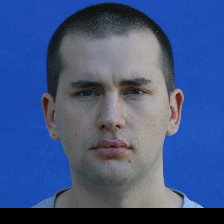}\\
\end{tabular}
\begin{tabular}{ccccc}
\includegraphics[width=\frontsizehalf, clip=true]{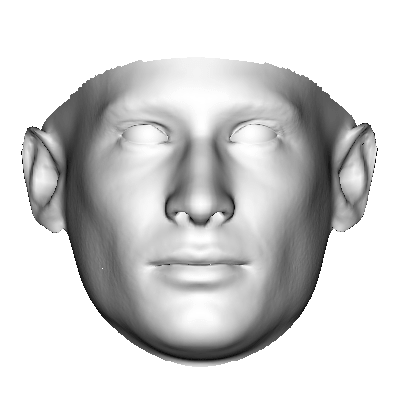}&
\includegraphics[width=\frontsizehalf, clip=true]{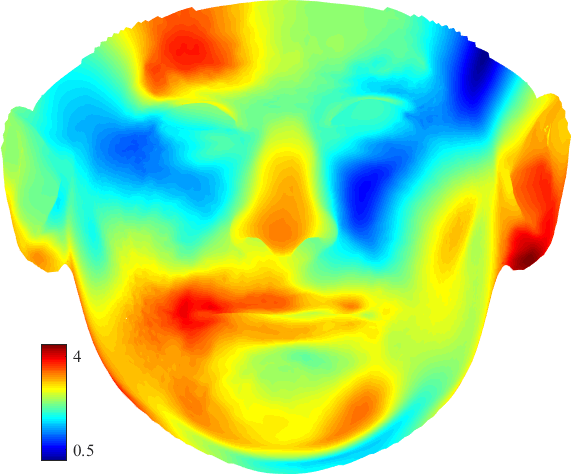}&
\includegraphics[width=\frontsizehalf, clip=true]{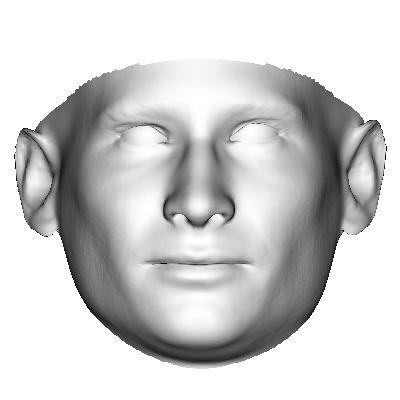}&
\includegraphics[width=\frontsizehalf, clip=true]{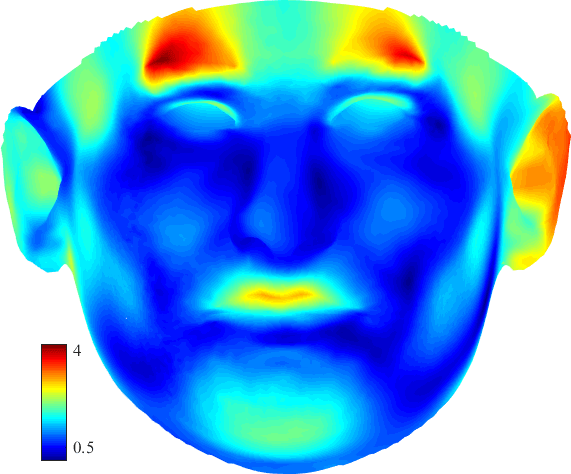}&
\includegraphics[width=\frontsizehalf, clip=true]{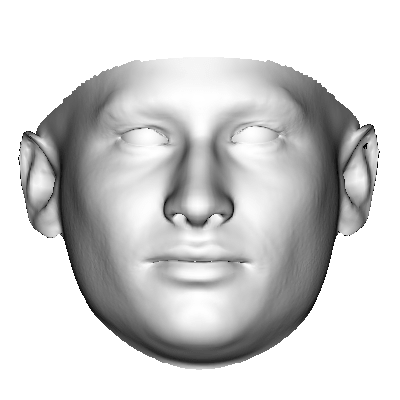}
\end{tabular}
\caption{\cite{tran2017regressing} regresses face shapes that are more different for for the same face viewed at different distances (2nd Row: 2.62mm, 4th Row: 2.5mm) than for different identities at the same distance (2nd Row: 1.79mm, 4th Row: 1.26mm).}
% \caption{\cite{tran2017regressing} regresses significantly different face shapes from the same face viewed at different distances (1.62mm difference between left and middle) while regressing more similar face shapes from different identities (0.88mm between middle and right).}
\label{fig:Tranvis}
\end{figure}
%

% ROC curve

\newcommand{\frontsize}{5cm}
\setlength{\tabcolsep}{1pt}
\begin{figure*}[!t]
\centering
\noindent\resizebox{\textwidth}{!}{
\begin{tabular}{ccccccc}

\includegraphics[height=\frontsize, clip=true,trim=50px 40px 70px 160px]{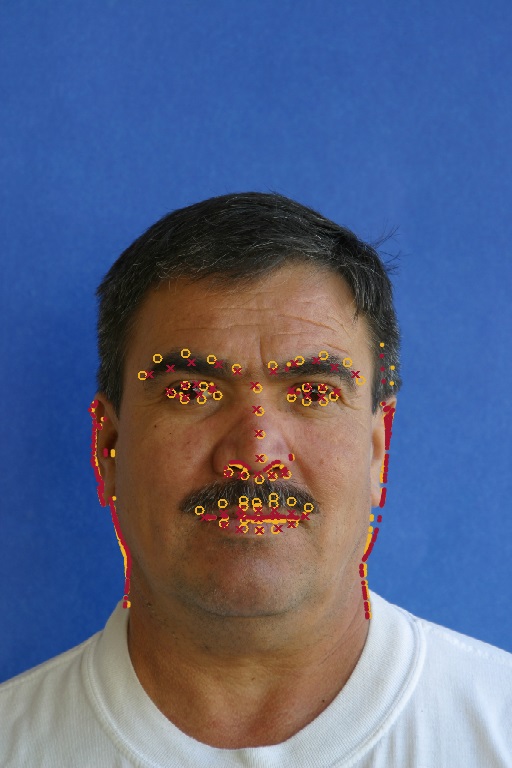}&
\includegraphics[height=\frontsize, clip=true,trim=160px 90px 170px 150px]{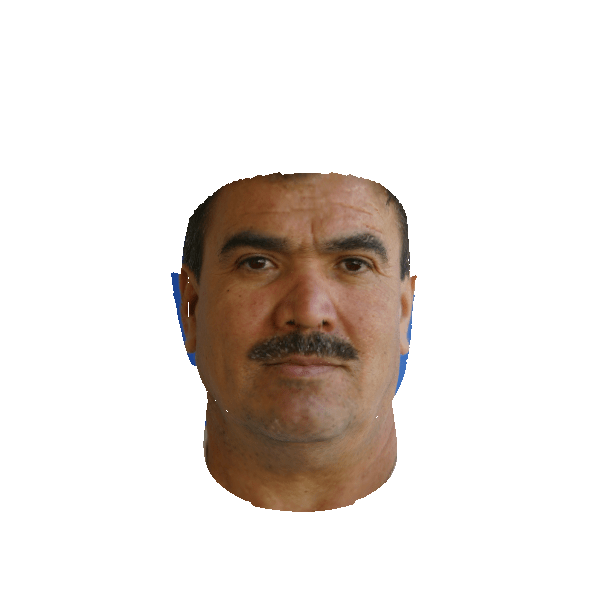}&
\includegraphics[height=\frontsize, clip=true,trim=160px 90px 170px 150px]{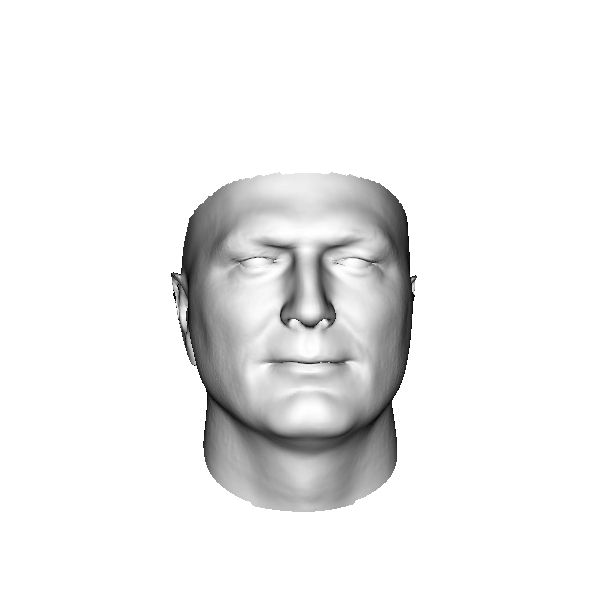}&
\includegraphics[height=\frontsize, clip=true,trim=40px 90px 270px 150px]{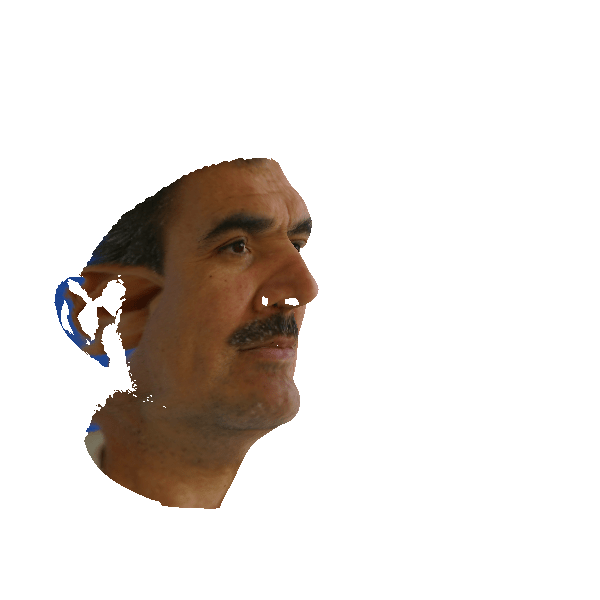}&
\includegraphics[height=\frontsize, clip=true,trim=10px	90px 300px 150px]{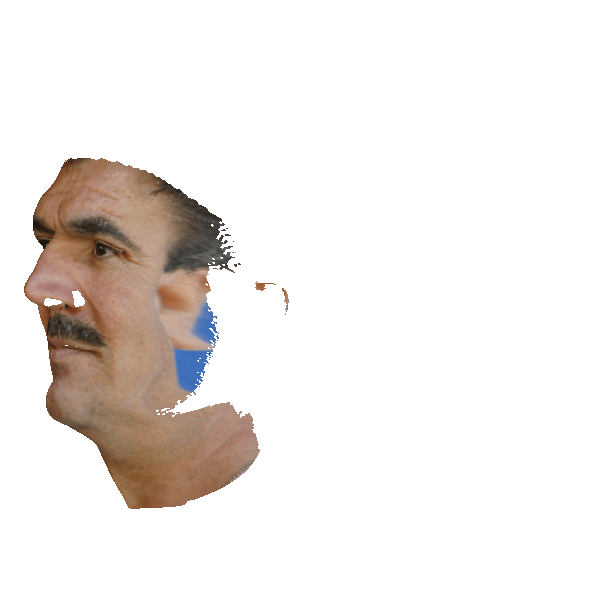}&
\includegraphics[height=\frontsize, clip=true,trim=40px 90px 270px 150px]{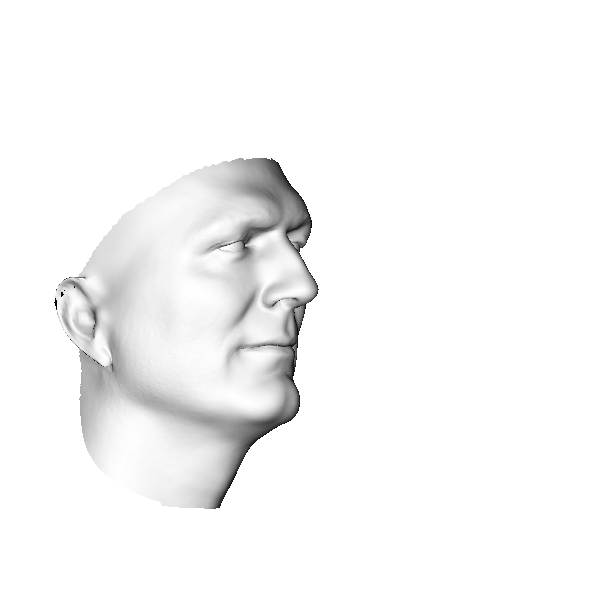}&
\includegraphics[height=\frontsize, clip=true,trim=10px	90px 300px 150px]{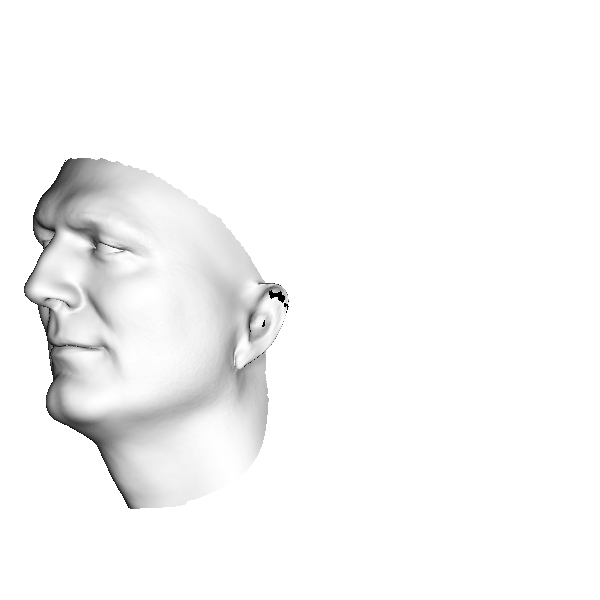}\\

\includegraphics[height=\frontsize, clip=true,trim=50px 40px 70px 160px]{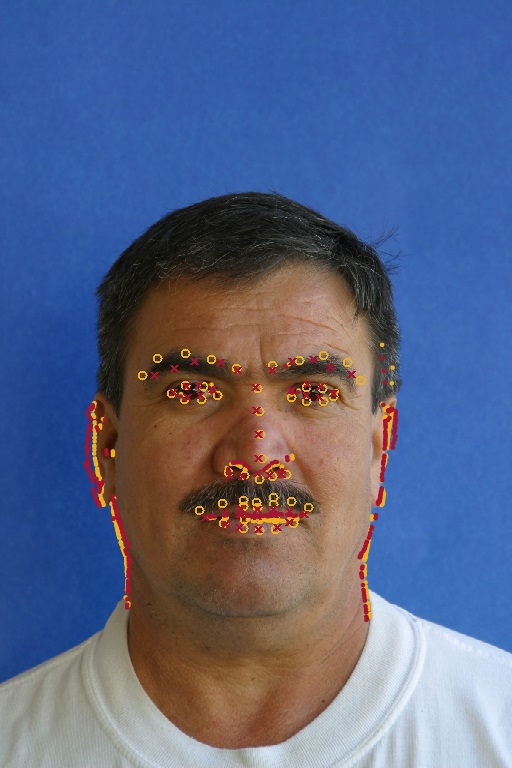}&
\includegraphics[height=\frontsize, clip=true,trim=160px 90px 170px 150px]{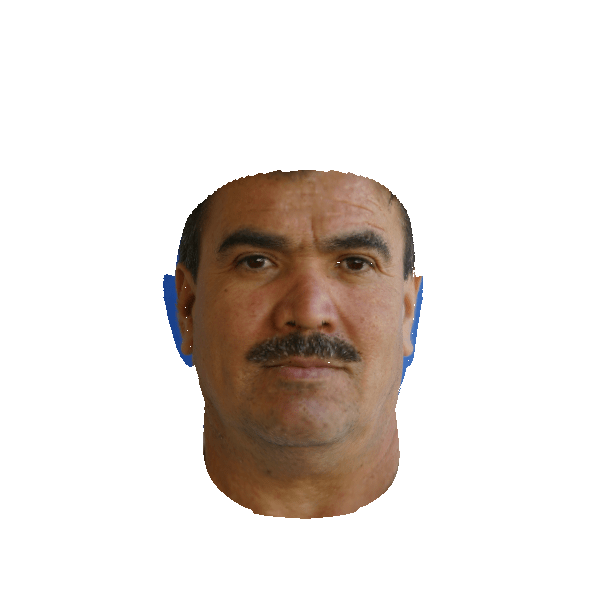}&
\includegraphics[height=\frontsize, clip=true,trim=160px 90px 170px 150px]{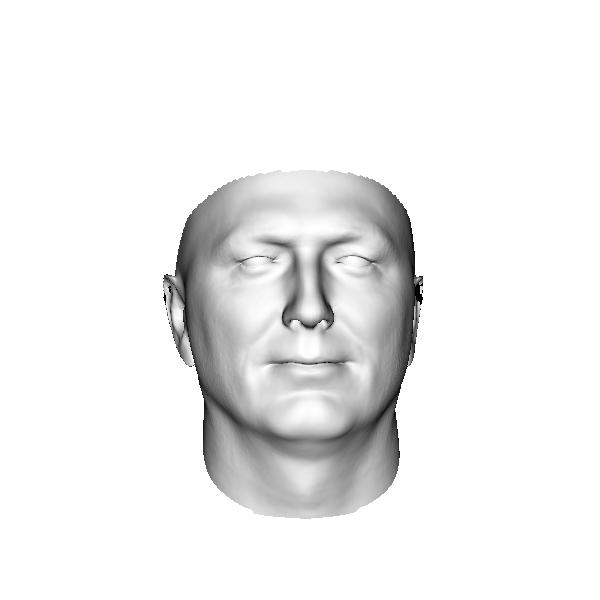}&
\includegraphics[height=\frontsize, clip=true,trim=60px 90px 270px 150px]{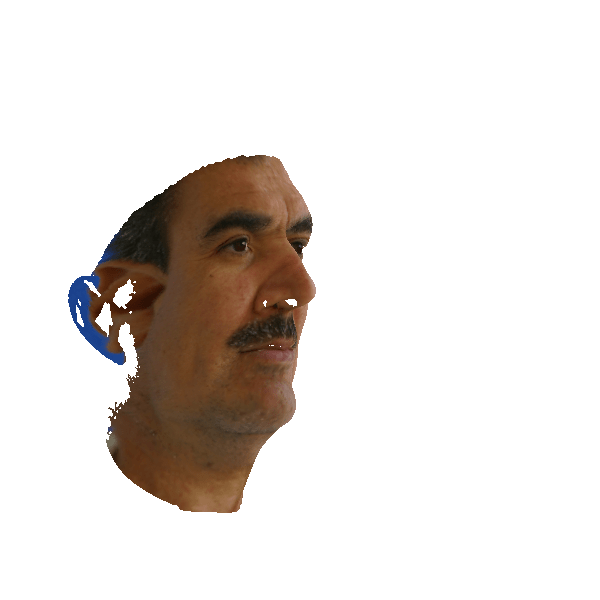}&
\includegraphics[height=\frontsize, clip=true,trim=10px	90px 320px 150px]{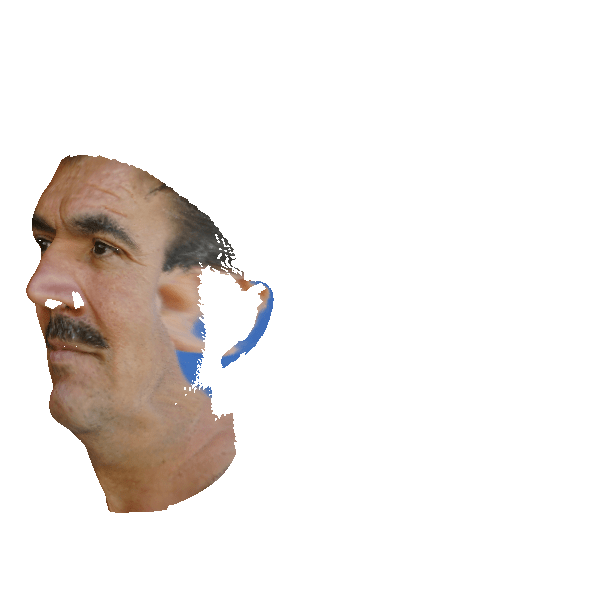}&
\includegraphics[height=\frontsize, clip=true,trim=60px 90px 270px 150px]{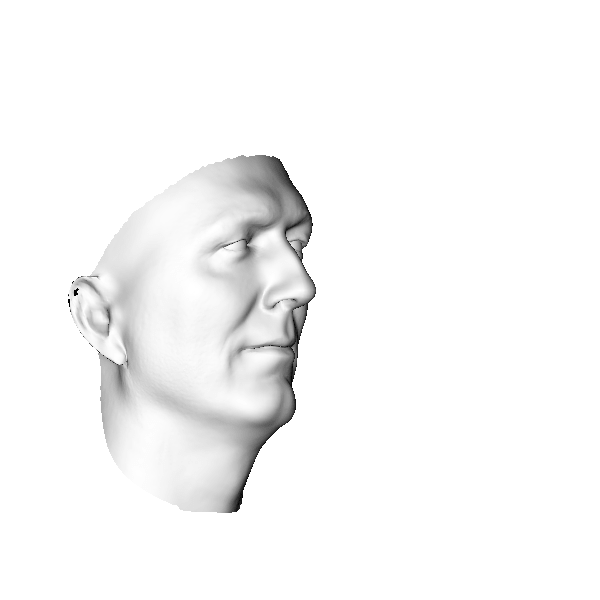}&
\includegraphics[height=\frontsize, clip=true,trim=10px	90px 320px 150px]{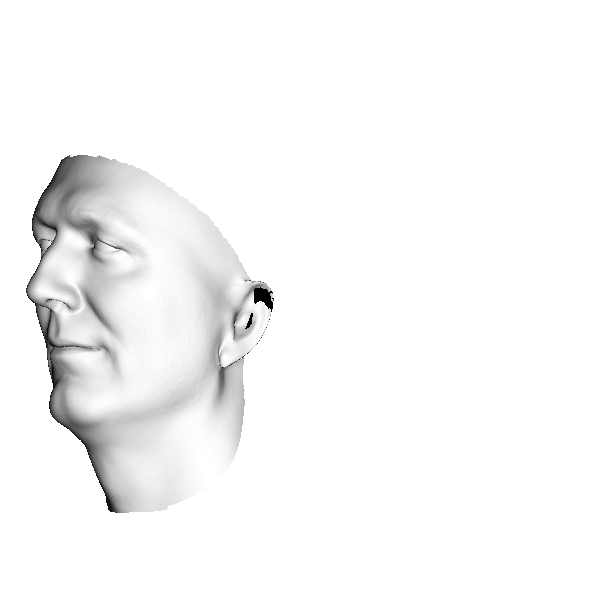}\\

\includegraphics[height=\frontsize, clip=true,trim=50px 40px 70px 160px]{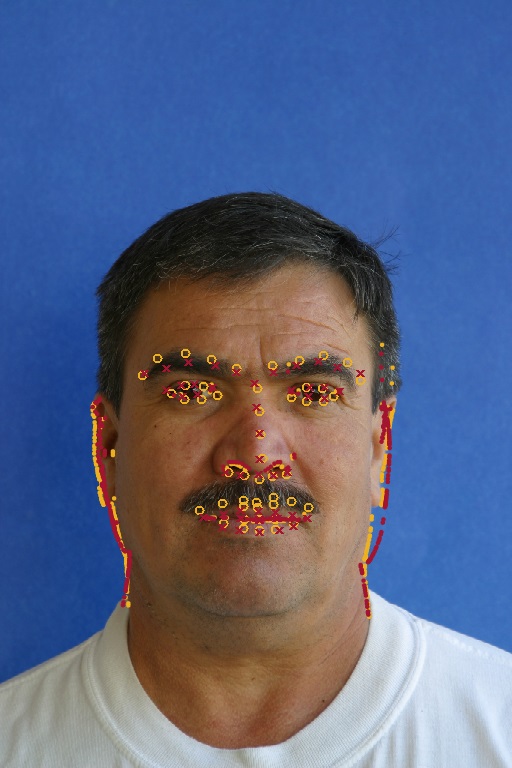}&
\includegraphics[height=\frontsize, clip=true,trim=160px 90px 170px 150px]{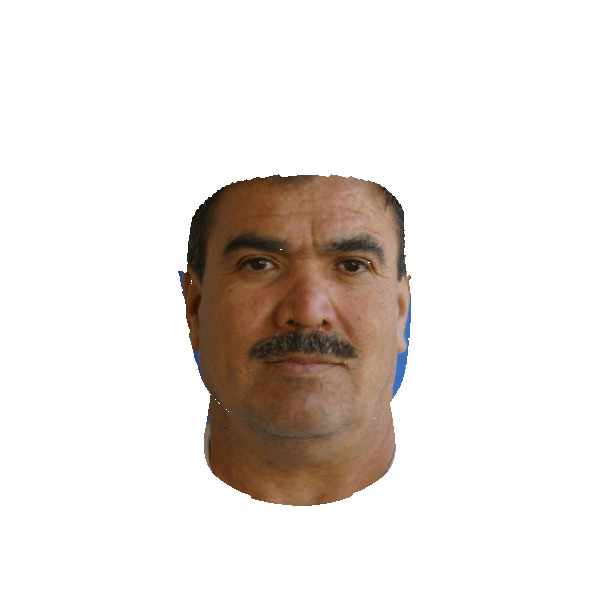}&
\includegraphics[height=\frontsize, clip=true,trim=160px 90px 170px 150px]{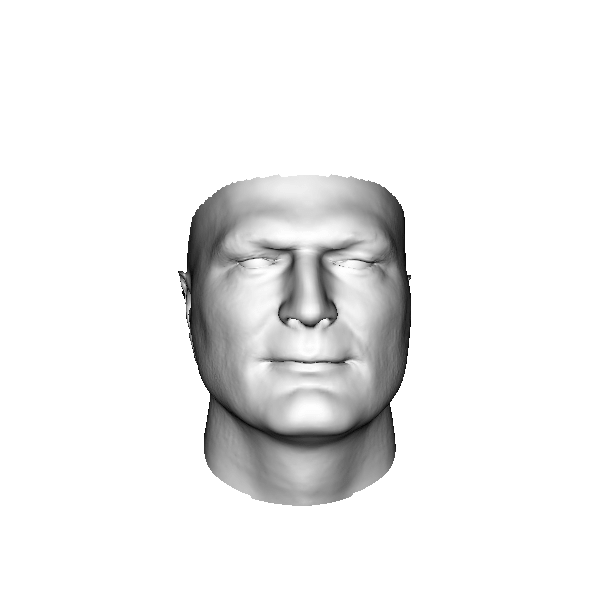}&
\includegraphics[height=\frontsize, clip=true,trim=40px 90px 270px 150px]{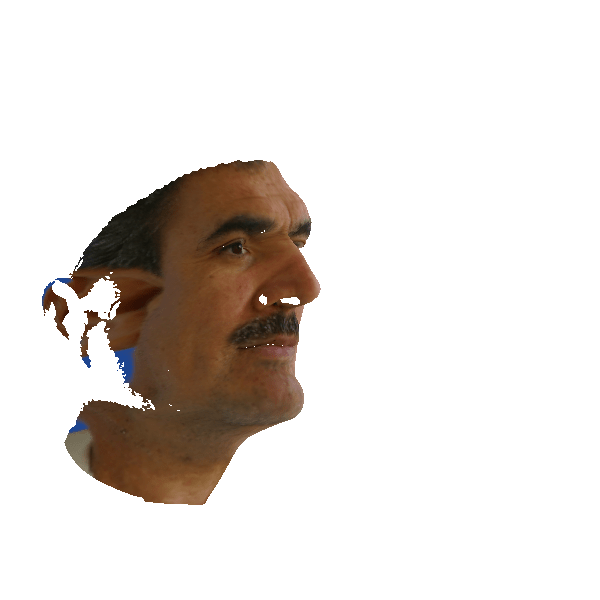}&
\includegraphics[height=\frontsize, clip=true,trim=10px	90px 300px 150px]{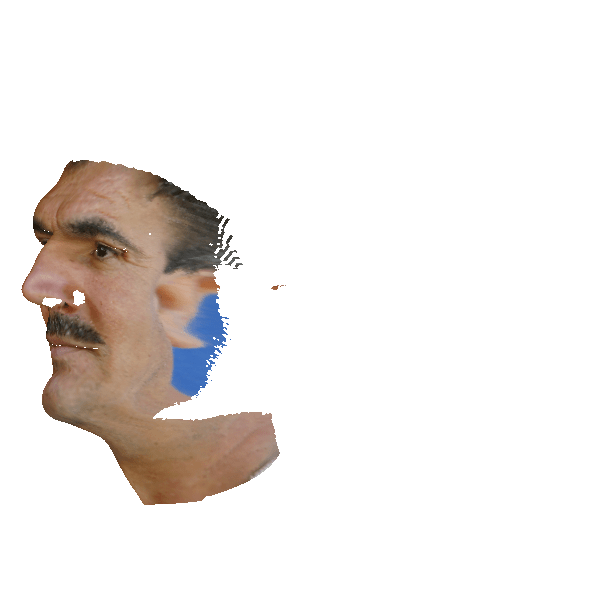}&
\includegraphics[height=\frontsize, clip=true,trim=40px 90px 270px 150px]{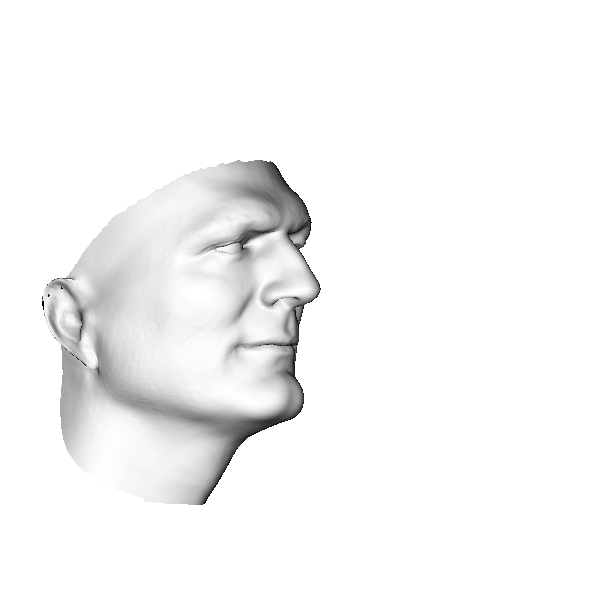}&
\includegraphics[height=\frontsize, clip=true,trim=10px	90px 300px 150px]{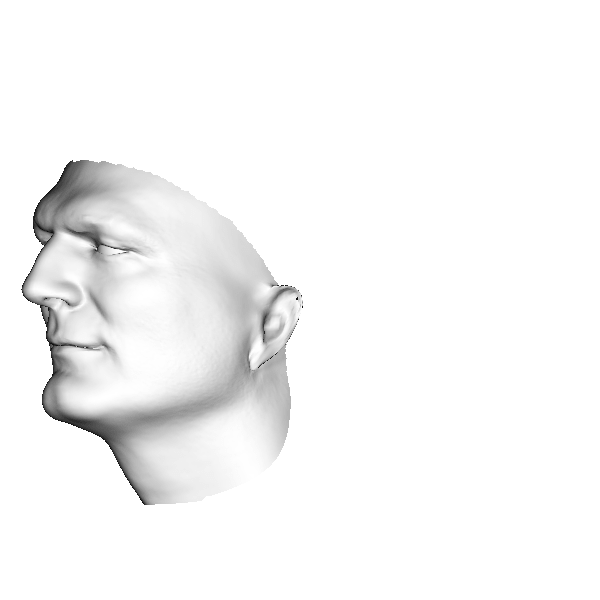}\\

\end{tabular}
}
\caption{Orthographic fitting with flexibility modes. 1st Row: Landmark and edge fitting. 2nd/3rd Row: The first plus and minus flexibility components. Landmark distance is 1.14\% and surface distance is 10mm.}
\label{fig:orthoFlex}
\end{figure*}

\begin{figure*}[!t]
\centering
\noindent\resizebox{\textwidth}{!}{
\begin{tabular}{ccccccc}

\includegraphics[height=\frontsize, clip=true,trim=220px 0px 180px 80px]{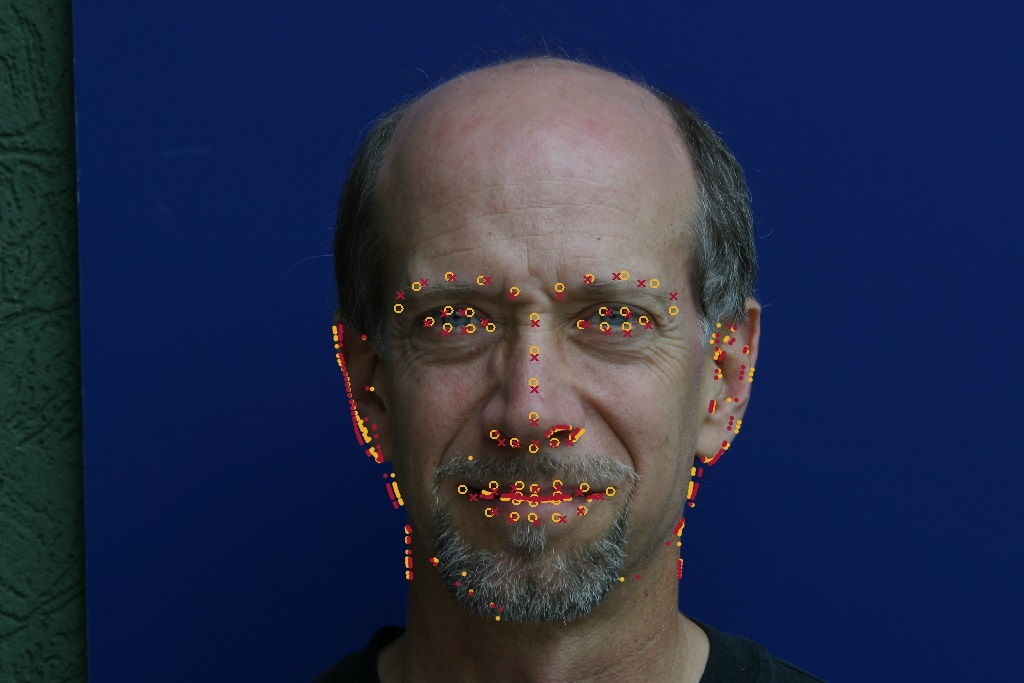}&
\includegraphics[height=\frontsize, clip=true,trim=140px 75px 155px 125px]{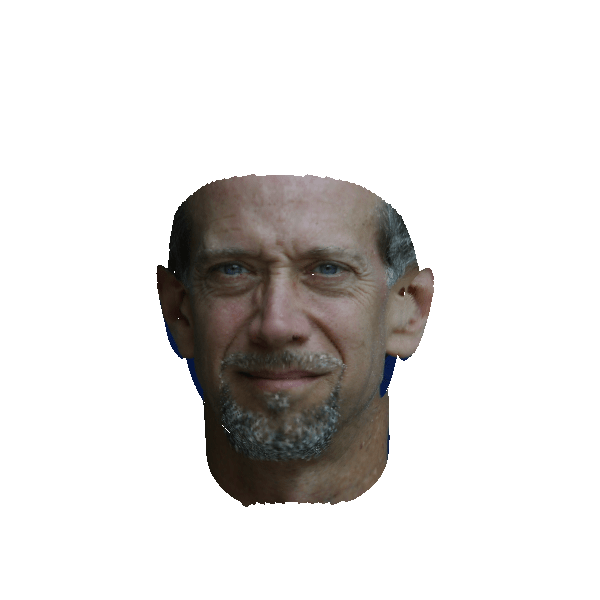}&
\includegraphics[height=\frontsize, clip=true,trim=140px 75px 155px 125px]{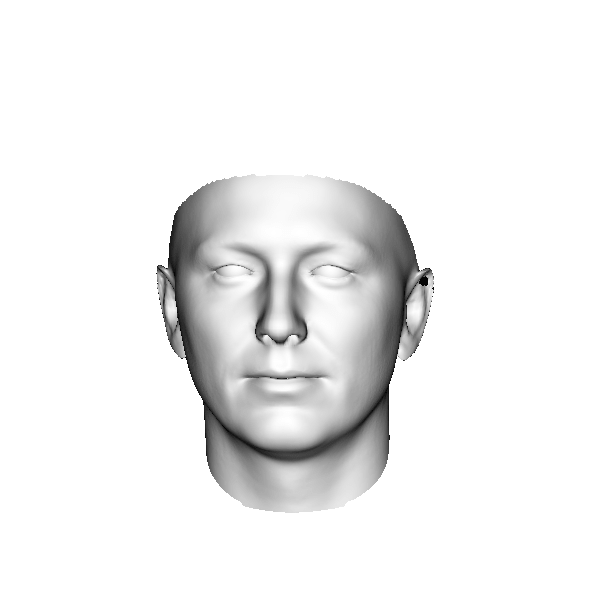}&
\includegraphics[height=\frontsize, clip=true,trim=50px 75px 260px 125px]{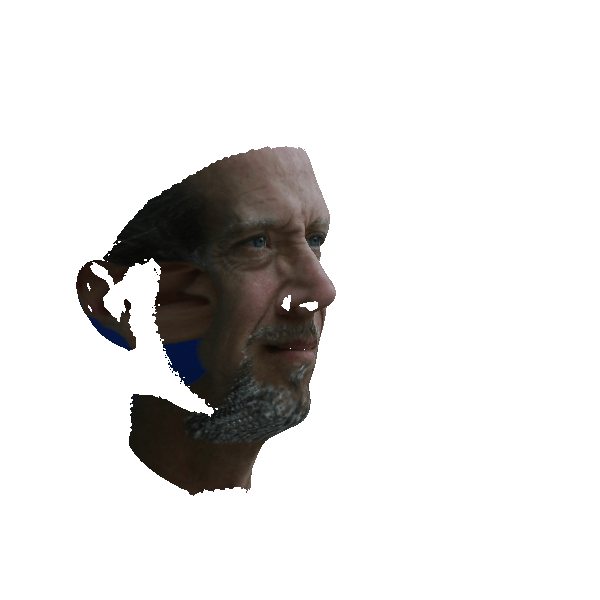}&
\includegraphics[height=\frontsize, clip=true,trim=0px	75px 310px 125px]{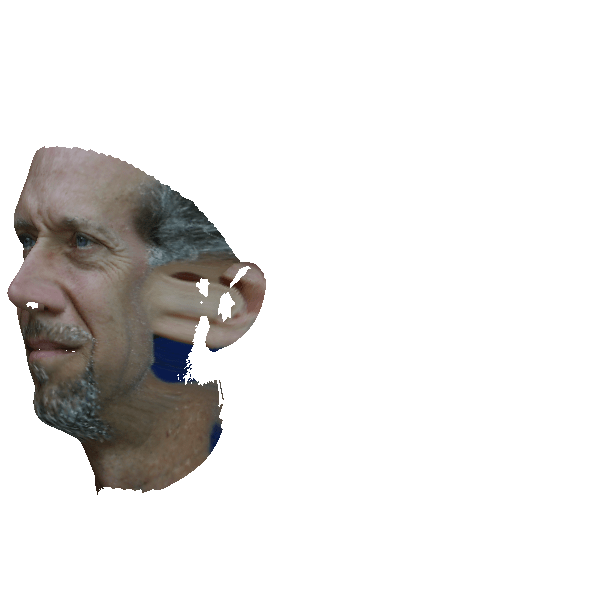}&
\includegraphics[height=\frontsize, clip=true,trim=50px 75px 260px 125px]{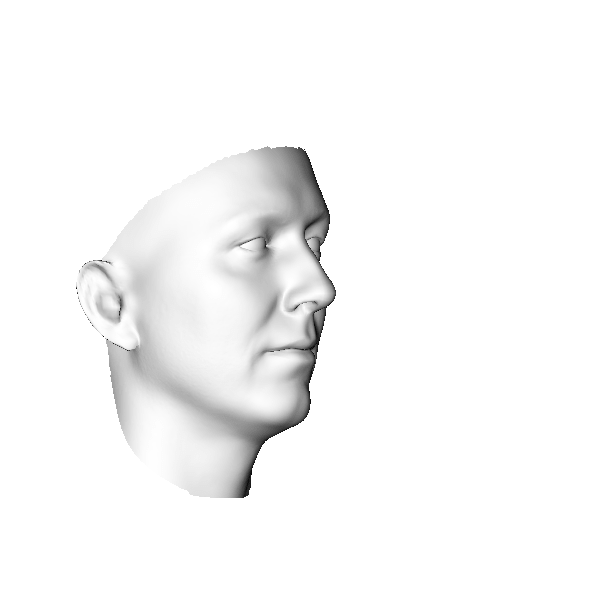}&
\includegraphics[height=\frontsize, clip=true,trim=0px	75px 310px 125px]{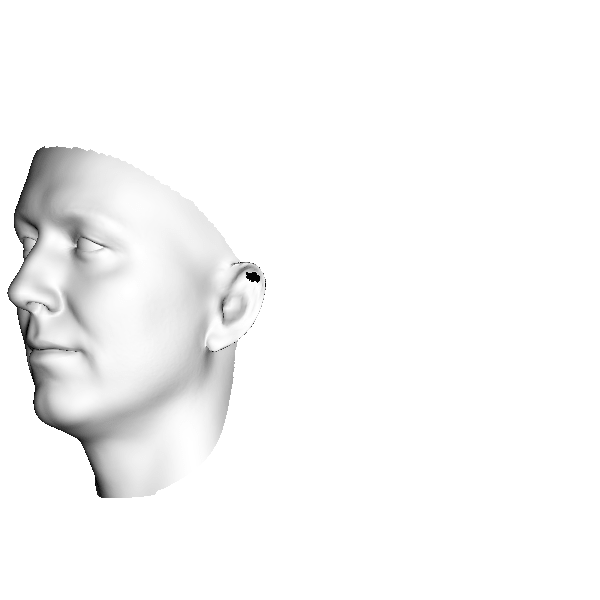}\\

\includegraphics[height=\frontsize, clip=true,trim=220px 0px 180px 80px]{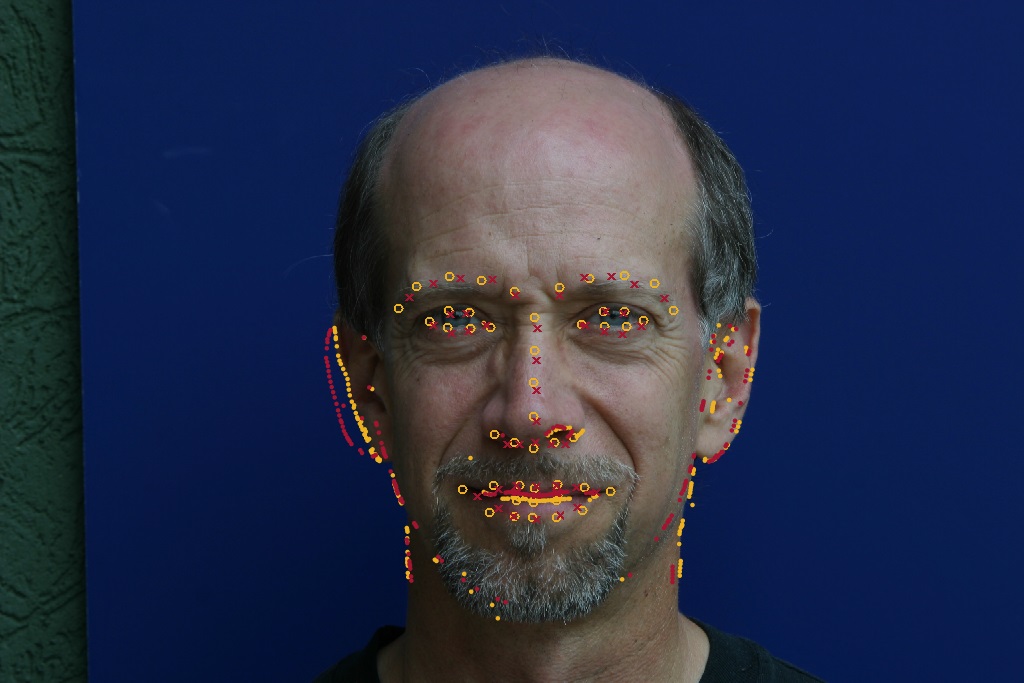}&
\includegraphics[height=\frontsize, clip=true,trim=140px 75px 155px 125px]{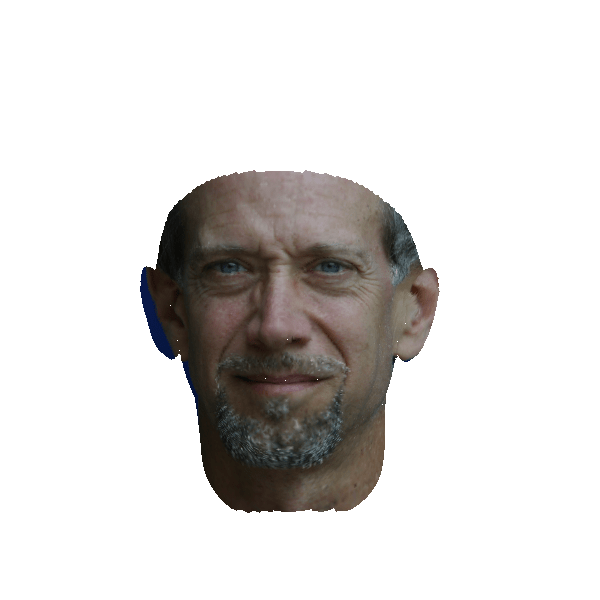}&
\includegraphics[height=\frontsize, clip=true,trim=140px 75px 155px 125px]{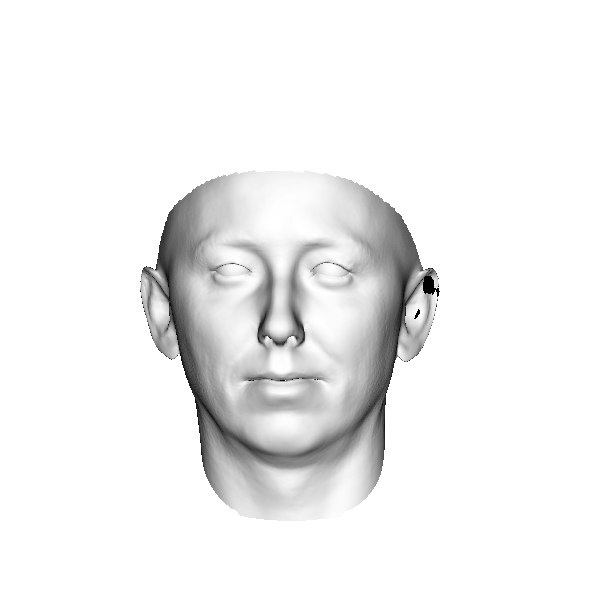}&
\includegraphics[height=\frontsize, clip=true,trim=80px 75px 260px 125px]{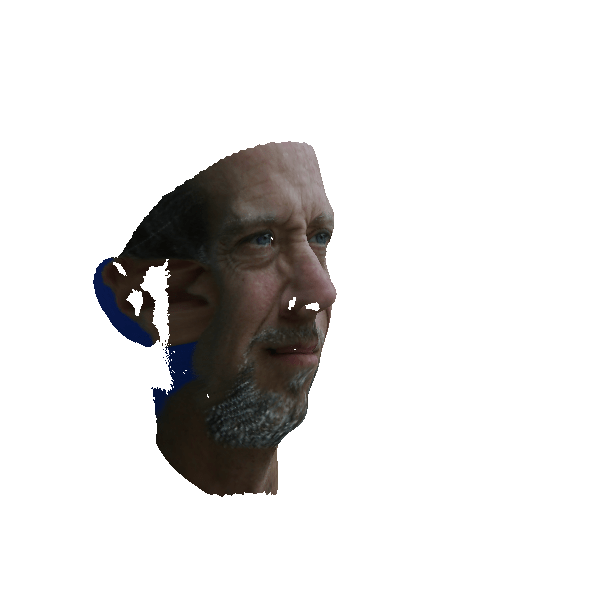}&
\includegraphics[height=\frontsize, clip=true,trim=0px	75px 350px 125px]{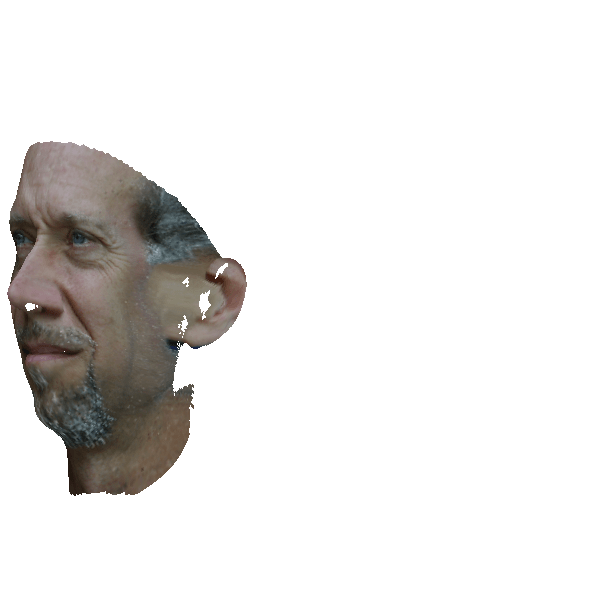}&
\includegraphics[height=\frontsize, clip=true,trim=80px 75px 260px 125px]{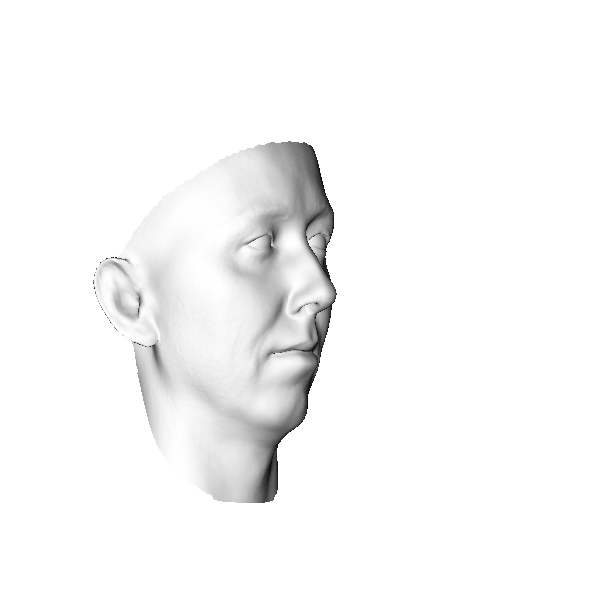}&
\includegraphics[height=\frontsize, clip=true,trim=0px	75px 350px 125px]{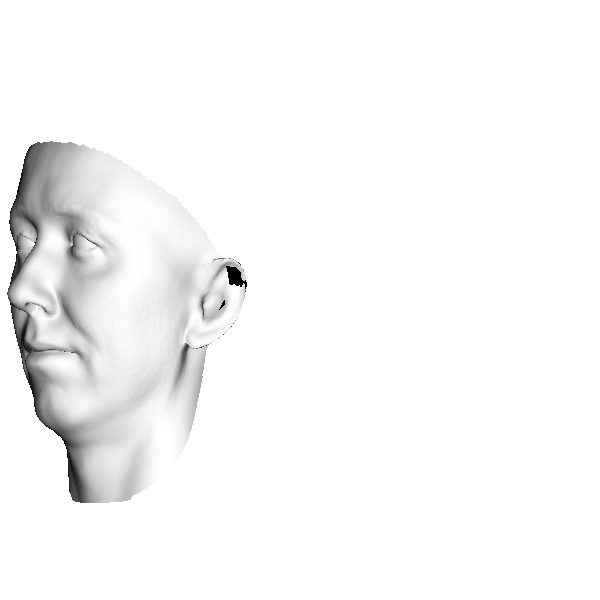}\\

\includegraphics[height=\frontsize, clip=true,trim=220px 0px 180px 80px]{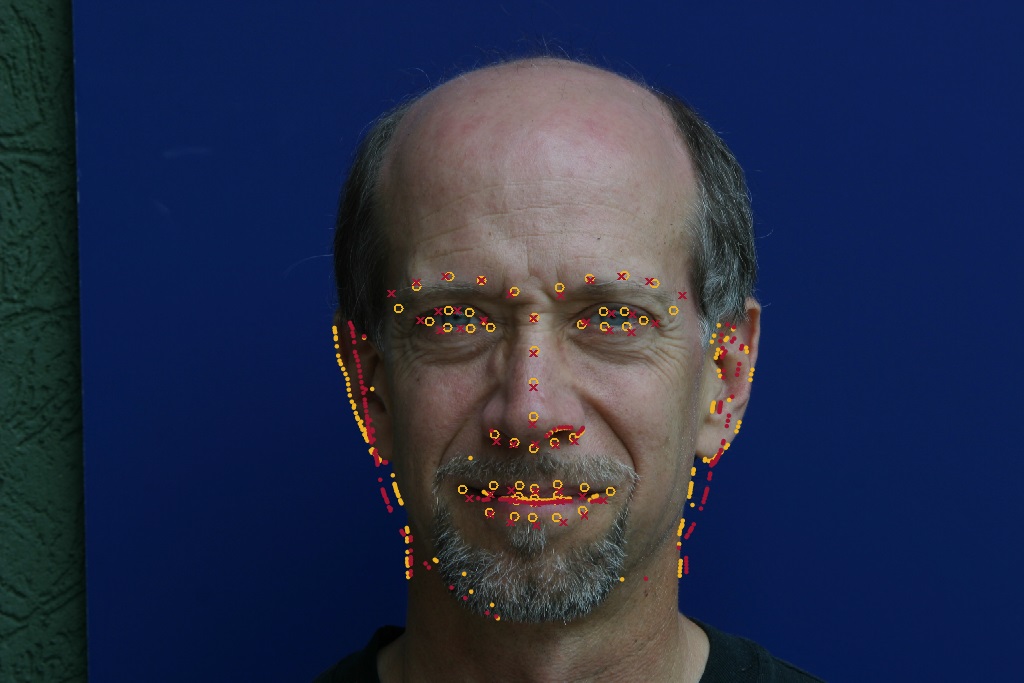}&
\includegraphics[height=\frontsize, clip=true,trim=140px 75px 155px 125px]{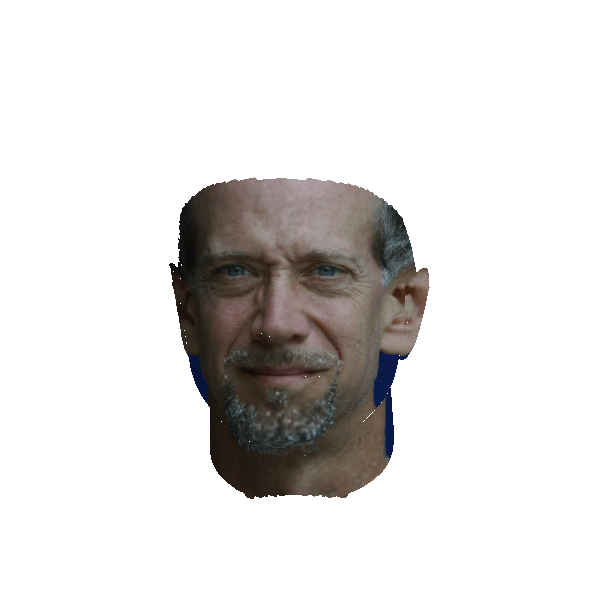}&
\includegraphics[height=\frontsize, clip=true,trim=140px 75px 155px 125px]{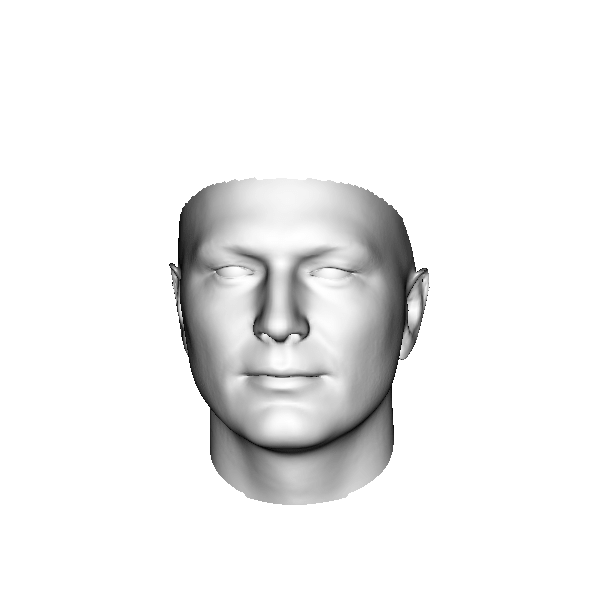}&
\includegraphics[height=\frontsize, clip=true,trim=50px 75px 260px 125px]{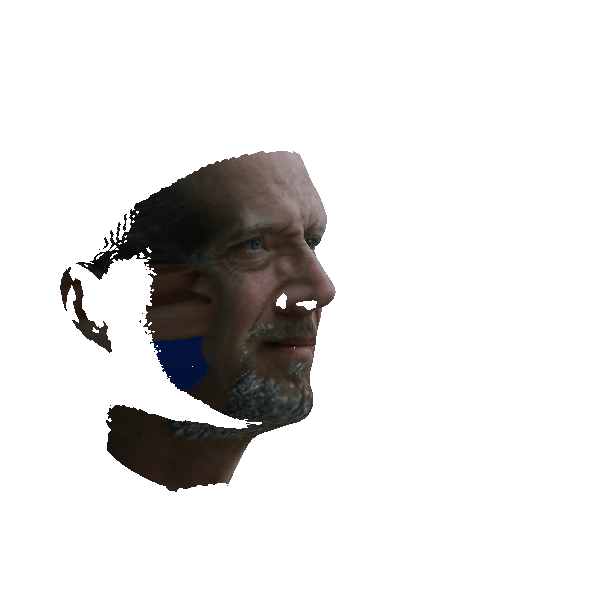}&
\includegraphics[height=\frontsize, clip=true,trim=0px	75px 310px 125px]{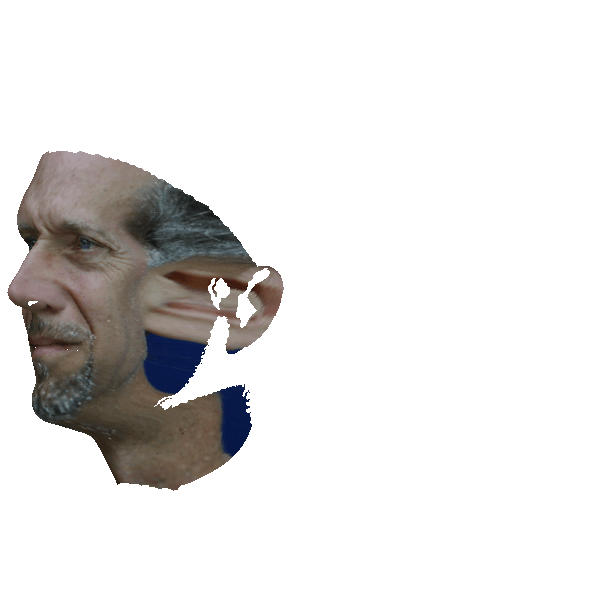}&
\includegraphics[height=\frontsize, clip=true,trim=50px 75px 260px 125px]{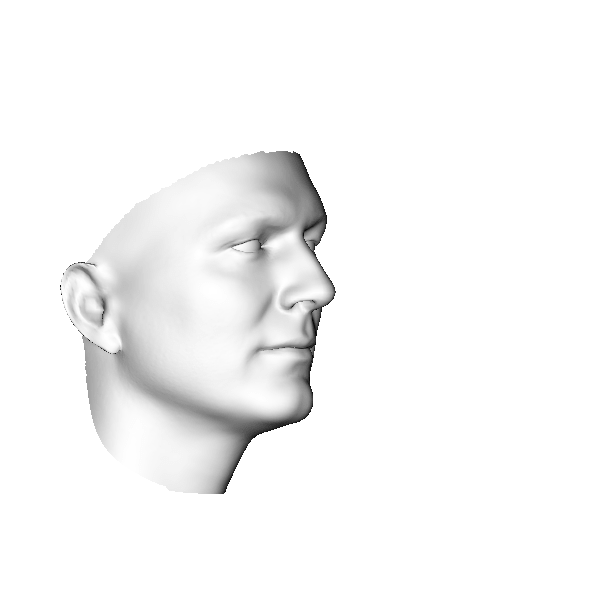}&
\includegraphics[height=\frontsize, clip=true,trim=0px	75px 310px 125px]{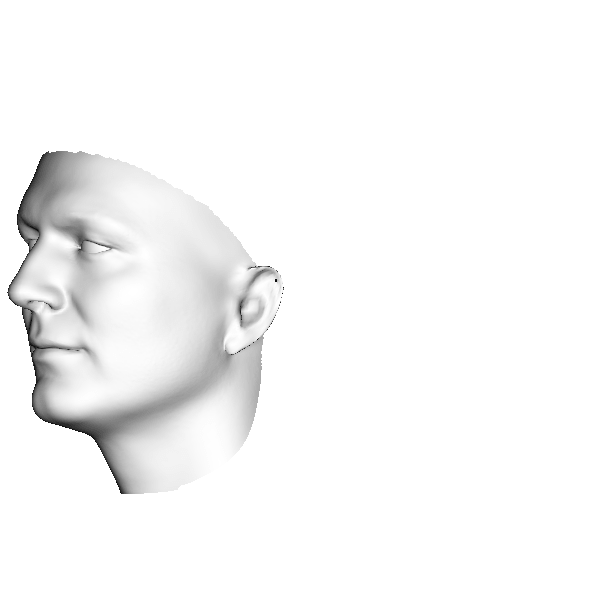}\\

\end{tabular}
}
\caption{Perspective fitting with flexibility modes. 1st Row: Landmark and edge fitting. 2nd/3rd Row: The first plus and minus flexibility components. Landmark distance is 1.79\% and surface distance is 10mm.}
\label{fig:persFlex}
\end{figure*}

In Figures \ref{fig:orthoFlex} and \ref{fig:persFlex} we show qualitative examples of the flexibility modes. We fit to a real image under both orthographic and perspective projection. We then compute the first flexibility mode and vary the shape in both directions such that the mean surface distance is 10mm. Despite the large change in the surface, the landmarks only vary by 1.14\% for orthographic and 1.79\% for perspective fitting. The correspondence when the texture is sampled onto the mesh remains similar. In other words, three very different surfaces provide plausible 3D explanations of the 2D data.

% Shading: unlike the bas-relief ambiguity where the transformation of the surface is linear, perspective transformation induces a non-linear transformation on the surface normals and hence on the appearance of the face. 

%\section{Discussion}

%For face recognition: if 3D shape similarity is to be used, the perspective ambiguity must be factored into this. Perhaps measure similarity after optimal projection to 2D?

\section{Conclusions}

In this paper we have studied ambiguities that arise when 3D face shape is estimated from monocular 2D geometric information. We have shown that 2D geometry (either sparse landmarks, semi-dense contours or dense vertex information) can be explained by a space of possible faces which vary significantly in 3D shape. We consider it surprising that the natural variability in face shape should include variations consistent with perspective transformation and that there are degrees of flexibility in face shape that have only a small effect on 2D geometry when pose is fixed. There are a number of interesting implications of these ambiguities.

In forensic image analysis, metric distances between features have been used as a way of comparing the identity of two face photographs. For example, \cite{porter2000anatomical} normalise face images by the interocular distance before using measurements such as the width of the face, nose and mouth to compare identities. We have shown that, after such normalisation, all distances between anthropometric features can be equal (up to the accuracy of landmarking) for two very different faces. This casts doubt on the use of such techniques in forensic image analysis and perhaps partially explains the studies that have demonstrated the weakness of these approaches \citep{kleinberg2007failure}.

Clearly, any attempt to reconstruct 3D face shape using 2D geometric information alone (such as in \citep{Blanz:04b,OSPAMI,Patel:09,Knothe:06,Bas:16}) will be subject to the ambiguity. Hence, the range of possible solutions is large and the likely accuracy low. If estimated 3D face shape is to be used for recognition, then the dissimilarity measure must account for the ambiguities we have described. On the other hand, CNN-based methods that learn to exploit any combination of features cannot necessarily overcome this uncertainty, as our results show. We believe that discriminative methods will require richer training data (either synthetic or real) containing significant variation in subject-camera distance, including small distances. Typically, there has been a reliance on web-crawled image databases, mainly of celebrities. These do not usually contain images at selfie distance and so new databases may be required.

For some face analysis problems, the purpose of fitting a statistical shape model is simply to establish correspondence. For example, it may be that face texture will be processed on the surface of the mesh, or that correspondence is required in order to compare different face textures for recognition. In such cases, these ambiguities are not important. Any solution that fits the dense 2D shape features (i.e. any from within the space of solutions described by the ambiguity) will suffice to correctly establish correspondence.

There are many ways in which the work can be extended. First, our model fitting approach could be cast in probabilistic terms. By seeking the least squares solution, we are obtaining the maximum likelihood explanation of the data under an assumption of Gaussian noise on the 2D landmarks. Our flexibility modes capture the likely parts of the posterior distribution but a fully probabilistic setting would allow the posterior to be explicitly modelled and uncertainty quantified.
Second, it would be interesting to investigate whether additional cues resolve the ambiguities. For example, an interesting follow-up to the work of \cite{Amberg:07} would be to investigate whether there is an ambiguity in uncalibrated {\it stereo} face images. Alternatively, we could investigate whether photometric cues (shading, shadowing and specularities) or statistical texture cues help to resolve the ambiguity. In the case of shading, it is not clear that this will be the case. Assuming illumination is unknown, it is possible that a transformation of the lighting environment could lead to shading which is consistent with (or at least close to) that of the target face \citep{smith2016perspective}.

%Bas relief ambiguity in the space of faces

\subsection*{Reproducible research}

A Matlab implementation of the fitting algorithms, the scripts necessary to recreate the results in this paper and videos visualising the ambiguities is available at:
% need to update this
{\footnotesize \url{http://www-users.cs.york.ac.uk/wsmith/faceambiguity}}.
For the purposes of creating the images in this paper, we developed a full featured off-screen renderer in Matlab. We make this publicly available at: {\footnotesize \url{https://github.com/waps101/MatlabRenderer}}.
%\vspace{-4mm}

% Can use something like this to put references on a page
% by themselves when using endfloat and the captionsoff option.
%\ifCLASSOPTIONcaptionsoff
%  \newpage
%\fi

% trigger a \newpage just before the given reference
% number - used to balance the columns on the last page
% adjust value as needed - may need to be readjusted if
% the document is modified later
%\IEEEtriggeratref{8}
% The "triggered" command can be changed if desired:
%\IEEEtriggercmd{\enlargethispage{-5in}}

% references section

% BibTeX users please use one of
\bibliographystyle{spbasic}      % basic style, author-year citations

\begin{appendices}

\section{SNLS derivatives}

Here we provide all of the derivatives required to optimise the SNLS objective functions. Specifically, we show how to compute the Jacobian matrices of the residual functions for the orthographic \eqref{eqn:orthoresSLS} and perspective case linearised via the DLT \eqref{eqn:perspresDLT}.

\paragraph{Matrix derivative identities}

The following identities are used in our derivations.

The derivatives of the axis-angle to rotation matrix function in \eqref{eqn:rtoR} are given by \cite{gallego2015compact}:
\begin{equation*}
    \frac{\partial\mathbf{R}}{\partial r_i} = 
    \begin{cases}
    \left[ \mathbf{e}_i \right]_{\times} & \textrm{if}\ \mathbf{r}=\mathbf{0} \\
    \frac{r_i\left[ \mathbf{r} \right]_{\times} + \left[ \mathbf{r}\times({\bf I}-\mathbf{R}(\mathbf{r}))\mathbf{e}_i \right]_{\times}}{\|\mathbf{r}\|^2}\mathbf{R}(\mathbf{r}) & \textrm{otherwise}
    \end{cases}
\end{equation*}
where $\mathbf{e}_i$ is the $i$th vector of the standard basis in $\R^3$.

The scalar derivative of the Kronecker product is:
\begin{equation*}
    \frac{\partial (\mathbf{X} \otimes \mathbf{Y})}{\partial x} = \frac{\partial \mathbf{X}}{\partial x} \otimes \mathbf{Y} + \mathbf{X} \otimes \frac{\partial \mathbf{Y}}{\partial x}.
\end{equation*}
For the special case involving the identity matrix, i.e. where $\mathbf{X}=\mathbf{I}$, this simplifies to:
\begin{equation*}
    \frac{\partial (\mathbf{I} \otimes \mathbf{Y})}{\partial x} = \mathbf{I} \otimes \frac{\partial \mathbf{Y}}{\partial x}.
\end{equation*}

The scalar derivative of the pseudoinverse $\mathbf{A}^+(x)$ of $\mathbf{A}$ at $x$ is given by:
\begin{multline*}
    \frac{\partial \mathbf{A}^+}{\partial x} = 
    -\mathbf{A}^+ \frac{\partial \mathbf{A}}{\partial x} \mathbf{A}^+ +
    \mathbf{A}^+\mathbf{A}^{+{\textrm{T}}}\frac{\partial \mathbf{A}^{\textrm{T}}}{\partial x}(\mathbf{I}-\mathbf{A}\mathbf{A}^+) + \\
    (\mathbf{I}-\mathbf{A}^+\mathbf{A})\frac{\partial \mathbf{A}^{\textrm{T}}}{\partial x}\mathbf{A}^{+\textrm{T}}\mathbf{A}^+
\end{multline*}
% Source: https://en.wikipedia.org/wiki/Moore%E2%80%93Penrose_inverse#Derivative

\paragraph{Orthographic case}

The derivatives of the matrix $\mathbf{A}(\mathbf{r},s)$ are given by:
\begin{equation*}
    \frac{\partial \mathbf{A}}{\partial s} = \begin{bmatrix} \left( \mathbf{I}_L \otimes \mathbf{P}\mathbf{R}(\mathbf{r}) \right)\mathbf{Q}_L & \mathbf{1}_L \otimes \mathbf{I}_2 \end{bmatrix},
\end{equation*}
\begin{equation*}
    \frac{\partial \mathbf{A}}{\partial r_i} = \begin{bmatrix} s\left( \mathbf{I}_L \otimes \mathbf{P}\frac{\partial\mathbf{R}}{\partial r_i} \right)\mathbf{Q}_L & \mathbf{0}_{2L\times 2} \end{bmatrix}.
\end{equation*}
The derivatives of the vector $\mathbf{y}(\mathbf{r},s)$ are given by:
\begin{equation*}
    \frac{\partial \mathbf{y}}{\partial s} = \left(\mathbf{I}_L \otimes \mathbf{P}\mathbf{R}(\mathbf{r}) \right)\mathbf{\bar{s}}
\end{equation*}
\begin{equation*}
    \frac{\partial \mathbf{y}}{\partial r_i} = s\left[ \left(\mathbf{I}_L \otimes \mathbf{P}\frac{\partial \mathbf{R}}{\partial r_i} \right)\mathbf{\bar{s}}\right].
\end{equation*}

From the components above we can compute the derivatives of the residual function:
\begin{multline*}
    \frac{\partial \mathbf{d}_{\textrm{ortho}}}{\partial s} = 
    \left(  
    % dAA^+/ds
    \mathbf{A}(\mathbf{r},s)\frac{\partial \mathbf{A}^+}{\partial s} + \frac{\partial \mathbf{A}}{\partial s}\mathbf{A}^+(\mathbf{r},s)
    \right)
    \mathbf{y}(\mathbf{r},s) + \\
    \mathbf{A}(\mathbf{r},s)\mathbf{A}^+(\mathbf{r},s) \frac{\partial \mathbf{y}}{\partial s} - \frac{\partial \mathbf{y}}{\partial s},
\end{multline*}
\begin{multline*}
    \frac{\partial \mathbf{d}_{\textrm{ortho}}}{\partial r_i} = 
    \left(  
    % dAA^+/ds
    \mathbf{A}(\mathbf{r},s)\frac{\partial \mathbf{A}^+}{\partial r_i} + \frac{\partial \mathbf{A}}{\partial r_i}\mathbf{A}^+(\mathbf{r},s)
    \right)
    \mathbf{y}(\mathbf{r},s) + \\
    \mathbf{A}(\mathbf{r},s)\mathbf{A}^+(\mathbf{r},s) \frac{\partial \mathbf{y}}{\partial r_i} - \frac{\partial \mathbf{y}}{\partial r_i}.
\end{multline*}
Finally, the Jacobian, $\mathbf{J}_{\mathbf{d}_{\textrm{ortho}}}({\bf r},s)$, is obtained by stacking these four vectors into a $2L\times 4$ matrix:
\begin{equation*}
\mathbf{J}_{\mathbf{d}_{\textrm{ortho}}}({\bf r},s) = \begin{bmatrix}
\frac{\partial \mathbf{d}_{\textrm{ortho}}}{\partial r_1} & \frac{\partial \mathbf{d}_{\textrm{ortho}}}{\partial r_2} & \frac{\partial \mathbf{d}_{\textrm{ortho}}}{\partial r_3} & \frac{\partial \mathbf{d}_{\textrm{ortho}}}{\partial s}
\end{bmatrix}.
\end{equation*}

\paragraph{Perspective case} 

% \begin{equation*}
%  {\bf E}(f) = {\bf I}_L \otimes {\bf K}(f) 
% \end{equation*}
% \begin{equation*}
%  {\bf F}({\bf r}) = \begin{bmatrix}
%  \left( {\bf I}_L \otimes {\bf R}({\bf r}) \right) {\bf Q}_L & {\bf 1}_L \otimes {\bf I}_3
% \end{bmatrix}
% \end{equation*}
% \begin{equation*}
%  \mathbf{B}(\mathbf{r},f) = {\bf DE}(f){\bf F}({\bf r})
% \end{equation*}
% \begin{equation*}
% \mathbf{z}(\mathbf{r},f) =
%  -{\bf D}\left({\bf I}_L \otimes \left[{\bf K}(f)\mathbf{R}(\mathbf{r})\right]\right)\overline{\bf s}
% \end{equation*}

The derivatives of the matrix $\mathbf{B}(\mathbf{r},f)$ are given by:
\begin{equation*}
\frac{\partial \mathbf{B}}{\partial f} = {\bf D} \frac{\partial \mathbf{E}}{\partial f}  {\bf F}({\bf r})
\quad\textrm{and}\quad
\frac{\partial \mathbf{B}}{\partial r_i} = {\bf DE}(f) \frac{\partial \mathbf{F}}{\partial r_i},
\end{equation*}
where
\begin{equation*}
\frac{\partial \mathbf{E}}{\partial f} = {\bf I}_L \otimes \frac{\partial \mathbf{K}}{\partial f}
\quad\textrm{and}\quad
\frac{\partial \mathbf{F}}{\partial r_i} = \begin{bmatrix}
 \left( {\bf I}_L \otimes \frac{\partial \mathbf{R}}{\partial r_i} \right) {\bf Q}_L & \mathbf{0}_{3L\times 3}
\end{bmatrix},
\end{equation*}
and
\begin{equation*}
\frac{\partial \mathbf{K}}{\partial f} = \begin{bmatrix}
1 & 0 & 0  \\
0 & 1 & 0  \\
0 & 0 & 0 
\end{bmatrix}.
\end{equation*}
The derivatives of the vector $\mathbf{z}(\mathbf{r},f)$ are given by:
\begin{equation*}
\frac{\partial \mathbf{z}}{\partial f} =
 -{\bf D}\left({\bf I}_L \otimes \left[ \frac{\partial \mathbf{K}}{\partial f} \mathbf{R}(\mathbf{r})\right]\right)\overline{\bf s},
\end{equation*}
\begin{equation*}
\frac{\partial \mathbf{z}}{\partial r_i} =
 -{\bf D}\left({\bf I}_L \otimes \left[{\bf K}(f) \frac{\partial \mathbf{R}}{\partial r_i} \right]\right)\overline{\bf s}.
\end{equation*}
From the components above we can compute the derivatives of the residual function:
\begin{multline*}
    \frac{\partial \mathbf{d}_{\textrm{persp}}^{\textrm{DLT}}}{\partial f} = 
    \left(  
    % dAA^+/ds
    \mathbf{B}(\mathbf{r},f)\frac{\partial \mathbf{B}^+}{\partial f} + \frac{\partial \mathbf{B}}{\partial f}\mathbf{B}^+(\mathbf{r},f)
    \right)
    \mathbf{z}(\mathbf{r},f) + \\
    \mathbf{B}(\mathbf{r},f)\mathbf{B}^+(\mathbf{r},f) \frac{\partial \mathbf{z}}{\partial f} - \frac{\partial \mathbf{z}}{\partial f},
\end{multline*}
\begin{multline*}
    \frac{\partial \mathbf{d}_{\textrm{persp}}^{\textrm{DLT}}}{\partial r_i} = 
    \left(  
    % dAA^+/ds
    \mathbf{B}(\mathbf{r},f)\frac{\partial \mathbf{B}^+}{\partial r_i} + \frac{\partial \mathbf{B}}{\partial r_i}\mathbf{B}^+(\mathbf{r},f)
    \right)
    \mathbf{z}(\mathbf{r},f) + \\
    \mathbf{B}(\mathbf{r},f)\mathbf{B}^+(\mathbf{r},f) \frac{\partial \mathbf{z}}{\partial r_i} - \frac{\partial \mathbf{z}}{\partial r_i}.
\end{multline*}
Finally, the Jacobian, $\mathbf{J}_{\mathbf{d}_{\textrm{persp}}^{\textrm{DLT}}}({\bf r},f)$, is obtained by stacking these four vectors into a $3L\times 4$ matrix:
\begin{equation*}
\mathbf{J}_{\mathbf{d}_{\textrm{persp}}^{\textrm{DLT}}}({\bf r},f) = \begin{bmatrix}
\frac{\partial \mathbf{d}_{\textrm{persp}}^{\textrm{DLT}}}{\partial r_1} & \frac{\partial \mathbf{d}_{\textrm{persp}}^{\textrm{DLT}}}{\partial r_2} & \frac{\partial \mathbf{d}_{\textrm{persp}}^{\textrm{DLT}}}{\partial r_3} & \frac{\partial \mathbf{d}_{\textrm{persp}}^{\textrm{DLT}}}{\partial f}
\end{bmatrix}.
\end{equation*}

\end{appendices}

% that's all folks
\end{document}